%% file: main.tex
\documentclass[]{TEAI}

\input{preamble}

\title{Does the Same Token Mean the Same State? \\ MoE Routing as Signal for Reasoning Control}

\author{
Kang Chen\textsuperscript{1},
Minshen Yu\textsuperscript{1},
Junjie Nian\textsuperscript{1},
Yaoning Wang\textsuperscript{1},
Yixin Cao\textsuperscript{1,2$\dagger$},
Yugang Jiang\textsuperscript{1}
}

\affiliation[1]{\mbox{Fudan University}}
\affiliation[2]{\mbox{Shanghai Innovation Institute}}

\checkdata[Email]{\email{kchen24@m.fudan.edu.cn}, \email{yxcao@fudan.edu.cn}$^\dagger$ (Correspondence)}
\checkdata[Web]{\url{https://CckFdu.com/RAD}}

\abstract{
  In sparse Mixture-of-Experts language models, does the same token id imply the same router state and the same experts producing it? Holding the emitted token id fixed at repeated anchors, we find it does not: the experts that produce it still separate task context, trajectory history, and reasoning-effort mode. This residual structure supports test-time control: near \emph{boundary} anchors (the final-response transition) and \emph{delimiter} anchors (which open the answer, e.g.\ \texttt{\textbackslash boxed\{} or code fences), routing neighborhoods already align with final-answer basins at a marker-only readout and strongest when the routing is read at the answer opening. We operationalize this as \textbf{RAD} (Routing Agreement Decoding), an answer-string-free multi-rollout selector: it locates a fixed anchor, represents each rollout by its anchor-window MoE routing states, and returns the densest Weighted-Jaccard $K$-NN route-basin center, without parsing, normalizing, executing, or voting over answer strings. Across 10 sparse-MoE configurations (gpt-oss, Qwen3-MoE) and 6 datasets spanning math, GPQA, and code, RAD is on par with Majority where string voting is well-posed, with small positive paired deltas (RAD $73.9$ / RAD+DC $74.2$ vs.\ Majority $73.6$).  Like majority voting, RAD is not a verifier: a dense \emph{wrong} basin can still win. Its value is the interface: the same selector gives direct pass@1 on code, where exact-string voting is ill-defined, and the same routing-density principle, re-anchored to the agentic boundary, improves best-of-16 patch selection on SWE-bench Verified over random, where patches have no answer string to vote on.
}

\begin{document}
\maketitle

\vspace{-0.5em}

\input{sections/introduction}

\input{sections/related_work}

\input{sections/analysis}

\input{sections/method}

\input{sections/experiments}

\input{sections/limitations}

\input{sections/conclusion}

\section*{Acknowledgments}
We thank Yicheng Yang and Weijian Shi for assistance with improving the visual presentation of Figures~\ref{fig:exp1} and~\ref{fig:pipeline}.

\small
\bibliographystyle{plainnat}
\bibliography{references}

\appendix

\section{Technical appendices and supplementary material}

This appendix contains the full answer-string-free protocol details, exact-statistics derivations, anchor coverage tables, ablation studies, failure-case analyses, and per-dataset / per-model results that the main text references but does not include for space.

\input{sections/appendix}

\end{document}

%% file: preamble.tex
\usepackage[utf8]{inputenc}

\usepackage{nicefrac}

\usepackage{longtable}

\usepackage{wrapfig}

\usepackage{algorithm}
\usepackage{algorithmic}

\usepackage{mdframed}
\definecolor{harmonyblue}{RGB}{225,244,248}
\mdfdefinestyle{harmonystyle}{%
  backgroundcolor=harmonyblue,linecolor=harmonyblue,linewidth=0pt,%
  innerleftmargin=8pt,innerrightmargin=8pt,innertopmargin=2pt,innerbottommargin=2pt,%
  skipabove=3pt,skipbelow=3pt,leftmargin=0pt,rightmargin=0pt}

%% file: sections/introduction.tex
\vspace*{-2.5em}
\noindent\includegraphics[width=\linewidth]{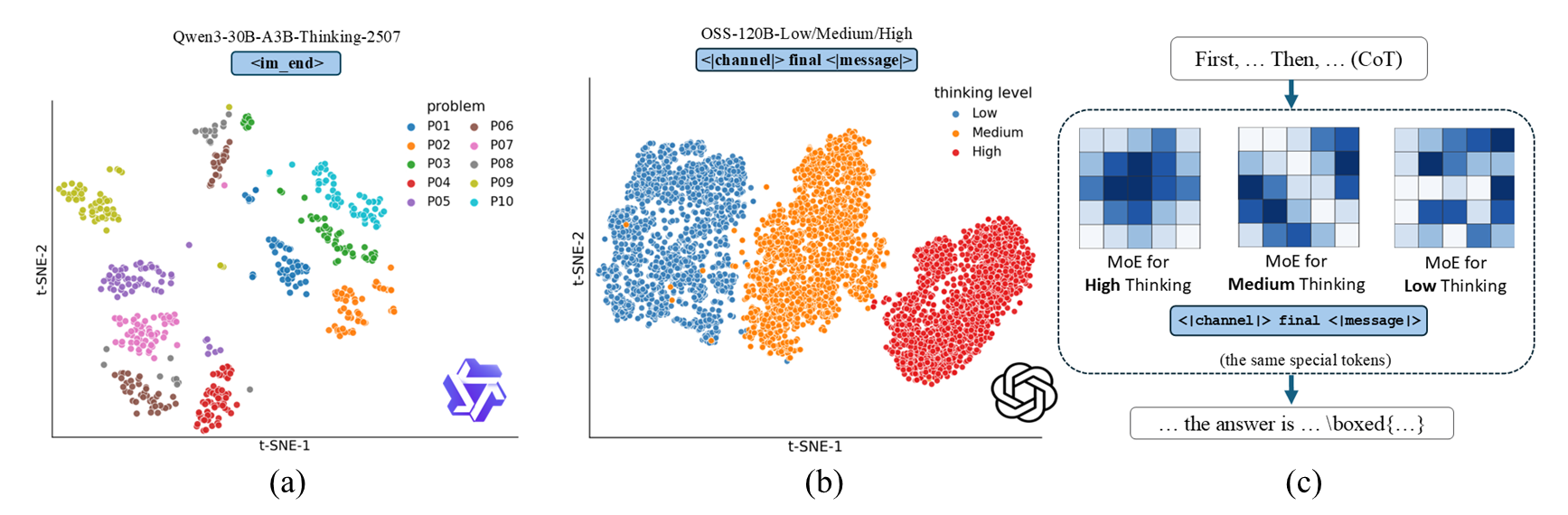}\par
\captionof{figure}{\textbf{Same token, different generating experts.} At a fixed emitted token id, the generating experts cluster by problem identity (Qwen EOS) and reasoning-effort mode (OSS boundary). \emph{A,B}: t-SNE of pairwise WJ; \emph{C}: schematic.}
\label{fig:exp1}
\clearpage

\section{Introduction}
\label{sec:intro}

Multi-rollout reasoning improves LLM reliability: instead of trusting a single sample, a system draws many reasoning traces and aggregates them, often via self-consistency or majority voting~\citep{wei2022chain,wang2023selfconsistency}, on the bet that correct paths recur more stably.

The bottleneck is the interface, not the voting rule. Aggregation happens \emph{after} generation, over a visible answer string the controller extracts, canonicalizes, and votes over. This is fine for short canonical answers, but fragile or ill-defined when outputs are open-ended, executable, or lexically diverse. One wants a selector that arbitrates completed rollouts \emph{without per-dataset answer-string extraction or voting}, yet stays on par with string voting where it is well-posed.

Sparse Mixture-of-Experts models expose another signal. To produce each token, an MoE router engages a few experts with routing weights~\citep{shazeer2017outrageously,fedus2022switch} (in effect, a record of \emph{which experts drove that token's generation}), not decoded text, yet produced online from the hidden state that drives generation. This raises two coupled questions, one descriptive, one operational: \textbf{\mbox{Does the same token mean the same internal state?} And if not, can the residual routing geometry arbitrate among completed rollouts where answer-string voting is undefined?} If so, routing is a white-box agreement signal voting cannot see.

We study this with \textbf{anchor-conditioned routing}: align rollouts at repeated anchor token ids and compare routing at those fixed positions, separating anchor type from window width, across \emph{trajectory}, \emph{boundary}, and \emph{delimiter} anchor families (definitions and \texttt{type@W} notation in \S\ref{sec:method}).

First, descriptively (Figure~\ref{fig:exp1}), token identity is not a sufficient statistic for router state: at the same emitted token id, the experts that produce it still separate problem identity and reasoning-effort regimes, under $5\%$ cross-problem and cross-effort nearest-neighbor leakage in routing space.

Second, operationally, routing neighborhoods align with \emph{answer basins} (rollouts sharing a normalized answer): \texttt{delimiter@marker} is already informative without conditioning on answer contents, and the wider \texttt{delimiter@16} gives the strongest $K$-NN basin purity (the headline setting, Table~\ref{tab:main}), which tracks selector accuracy, so dense routing neighborhoods approximate dense answer basins.

These observations motivate \textbf{Routing Agreement Decoding (RAD)}. For one problem's completed rollout pool, RAD locates a fixed anchor by token id, represents each rollout by its anchor-window routing, and selects the densest Weighted-Jaccard $K$-NN route basin (notation in \S\ref{sec:method}). During selection, RAD never parses, normalizes, compares, executes, or votes over final answer contents. The selected answer string is read only \emph{after} this routing-only decision. Like Majority voting, it estimates shared commitment rather than truth: a dense route basin can be wrong, exactly as textual Majority follows a dense wrong consensus, differing only in interface: Majority counts answer strings, RAD aggregates routing structure.

We evaluate RAD across 10 sparse-MoE configurations and 6 datasets (gpt-oss and Qwen3-MoE; math, GPQA, LiveCodeBench). On the well-posed math+GPQA regime it is on par with textual Majority, modestly ahead in our pools, reading no answer strings. Its distinctive value is the answer-string-free regime: on code, where raw outputs are lexically fragmented and exact-string voting is ill-defined, the same selector returns direct pass@1; and on agentic SWE-bench Verified, re-anchored at the boundary, it beats random selection by $+4$--$5$\,pp on the decidable subset where patches have no votable answer string.

\paragraph{Contributions.} We (i)~show identical token ids can be generated through different MoE expert routes; (ii)~show routing neighborhoods align with final-answer basins, strongest at delimiter readouts; and (iii)~propose \textbf{RAD}, a routing-based selector that is competitive with textual Majority in well-posed string voting settings, applies directly to code-generation outputs, and extends, via boundary re-anchoring, to agentic SWE-bench Verified patch selection.

%% file: sections/related_work.tex
\section{Related Work}
\label{sec:related}

\paragraph{Multi-rollout reasoning and test-time control.}
Chain-of-thought prompting and self-consistency improve reasoning by sampling multiple paths and aggregating final answers~\citep{wei2022chain,wang2023selfconsistency}. Recent controllers reduce this cost using early stopping, rationale or prefix agreement, confidence scores, or thought-level pruning~\citep{li2024escape,wan2025rasc,hong2025slim,jindal2026polr,fu2026deepconf}. These methods arbitrate through visible text, answer strings, or scalar scores. RAD instead selects from a completed rollout pool using MoE routing agreement and never compares final answer strings, which lets the same selector extend to agentic best-of-$N$ patch selection, where candidates are multi-step trajectories that expose no answer string to vote on.

\paragraph{Internal signals for selection.}
Mechanism-aware evaluation studies how internal activations, neurons, or utilization patterns relate to model behavior~\citep{cao2025mui,zhang2025reasoning,chen2025nad,chen2026nex}. RAD belongs to this white-box family, but uses sparse MoE routing states rather than dense hidden activations or neuron caches, and treats routing density as an answer-basin signal rather than a correctness oracle.

\paragraph{MoE routing.}
Sparse MoE models route each token through a small subset of experts~\citep{shazeer2017outrageously,lepikhin2021gshard,fedus2022switch,jiang2024mixtral,dai2024deepseekmoe}. Prior work shows that routing is shaped by token identity as well as hidden-state geometry~\citep{xue2024openmoe,hayashi2026layerwise,wang2026myth,ying2025moemui}. RAD turns this confound into an alignment device: repeated anchor token ids hold the surface token fixed, while residual routing geometry is used to identify redundancy at trajectory anchors and dense answer basins around boundary and delimiter anchors.

%% file: sections/analysis.tex
\section{Analysis of Anchor-Conditioned Routing}
\label{sec:analysis}

We use anchor-conditioned routing to ask whether MoE router states (which experts drive each token's generation) contain rollout-level structure beyond token identity. Exp~I holds the emitted token id fixed and tests whether the experts that generate it still separate upstream context. Exp~II tests whether routing neighborhoods align with final-answer basins near final-answer production. We do not interpret individual experts or claim that routing predicts correctness.

\subsection{Answer-string-free protocol, route representations, and diagnostic}
\label{sec:analysis:protocol}

\paragraph{Setup and access policy.}
For each problem $p$, we sample $N{=}64$ rollouts and record both token ids and sparse MoE route vectors $R_i(t)$ (selected expert ids and post-softmax weights across layers). The index convention is token-synchronous but prefix-causal: $R_i(t)$ is recorded in the forward pass that predicts $y_{i,t}$ from $y_{i,<t}$, so it is aligned with token $t$, not conditioned on $y_{i,t}$ itself; in words, $R_i(t)$ records which experts drove the generation of $y_{i,t}$. Anchors are matched on the recorded token-id stream, not re-tokenized text. We use three anchor types: \emph{trajectory anchors}, recurring intermediate markers such as \texttt{so}, \texttt{now}, or paragraph breaks; \emph{boundary anchors}, final-response transition markers such as \texttt{<|channel|>final<|message|>} or \texttt{</think>}; and \emph{delimiter anchors}, fixed answer-opening delimiters such as \texttt{\textbackslash boxed\{} or code fences. Thus \texttt{delimiter@marker} is marker-only and contains no answer-region rows, whereas the headline \texttt{delimiter@16} is a fixed post-marker routing readout. At runtime, RAD uses token ids only for anchor lookup and routing tensors only in fixed windows; it never inspects decoded answer text, answer-class/majority labels, or correctness/execution signals (see Appendix~\ref{app:no-answer} for the access boundary).

\paragraph{Route representation and Weighted Jaccard.}
At token $t$ and MoE layer $\ell$, the router selects $k_{\mathrm{route}}$ experts with normalized weights. Across our model pool $k_{\mathrm{route}}\in\{4,8,10\}$; for Qwen3-Next the constant shared expert is excluded from the WJ histogram (Appendix~\ref{app:routing-capture}). Concatenating the per-layer sparse vectors gives $R_i(t)\in\mathbb{R}^{L\cdot E}$, and averaging over an anchor window yields a per-rollout route vector. We compare two route vectors $u,v$ by Weighted Jaccard:
\begin{equation}
\mathrm{WJ}(u,v) \;=\; \frac{\sum_e \min(u_{e},v_{e})}{\sum_e \max(u_{e},v_{e})}.
\label{eq:wj}
\end{equation}
WJ preserves routed expert mass rather than binarizing fired experts; the binary comparison appears in \S\ref{sec:exp:analysis}.

\paragraph{Diagnostic metric.}
For analysis only, we measure chance-corrected $K$-NN answer-basin purity:
\begin{equation}
A_{K_{\mathrm{nn}},r} \;=\; \frac{p_{K_{\mathrm{nn}},r} - b}{1 - b},
\quad
p_{K_{\mathrm{nn}},r} = \frac{1}{N}\sum_{i}\frac{1}{K_{\mathrm{nn}}}\sum_{j\in\mathrm{NN}_{K_{\mathrm{nn}}}(i,r)}\mathbf{1}[a_i=a_j],
\quad
b = \frac{\sum_{a} n_a(n_a-1)}{N(N-1)}.
\label{eq:knn-purity}
\end{equation}
Neighbors are defined in routing space (WJ at readout $r$) and $a_i$ is the normalized final-answer basin label. $A_{K_{\mathrm{nn}},r}{=}0$ means routing neighbors are random with respect to answer basins; $A_{K_{\mathrm{nn}},r}{=}1$ means perfect local basin alignment. RAD never reads this diagnostic at runtime; implementation details (degenerate cohorts, chance-baseline forms, ground-truth vs.\ majority robustness) are in Appendix~\ref{app:knn-purity}. After selection, we report accuracy / pass@1, token saving, and lift over the exact random 95\% envelope.

\subsection{Exp~I: The same token can be generated by different experts}
\label{sec:analysis:exp1}

A first concern is that anchor-conditioned routing might simply recover token identity. Exp I addresses this by holding the emitted anchor token id fixed across rollouts and comparing the prefix-causal MoE routes used to produce that fixed token. Each route is recorded while predicting the anchor and therefore does not condition on the anchor token itself (\S\ref{sec:analysis:protocol}). We test whether these routes nevertheless separate upstream factors that token identity alone cannot encode. Figure~\ref{fig:exp1} (introduced in \S\ref{sec:intro}) visualizes this with pairwise WJ distances over per-rollout sparse expert-weight histograms; color labels are used only for visualization, not to fit the embedding.

\begin{itemize}
\item \textbf{Qwen EOS token.} At the same Qwen \texttt{<|im\_end|>} token on BRUMO25, the generating route preserves problem identity across ten random problems, with less than $5\%$ cross-problem neighborhood leakage.
\item \textbf{OSS boundary anchor.} At the fixed OSS final-response boundary anchor (the Harmony final-channel triplet \texttt{<|channel|>final<|message|>}) on AIME25, the generating route separates Low, Medium, and High reasoning-budget settings, with less than $1\%$ cross-level contamination.
\end{itemize}

\paragraph{Insight 1.}
\emph{Token identity is not a sufficient statistic for routing state.} Even when the emitted token id is held fixed, the experts that generate it retain information about task family, problem identity, and reasoning-budget mode. Anchor-conditioned routing is therefore not vacuous: an identical anchor token can be generated by a different set of experts, because the upstream context driving that generation step differs.

\paragraph{Bridge to Exp~II.}
Exp~I is a validity test, not yet a controller test. A route representation can separate tasks, problems, or budget modes without being useful for multi-rollout selection. A controller needs a more specific property: routing neighborhoods should reflect rollout-level relations that matter for control, especially same-answer-basin agreement near final-answer production. Exp~II tests this operational condition; only an affirmative answer there licenses RAD in \S\ref{sec:method}.


\subsection{Exp~II: routing neighborhoods align with answer basins, strongest at delimiter windows}
\label{sec:analysis:exp2}

Exp~II asks whether the non-token residual from Exp~I aligns with the answer-basin structure that a multi-rollout controller can exploit. We use $A_{K_{\mathrm{nn}},r}$ from Eq.~\ref{eq:knn-purity} as a post-hoc diagnostic. High $A_{K_{\mathrm{nn}},r}$ means nearby rollouts in routing space tend to share a normalized final-answer basin; low $A_{K_{\mathrm{nn}},r}$ means routing neighborhoods do not predict answer agreement. Again, $A_{K_{\mathrm{nn}},r}$ is never read by the selector at runtime; it is used only for analysis. Appendix~\ref{app:knn-purity} reports robustness checks, including ground-truth-labelled variants.

\begin{figure}[t]
  \centering
  \begin{subfigure}[b]{0.63\linewidth}
    \centering
    \includegraphics[width=\linewidth]{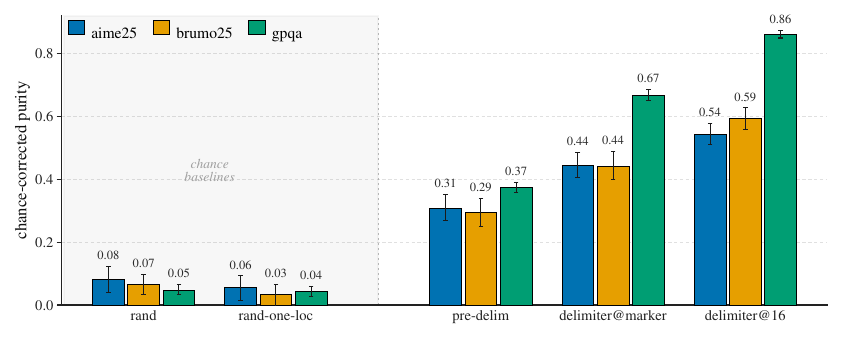}
  \end{subfigure}
  \hfill
  \begin{subfigure}[b]{0.35\linewidth}
    \centering
    \includegraphics[width=\linewidth]{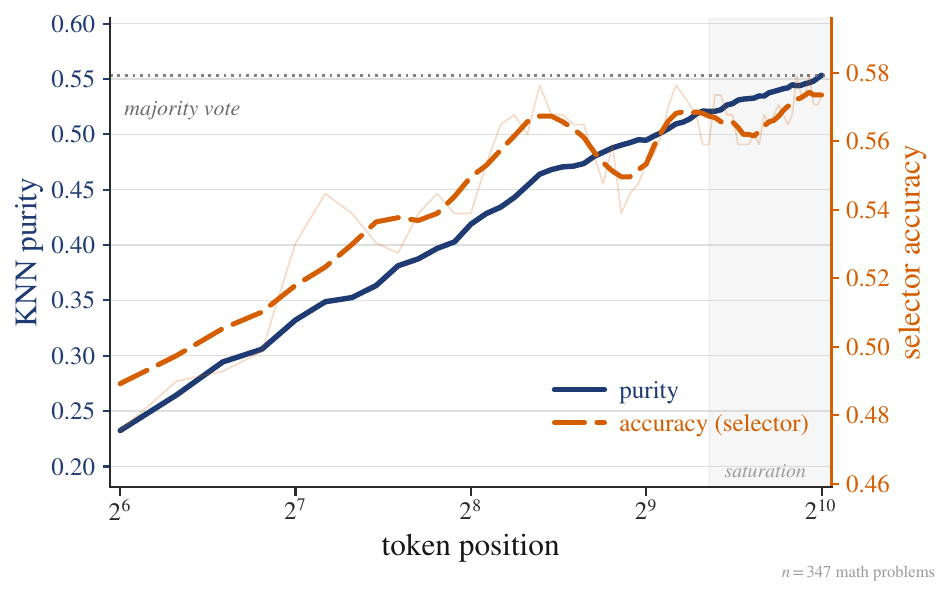}
  \end{subfigure}
  \caption{\textbf{Delimiter-window routing is the strongest basin-aligned readout, and its alignment tracks selector accuracy.} \emph{Left:} chance-corrected $K$-NN purity ($K_{\mathrm{nn}}{=}1$) for five routing readouts: \texttt{rand}, \texttt{rand-one-loc}, \texttt{pre-delim}, \texttt{delimiter@marker}, and \texttt{delimiter@16}; 8 MoE models, $N{=}1{,}180$ (model,problem) cells, 95\% bootstrap CIs. The marker-only \texttt{delimiter@marker} readout is already informative, while \texttt{delimiter@16} reaches $0.86/0.59/0.54$ on gpqa/brumo25/aime25. \emph{Right:} math-only \texttt{delimiter@16} purity ($K_{\mathrm{nn}}{=}1$) and routing-density selector accuracy vs.\ $16t$, $t\in[4,64]$, 347 math (model, problem) cells (8 models $\times$ 4 datasets); dotted = $t{=}64$ majority (\S\ref{sec:analysis:exp2}).}
  \label{fig:evidence-a}
\end{figure}

\paragraph{Anchor dependence.}
Figure~\ref{fig:evidence-a} (left) compares five routing readouts: two length-matched random controls, a pre-delimiter 16-token window, the marker-only \texttt{delimiter@marker} readout, and the headline \texttt{delimiter@16} window. The \texttt{delimiter@16} window used by RAD wins by a wide margin, while \texttt{delimiter@marker} already carries nontrivial state without conditioning on answer contents, and the two random controls are at chance. Among the four length-matched 16-token readouts (all but the marker-only \texttt{delimiter@marker}), the \texttt{delimiter@16} gain is driven by \emph{where} routing is read, not merely by how many states are averaged. The full anchor-round trajectory across trajectory-anchor occurrences is deferred to Appendix~\ref{app:knn-purity}.

\paragraph{Accuracy correspondence.}
Figure~\ref{fig:evidence-a} (right) compares \texttt{delimiter@16} basin purity with routing-density selector accuracy as the per-rollout reading budget grows, $16t$ tokens for $t{=}4$ to $64$ (token positions $\{2^{6},\ldots,2^{10}\}$), with the rollout count fixed at $N{=}64$. On the math subset (aime24, aime25, brumo25, hmmt25), purity rises monotonically with $t$ and selector accuracy rises in step (Spearman $\rho{=}{+}1.00$ and ${+}0.72$), with the selector's accuracy approaching the $t{=}64$ textual-majority baseline near the right edge of the budget range. This supports a distribution-level claim: when delimiter-window routing neighborhoods better match answer-basin neighborhoods, selecting the densest routing basin better approximates selecting the most populated answer basin. Yet a dense route basin can still be wrong when many rollouts converge to the same erroneous solution (\S\ref{sec:limitations}); routing density estimates shared commitment, not correctness.

\paragraph{Insight 2.}
\emph{$K$-NN basin purity quantifies how strongly routing neighborhoods align with final-answer basins.} This alignment is strongest in \texttt{delimiter@16} (while the marker-only \texttt{delimiter@marker} is already informative, with full per-model selection accuracy in Appendix~\ref{app:marker-only}) and tracks selector accuracy across rollout budgets. RAD therefore selects among completed rollouts by WJ-KNN density in routing space, without using decoded answer strings: around boundary and delimiter anchors, the densest routing basin often approximates the densest answer basin.

%% file: sections/method.tex
\section{Method: Routing Agreement Decoding}
\label{sec:method}

\begin{figure}[t]
  \centering
  \includegraphics[width=\linewidth]{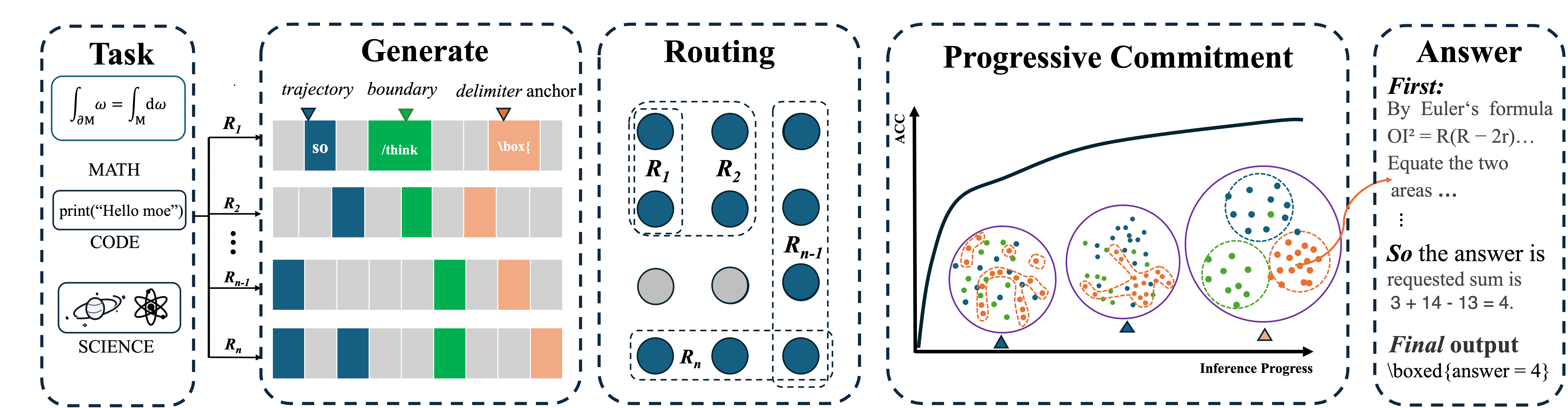}
  \caption{\textbf{RAD overview.} \emph{Task:} for each problem (mathematics, code, or science) we draw $N{=}64$ rollouts $R_1,\dots,R_N$. \emph{Generate:} rollouts are aligned by token id at a fixed anchor; RAD supports three anchor types, a \emph{trajectory} anchor (a discourse marker such as \texttt{So}), a \emph{boundary} anchor (a final-response marker such as \texttt{</think>}), and the headline \emph{delimiter} anchor (an answer-opening delimiter such as \texttt{\textbackslash boxed\{} or a code fence), read over a $W{=}16$ window and denoted \texttt{delimiter@16}. \emph{Routing:} the window is averaged into a per-rollout routing vector $z_i$ (\S\ref{sec:method:rad}), and pairwise Weighted-Jaccard similarities form the consensus matrix. \emph{Progressive commitment:} as inference proceeds the routing states concentrate into answer basins, and RAD selects the rollout with the highest $K$-NN agreement density $q_i$, a representative of the densest route basin. \emph{Answer:} the selected rollout's final answer string is read only once after selection, for evaluation; the selector never parses, compares, or votes over final answer strings.}
  \label{fig:pipeline}
\end{figure}

The empirical analysis of \S\ref{sec:analysis} motivates an anchor-aligned routing selector parameterized by an anchor type and a window width. First, routing must be conditioned on an anchor token id, since identical token ids can still be generated by different experts (\S\ref{sec:analysis:exp1}). Second, \S\ref{sec:analysis:exp2} shows an access/signal tradeoff: marker-only readouts such as \texttt{delimiter@marker} are already informative, while the wider \texttt{delimiter@16} readout has the strongest alignment with answer basins. The headline \emph{RAD} setting therefore reads routing in a fixed \texttt{delimiter@16} window and selects the densest Weighted-Jaccard $K$-NN route basin. Figure~\ref{fig:pipeline} illustrates the overall framework.

\paragraph{Anchor and window notation.}
We separate the alignment marker from the routing window. An anchor is a token position used only to align routing readouts across rollouts; a readout is an anchor type plus a window width. We use three anchor types. \emph{Trajectory anchors} are recurring discourse markers in the reasoning trace, used one at a time (by default \texttt{So}; \texttt{Now} and the paragraph break \texttt{.\textbackslash n\textbackslash n} are alternatives examined in the analysis, App.~\ref{app:frontier}, never combined in one readout). \emph{Boundary anchors} are final-response transition markers, such as \texttt{<|channel|>final<|message|>} or \texttt{</think>}. \emph{Delimiter anchors} are fixed answer-opening delimiters, such as \texttt{\textbackslash boxed\{} in math or code fences in programming tasks. We write \texttt{type@W} for the half-open $W$-token window $[a_i,a_i+W)$ and \texttt{type@marker} for the special case that spans \emph{all} tokens of the anchor marker: if the marker tokenizes to $m$ rows, \texttt{type@marker} is \texttt{type@}$m$, i.e.\ the window $[a_i,a_i+m)$ over the rows $R_i(a_i),\dots,R_i(a_i{+}m{-}1)$ (a multi-token delimiter such as \texttt{\textbackslash boxed\{} or a code fence contributes \emph{all} of its rows, with $m$ set by the tokenizer; the rows are averaged per token, Eq.~\ref{eq:w16}, so markers of different lengths stay comparable). Under the prefix-causal index (\S\ref{sec:analysis:protocol}) a marker row conditions at most on \emph{earlier marker tokens} (a fixed-format string before answer text), so \texttt{delimiter@marker} is a marker-only readout; the headline \texttt{delimiter@16} instead extends the window past the marker ($16{>}m$), so its rows beyond the marker condition on the first few answer-opening token positions, and it is used because it gives the strongest empirical selection signal.

\paragraph{Runtime interface and access boundary.}
For a problem $p$, let $\mathcal{D}_p=\{(y_i,R_i)\}_{i=1}^N$ be the completed rollout pool, with $y_i=(y_{i,0},\ldots,y_{i,T_i-1})$ the generated token sequence. The route vector $R_i(t)$ is the sparse $L\cdot E$ MoE routing vector recorded during the forward pass that predicts $y_{i,t}$ from the prefix $y_{i,<t}$; it is therefore index-aligned with token $y_{i,t}$ and its sampled logprob, but is \emph{not} conditioned on $y_{i,t}$ itself. The selector uses token ids only to locate anchors and extracts routing states only from the fixed readout window. It never parses, normalizes, compares, executes, or votes over final answer strings, and it does not use correctness labels, execution results, textual majority labels, or answer-token ids inside the window as features.

RAD is \emph{answer-string-free} in all variants. Its access level is configurable: \texttt{boundary@marker} and \texttt{delimiter@marker} are strict no-answer-content readouts, while the headline \texttt{delimiter@16} setting uses a stronger answer-opening routing window. Because the half-open \texttt{delimiter@16} window begins at the anchor (Eq.~\ref{eq:window}), its first row predicts the anchor token while later rows are autoregressively conditioned on previously generated answer-region tokens; the decoded answer content is never supplied to the selector.

\subsection{Routing Agreement Decoding (RAD)}
\label{sec:method:rad}

RAD implements the distributional implication of Insight~2: around boundary and delimiter anchors, \S\ref{sec:analysis} shows the route neighborhood aligns with answer basins, so RAD reads routing from a specified anchor-window readout and selects the densest $K$-NN route basin.

For each rollout $i$, let $a_i$ be the delimiter anchor located by the dataset-specific answer-opening rule, the final answer-opening delimiter: for math and GPQA the boxed-answer marker, and for LiveCodeBench the code-output marker (occurrence rules and fallbacks in Appendix~\ref{app:fallback}). RAD forms the half-open anchor window
\begin{equation}
\mathcal{W}_i=\{a_i,\, a_i+1,\, \ldots,\, a_i+W-1\}, \qquad W=16.
\label{eq:window}
\end{equation}
The rollout representation is the per-token average over this window (each per-layer block of $z_i$ then sums to one; see \S\ref{sec:analysis:protocol}):
\begin{equation}
z_i\;=\;\frac{1}{|\mathcal{W}_i|}\sum_{t\in\mathcal{W}_i} R_i(t).
\label{eq:w16}
\end{equation}
Pairwise similarities among $\{z_i\}_{i=1}^N$ form a Weighted-Jaccard consensus matrix $S\in[0,1]^{N\times N}$ with $S_{ij} = \mathrm{WJ}(z_i, z_j)$ (Eq.~\ref{eq:wj}). For each rollout, let $\mathrm{KNN}_{k_{\mathrm{rad}}}(i)$ be its $k_{\mathrm{rad}}$ most similar other rollouts under $S$; the user-facing default is $k_{\mathrm{rad}}=10$ on the full $N=64$ pool. RAD assigns each rollout a local agreement density and selects the densest:
\begin{equation}
q_i \;=\; \sum_{j\in\mathrm{KNN}_{k_{\mathrm{rad}}}(i)} S_{ij},
\qquad
i^{\star} \;=\; \arg\max_i q_i.
\label{eq:knn}
\end{equation}
The selected answer string of $y_{i^{\star}}$ is read only after the routing-only selection has been made for evaluation, and the post-hoc purity diagnostic of \S\ref{sec:analysis:exp2} motivates the headline \texttt{delimiter@16} readout but is not part of the runtime selector.

The K-NN density objective is deliberately local. A global medoid can be pulled toward the space between several basins, especially when multiple answer candidates are common. In contrast, $q_i$ rewards being surrounded by many nearby rollouts and therefore selects a representative of one high-mass route basin. Thus RAD is an agreement selector: it returns a representative of a high-mass route basin, not an
independently verified solution. A dense route basin can still be wrong when many
rollouts share the same erroneous commitment.\textbf{Slogan: Majority counts answer strings; RAD aggregates routing agreement.}

\paragraph{Agentic instantiation.}
The same density selector carries to multi-step agentic rollouts (best-of-$N$ patch selection, \S\ref{sec:exp:results}), with two changes forced by the setting rather than tuned. First, agentic outputs open no \texttt{\textbackslash boxed\{}/code-fence, so we drop the delimiter anchor for the \emph{boundary} anchor (the thinking$\to$action transition), unioning each rollout's per-step boundary rows into one routing fingerprint. Second, $k_{\mathrm{rad}}$ tracks the pool size: the deployed $k_{\mathrm{rad}}{=}10$ is for $N{=}64$, and we use $k_{\mathrm{rad}}{=}4$ for the smaller $N{=}16$ agentic pools. There we read the binary form of the kernel (set membership over fired (layer,\,expert) pairs), which the kernel ablation in \S\ref{sec:exp:analysis} finds statistically indistinguishable from Weighted Jaccard; everything else---the $K$-NN density $q_i$ and the answer-string-free access boundary---is unchanged. This is the configuration evaluated on SWE-bench Verified (App.~\ref{app:swebench}).

\paragraph{Confidence fusion (RAD+DC).}
RAD+DC uses DeepConf's per-rollout confidence scalar as a complementary confidence signal, but not DeepConf's answer-string vote. In our deployed fusion, rollouts are first filtered by the DeepConf confidence score to the highest-confidence half (top-$50\%$ by cBW), and the RAD density is then computed within the retained subset (Appendix~\ref{app:rollout}). This distinction matters on code: the original DeepConf baseline is a confidence-weighted self-consistency vote over normalized answer strings, so it is degenerate for raw code outputs, which have no canonical answer class, whereas RAD+DC stays well-defined because it uses only the scalar confidence attached to each rollout plus routing density, and then outputs a single selected rollout directly.

%% file: sections/experiments.tex
\section{Experiments}
\label{sec:exp}

\subsection{Setup, baselines, and exact random envelopes}
\label{sec:exp:setup}

We evaluate all selectors on the same fixed $N{=}64$ rollout pool per problem (each method's candidate-set eligibility and coverage are detailed in Appendix~\ref{app:fallback}).

\paragraph{Datasets and models.}
The full pool contains 10 sparse MoE reasoning model configurations and 6 datasets: AIME24/25, BRUMO25, HMMT25, GPQA, and LiveCodeBench v5, for $4{,}850$ model--problem cells and $310{,}400$ generated runs. The models cover \textbf{gpt-oss} (20B / 120B), \textbf{Qwen3-30B-A3B}, and \textbf{Qwen3-Next-80B-A3B}, spanning top-4, top-8, and top-10 routing regimes (architectural details in Appendix~\ref{app:routing-capture}; gpt-oss Low/Medium/High are Harmony prompt-level reasoning-effort settings, Appendix~\ref{app:oss-reasoning-effort}). The same selector is then carried, unchanged in principle, to a separate \emph{agentic} regime---best-of-$N$ patch selection on SWE-bench Verified~\citep{jimenez2024swebench,openai2024swebenchverified} (\S\ref{sec:exp:results})---run with three multi-step coding-agent backbones under their native harnesses (gpt-oss 20B/120B under \emph{harmonyagent}, Qwen3.6-35B-A3B under \emph{mini-swe-agent}; per-model harness and configuration in Appendix~\ref{app:swebench}), where each rollout is a trajectory ending in a code patch rather than a single answer string.

\paragraph{Inference setup.}
Rollouts use temperature $0.7$, top-$p$ $0.9$, and maximum context length $32{,}768$. We capture top-$k$ expert ids and post-softmax routing weights from vLLM during generation (Appendix~\ref{app:routing-capture}).

\paragraph{Baselines.}
Baselines include random / Avg@64, textual self-consistency Majority, and the original DeepConf confidence-weighted vote~\citep{fu2026deepconf}. We distinguish \emph{DeepConf-vote}, which votes over normalized answer classes and is therefore reported only where string voting is well-posed, from the scalar DeepConf confidence score used inside RAD+DC.

\subsection{Results}
\label{sec:exp:results}

\paragraph{Pool-answer selection.}
RAD improves over single-rollout accuracy across model groups (substantially in the well-posed math+GPQA regime, more modestly on code) and, on the regime where string voting is well-posed (math + GPQA), is on par with textual Majority, modestly ahead in our pools (Table~\ref{tab:main}): well-posed pooled problem-weighted average Majority $73.6$ vs.\ RAD $73.9$ / RAD+DC $74.2$, with RAD winning four of five well-posed datasets and trailing only on GPQA. The paired natural-pool check ($n{=}3180$) gives RAD$-$Majority $+0.28$\,pp (bootstrap $95\%$ CI $[-0.50,+1.07]$; McNemar $p{=}0.52$) and RAD+DC$-$Majority $+0.57$\,pp ($[-0.22,+1.35]$; $p{=}0.18$). On code-generation, exact-string Majority and the original DeepConf-vote are degenerate on raw outputs: code strings are near-unique, so a string vote collapses to near-singletons (shown as ``--''); RAD and RAD+DC still produce direct pass@1 selections, improving over the Avg@64 random-rollout floor. Crucially, this edge does not depend on reading answer-opening tokens: the strict \texttt{delimiter@marker} readout, which conditions on \emph{no} answer-region tokens, already reaches RAD $71.1$ / RAD+DC $72.8$ on the well-posed pool (Avg@64 floor $66.3$, Majority $73.6$) and $65.3$ on code ($+4.0$ over Avg@64), with the headline \texttt{delimiter@16} adding only the remaining ${\sim}1$--$3$\,pp (Table~\ref{tab:marker-only}, Appendix~\ref{app:marker-only}). RAD is thus competitive where voting works and defined where voting does not.

\input{tables/big_table_main}

\paragraph{Agentic best-of-$N$: SWE-bench Verified.}
Code generation already strains string voting; the agentic setting removes the answer string entirely, and this is where the interface gap is widest. On SWE-bench Verified each candidate is a multi-step trajectory whose output is a code patch scored pass/fail by a hidden test suite, with no short answer to extract or normalize, so textual Majority and DeepConf-vote \emph{degenerate} exactly as on raw code but completely: patches are near-unique strings, so an exact-string vote collapses to singletons. We apply the agentic instantiation of RAD defined in \S\ref{sec:method:rad} (boundary anchor, per-step union, binary $K$-NN at $k{=}4$), anchoring at each model's thinking$\to$action transition: \texttt{</think>} for Qwen3.6-35B-A3B under the \emph{mini-swe-agent} harness and \texttt{<|start|>} for gpt-oss under \emph{harmonyagent} (per-model configuration in Appendix~\ref{app:swebench}). Over $N{=}16$ rollouts on $500$ problems per model, RAD improves best-of-$N$ selection by $+4.7/+4.8/+5.0$\,pp on the decidable subset of problems whose rollouts split, and by $+2.2$ to $+2.9$\,pp end-to-end over the random Avg@16 floor (Table~\ref{tab:swebench}; per-model setup and analysis in Appendix~\ref{app:swebench}). The one recipe---a single boundary anchor, one $k$, no per-model tuning---transfers across all three backbones, so the routing-density principle that keeps RAD level with textual Majority where voting is well-posed keeps selecting useful patches where voting has nothing to count.

\input{tables/table2_swebench}

\subsection{Analysis}
\label{sec:exp:analysis}

\paragraph{Robustness: kernel, layer depth, and reading cost.}
The selection signal is robust to the main implementation choices; the full tables and figures are in the appendix. \emph{(i)~Kernel.} The similarity kernel is not load-bearing: a within-cell paired ablation over $15{,}188$ problem cells finds binary and weighted Jaccard statistically indistinguishable, with a pooled $|\mathrm{WJ}-\mathrm{JA}|=0.22$\,pp (exact McNemar $p=0.25$) and power to exclude a pooled $\geq 1$\,pp gap (Appendix~\ref{app:wjja}). So the agreement signal lives in the routing geometry, not the weighting; this equivalence is also what licenses the binary-Jaccard fingerprint deployed in the agentic SWE-bench instantiation (\S\ref{sec:exp:results}). \emph{(ii)~Layer depth.} The signal is redundant across depth: leaving out any single layer-decile never significantly changes selector accuracy (max $|\Delta|\le 0.47$\,pp; $0/10$ leave-one-out drops survive Holm), and the shallowest tenth of layers alone matches the full network to within $\le 1.7$\,pp, so only shallow-layer routing need be captured, reducing routing-capture and storage cost, though not generation cost in this completed-rollout setting (Appendix~\ref{app:layer-depth}). \emph{(iii)~Reading cost.} How much routing to read is a cost--certainty tradeoff: early \emph{trajectory} anchors (\texttt{So}/\texttt{Now}) are read cheaply and near-universally, while late \emph{boundary}/\emph{delimiter} anchors cover fewer problems at low budget but give the largest uncertainty reduction once reached; the two cross at a per-rollout budget of $B\approx2$--$4$k, and the deployed \texttt{delimiter@16} readout sits at the high-budget, high-certainty end (Appendix~\ref{app:frontier}, Table~\ref{tab:frontier}).

\begin{figure}[t]
  \centering
  \begin{minipage}[c]{0.49\linewidth}\centering\includegraphics[width=\linewidth]{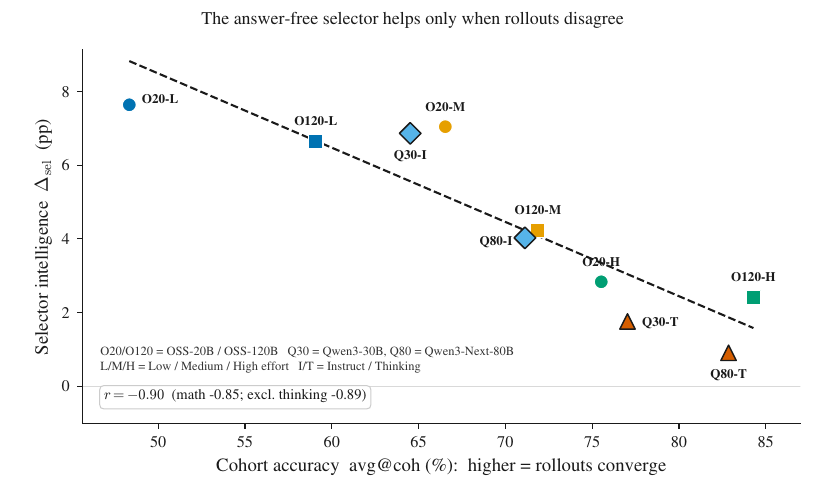}\end{minipage}\hfill
  \begin{minipage}[c]{0.49\linewidth}\centering\includegraphics[width=\linewidth]{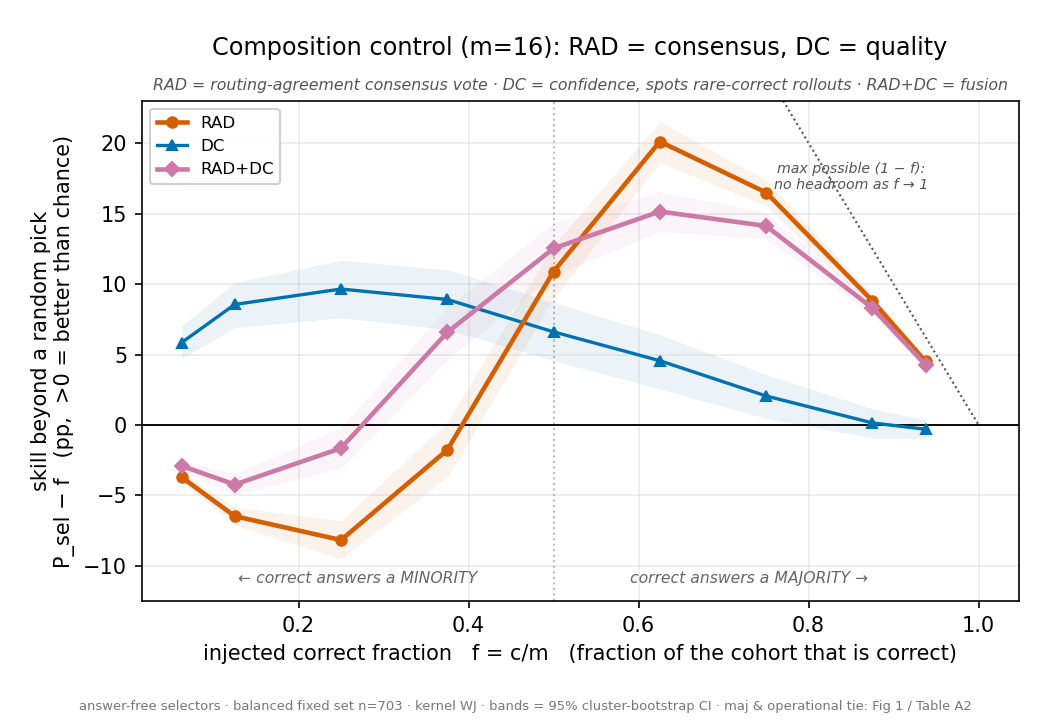}\end{minipage}
  \caption{\footnotesize \textbf{RAD is a soft consensus vote, not a verifier.} \textbf{(left)}~Selector intelligence $\Delta_{\mathrm{sel}}$ is largest when rollouts disagree and shrinks as cohorts converge (color $=$ effort, shape $=$ model; L/M/H, I/T $=$ Instruct/Thinking). \textbf{(right)}~In a composition control, RAD is below chance when correct rollouts are a minority and positive once they form a majority (the consensus signature) while DeepConf is the complementary rare-correct signal, so RAD+DC inherits both. Bands are $95\%$ cluster-bootstrap CIs; statistics and the delimiter-anchor replication are in the text (App.~\ref{app:effort},~\ref{app:rollout}).}
  \label{fig:cohort-agreement}\label{fig:rad-composition}
\end{figure}

\paragraph{What RAD selects: a soft consensus vote, not a verifier.}
Two controls (both panels of Figure~\ref{fig:cohort-agreement}) together show RAD is a soft consensus signal rather than a verifier: one for \emph{when} it helps (left), one for \emph{what} it selects (right).
\emph{When it helps.} RAD helps \emph{only when the parallel rollouts disagree}. Across the ten models, pure selector intelligence $\Delta_{\mathrm{sel}}$ falls as cohort accuracy avg@coh rises (Figure~\ref{fig:cohort-agreement}, left); the cross-model slope $r=-0.90$ is largely structural (dens and avg@coh are $99\%$ collinear), so we lead with the within-model, problem-level signal: within a model, a problem's advantage rises with its answer disagreement, partial correlation $+0.28$ ($p=2.9\times10^{-59}$) after partialling out difficulty (App.~\ref{app:effort}). Reasoning effort and the thinking-vs-instruct switch are knobs on this one axis: higher agreement shrinks the \emph{absolute} gain, but this is a \emph{ceiling} effect, not lost skill, and the operational lift $\Delta_{\mathrm{full}}$ stays $\sim+12$--$15$\,pp at every effort, with RAD at or above textual majority throughout. The agentic SWE-bench result is this same disagreement-gated mechanism at $N{=}16$: its lift is reported precisely on the decidable subset whose rollouts split into passes and fails---the only problems where selection can change the outcome---and concentrates there (Table~\ref{tab:swebench}).

\emph{What it selects.} A composition control isolates \emph{what} the selector selects: on a difficulty-balanced fixed set ($m{=}16$, $n{=}703$) we inject exactly $c$ correct rollouts and measure the pick's accuracy minus the random-pick baseline $f{=}c/m$ (Figure~\ref{fig:rad-composition}, right; boundary anchor, the deployed delimiter anchor replicates, App.~\ref{app:rollout}). RAD's lift over chance is a clean \emph{S-curve in $f$}: below chance when correct answers are a minority, crossing zero near $f{\approx}0.4$, and large-positive once they are a majority (App.~\ref{app:rollout}). RAD is therefore a soft, answer-string-free \emph{consensus} vote read from routing geometry. It tracks the most populated route basin, not correctness, and is \emph{below random} in the wrong-majority regime: it is not a verifier. The scalar DeepConf confidence signal is a second, complementary per-rollout \emph{quality} signal, positive precisely in the rare-correct regime where RAD fails; in the single-needle case ($c{=}1$) DeepConf finds the lone correct rollout above chance while RAD is below it, so a confidence-first selector is preferable at low pass rates (App.~\ref{app:rollout}). Consensus and confidence are empirically near-orthogonal: their per-rollout correlation is near zero (mean within-problem $r\approx0.03$ under both kernels; App.~\ref{app:rollout}). They also scale differently: RAD \emph{sharpens with the rollout count} ($+2.9$\,pp from $m{=}2$ to $64$, $95\%$ CI excludes $0$) while DeepConf does not. Fusing two non-redundant axes, RAD+DC inherits both and statistically \emph{ties} answer-aware self-consistency without reading a single answer string. The paired gap to Majority includes zero at both the deployed delimiter anchor ($+0.1$\,pp $[-0.8,+1.1]$) and the boundary anchor ($-0.4$\,pp $[-1.5,+0.6]$), at ${\sim}1$\,pp resolution (weighted Jaccard; the tie is fusion- and anchor-dependent, App.~\ref{app:rollout}). The selector's edge concentrates on mathematics (App.~\ref{app:rollout}).

\paragraph{Computational overhead.}
RAD adds ${\sim}7\%$ at rollout collection (top-$K$ routing capture in vLLM) and runs the selection step in real time on a single CPU thread. The full route cache used in this paper stores every per-token expert activation for offline analysis; at deployment, only the routing window around each delimiter anchor needs to be loaded into memory (Appendix~\ref{app:routing-capture}).

%% file: tables/big_table_main.tex
\providecommand{\best}[1]{\textbf{#1}}
\providecommand{\snd}[1]{\uline{#1}}
\providecommand{\mdl}[3]{\multirow{5}{*}{\shortstack[l]{#1\\[2pt]{\scriptsize #2}\\{\scriptsize #3}}}}
\definecolor{w16blue}{RGB}{225,244,248}
\definecolor{dgain}{RGB}{15,120,75}
\definecolor{dloss}{RGB}{176,0,32}
\begin{table}[t]
  \centering
  \footnotesize
  \setlength{\tabcolsep}{4.0pt}
  \renewcommand{\arraystretch}{1.15}
  \caption{\textbf{Pool-answer accuracy (\%) across four model families}, each pooled over its reasoning-effort / thinking--instruct variants; the sub-label gives the family's MoE routing width (routed-expert count and the top-$k$ experts activated per token; ``$+$shared'' = one always-on shared expert). \emph{Avg@64}: single random rollout; \emph{Cons@64}: self-consistency majority @64 (the textual \emph{Majority} baseline of the main text); \emph{DeepConf}: original confidence-weighted vote over normalized answer classes @64; \emph{RAD} is the headline \texttt{delimiter@16} setting ($K{=}10$, delimiter anchor \texttt{\textbackslash boxed\{}/code-fence, Weighted-Jaccard) and selects a rollout without reading any answer string; stricter no-answer-content readouts such as \texttt{delimiter@marker} and \texttt{boundary@marker} are analyzed in \S\ref{sec:analysis:exp2} and the appendices. \emph{RAD+DC} filters by the scalar DeepConf confidence score and then applies RAD density. Within each family, \textbf{bold} marks the best method per column; the green {\scriptsize\textcolor{dgain}{($+\Delta$)}} after each score is its gain over the Avg@64 random-rollout floor (per column). On Code (LCBv5) code outputs are near-unique strings, so exact-string Majority and the DeepConf-vote degenerate to near-singletons and are not meaningful (``--''), while RAD/RAD+DC still output a single selected rollout. \emph{Avg.}\ is the problem-weighted average over the five well-posed (math$+$GPQA) datasets. Well-posed pooled problem-weighted average over all ten configurations: Cons@64 73.6 vs.\ RAD 73.9 / RAD+DC 74.2 ($n{=}3180$: RAD$-$Cons $+0.28$ pp, $95\%$ bootstrap CI $[-0.50,+1.07]$, McNemar $p{=}0.52$; RAD+DC$-$Cons $+0.57$ pp, $[-0.22,+1.35]$, $p{=}0.18$). Strict \texttt{delimiter@marker} (reading no answer-region tokens) reaches RAD $71.1$ / RAD+DC $72.8$ well-posed and $65.3$ on code (Appendix~\ref{app:marker-only}).}
  \label{tab:main}
  \begin{tabular}{@{}ll rrrr r r @{\hspace{9pt}} r@{}}
    \toprule
    \multirow{2}{*}{\textbf{Model}} & \multirow{2}{*}{\textbf{Method}} & \multicolumn{4}{c}{\textbf{Math Reasoning}} & \multirow{2}{*}{\textbf{GPQA}} & \textbf{Avg.} & \multirow{2}{*}{\textbf{Code}} \\
    \cmidrule(lr){3-6}
     & & \textbf{AIME24} & \textbf{AIME25} & \textbf{BRUMO25} & \textbf{HMMT25} & & {\scriptsize math+GPQA} & \\
    \midrule
    \mdl{gpt-oss-20B}{32 experts, top-4}{Low/Med/High avg.} & Avg@64 & 65.0 & 59.5 & 64.2 & 41.2 & 60.0 & 59.0 & 50.8 \\
     & Cons@64 & 78.9\,{\scriptsize\textcolor{dgain}{(+13.9)}} & 73.3\,{\scriptsize\textcolor{dgain}{(+13.9)}} & \best{76.7}\,{\scriptsize\textcolor{dgain}{(+12.5)}} & 52.2\,{\scriptsize\textcolor{dgain}{(+11.0)}} & \best{67.0}\,{\scriptsize\textcolor{dgain}{(+7.0)}} & 68.2\,{\scriptsize\textcolor{dgain}{(+9.2)}} & -- \\
     & DeepConf & 78.9\,{\scriptsize\textcolor{dgain}{(+13.9)}} & 73.3\,{\scriptsize\textcolor{dgain}{(+13.9)}} & 74.4\,{\scriptsize\textcolor{dgain}{(+10.3)}} & 54.4\,{\scriptsize\textcolor{dgain}{(+13.2)}} & 66.8\,{\scriptsize\textcolor{dgain}{(+6.9)}} & 68.1\,{\scriptsize\textcolor{dgain}{(+9.1)}} & -- \\
     & \cellcolor{w16blue}\best{RAD} & \cellcolor{w16blue}\best{81.1}\,{\scriptsize\textcolor{dgain}{(+16.1)}} & \cellcolor{w16blue}\best{74.4}\,{\scriptsize\textcolor{dgain}{(+15.0)}} & \cellcolor{w16blue}\best{76.7}\,{\scriptsize\textcolor{dgain}{(+12.5)}} & \cellcolor{w16blue}60.0\,{\scriptsize\textcolor{dgain}{(+18.8)}} & \cellcolor{w16blue}65.3\,{\scriptsize\textcolor{dgain}{(+5.3)}} & \cellcolor{w16blue}68.2\,{\scriptsize\textcolor{dgain}{(+9.2)}} & \cellcolor{w16blue}60.5\,{\scriptsize\textcolor{dgain}{(+9.7)}} \\
     & \cellcolor{w16blue}\best{RAD+DC} & \cellcolor{w16blue}80.0\,{\scriptsize\textcolor{dgain}{(+15.0)}} & \cellcolor{w16blue}72.2\,{\scriptsize\textcolor{dgain}{(+12.8)}} & \cellcolor{w16blue}\best{76.7}\,{\scriptsize\textcolor{dgain}{(+12.5)}} & \cellcolor{w16blue}\best{62.2}\,{\scriptsize\textcolor{dgain}{(+21.0)}} & \cellcolor{w16blue}66.0\,{\scriptsize\textcolor{dgain}{(+6.0)}} & \cellcolor{w16blue}\best{68.6}\,{\scriptsize\textcolor{dgain}{(+9.5)}} & \cellcolor{w16blue}\best{61.3}\,{\scriptsize\textcolor{dgain}{(+10.5)}} \\
    \midrule
    \mdl{gpt-oss-120B}{128 experts, top-4}{Low/Med/High avg.} & Avg@64 & 73.4 & 71.7 & 72.5 & 55.9 & 70.6 & 69.7 & 70.6 \\
     & Cons@64 & 83.3\,{\scriptsize\textcolor{dgain}{(+9.9)}} & 81.1\,{\scriptsize\textcolor{dgain}{(+9.4)}} & \best{80.0}\,{\scriptsize\textcolor{dgain}{(+7.5)}} & 71.1\,{\scriptsize\textcolor{dgain}{(+15.2)}} & 74.1\,{\scriptsize\textcolor{dgain}{(+3.5)}} & 75.9\,{\scriptsize\textcolor{dgain}{(+6.1)}} & -- \\
     & DeepConf & \best{84.4}\,{\scriptsize\textcolor{dgain}{(+11.1)}} & 82.2\,{\scriptsize\textcolor{dgain}{(+10.5)}} & \best{80.0}\,{\scriptsize\textcolor{dgain}{(+7.5)}} & \best{73.3}\,{\scriptsize\textcolor{dgain}{(+17.4)}} & 73.9\,{\scriptsize\textcolor{dgain}{(+3.3)}} & \best{76.2}\,{\scriptsize\textcolor{dgain}{(+6.5)}} & -- \\
     & \cellcolor{w16blue}\best{RAD} & \cellcolor{w16blue}83.3\,{\scriptsize\textcolor{dgain}{(+9.9)}} & \cellcolor{w16blue}\best{83.3}\,{\scriptsize\textcolor{dgain}{(+11.6)}} & \cellcolor{w16blue}77.8\,{\scriptsize\textcolor{dgain}{(+5.2)}} & \cellcolor{w16blue}70.0\,{\scriptsize\textcolor{dgain}{(+14.1)}} & \cellcolor{w16blue}\best{74.4}\,{\scriptsize\textcolor{dgain}{(+3.8)}} & \cellcolor{w16blue}76.0\,{\scriptsize\textcolor{dgain}{(+6.3)}} & \cellcolor{w16blue}\best{77.0}\,{\scriptsize\textcolor{dgain}{(+6.5)}} \\
     & \cellcolor{w16blue}\best{RAD+DC} & \cellcolor{w16blue}82.2\,{\scriptsize\textcolor{dgain}{(+8.8)}} & \cellcolor{w16blue}\best{83.3}\,{\scriptsize\textcolor{dgain}{(+11.6)}} & \cellcolor{w16blue}\best{80.0}\,{\scriptsize\textcolor{dgain}{(+7.5)}} & \cellcolor{w16blue}71.1\,{\scriptsize\textcolor{dgain}{(+15.2)}} & \cellcolor{w16blue}73.7\,{\scriptsize\textcolor{dgain}{(+3.2)}} & \cellcolor{w16blue}75.8\,{\scriptsize\textcolor{dgain}{(+6.0)}} & \cellcolor{w16blue}76.4\,{\scriptsize\textcolor{dgain}{(+5.9)}} \\
    \midrule
    \mdl{Qwen3-30B-A3B}{128 experts, top-8}{Think/Instruct avg.} & Avg@64 & 79.7 & 70.7 & 77.4 & 51.6 & 68.6 & 69.1 & 58.2 \\
     & Cons@64 & 88.3\,{\scriptsize\textcolor{dgain}{(+8.6)}} & 81.7\,{\scriptsize\textcolor{dgain}{(+11.0)}} & 86.7\,{\scriptsize\textcolor{dgain}{(+9.3)}} & 58.3\,{\scriptsize\textcolor{dgain}{(+6.7)}} & 70.2\,{\scriptsize\textcolor{dgain}{(+1.6)}} & 73.4\,{\scriptsize\textcolor{dgain}{(+4.3)}} & -- \\
     & DeepConf & 88.3\,{\scriptsize\textcolor{dgain}{(+8.6)}} & 80.0\,{\scriptsize\textcolor{dgain}{(+9.3)}} & 86.7\,{\scriptsize\textcolor{dgain}{(+9.3)}} & 58.3\,{\scriptsize\textcolor{dgain}{(+6.7)}} & 70.5\,{\scriptsize\textcolor{dgain}{(+1.8)}} & 73.4\,{\scriptsize\textcolor{dgain}{(+4.3)}} & -- \\
     & \cellcolor{w16blue}\best{RAD} & \cellcolor{w16blue}\best{90.0}\,{\scriptsize\textcolor{dgain}{(+10.3)}} & \cellcolor{w16blue}\best{85.0}\,{\scriptsize\textcolor{dgain}{(+14.3)}} & \cellcolor{w16blue}88.3\,{\scriptsize\textcolor{dgain}{(+10.9)}} & \cellcolor{w16blue}\best{63.3}\,{\scriptsize\textcolor{dgain}{(+11.7)}} & \cellcolor{w16blue}\best{70.7}\,{\scriptsize\textcolor{dgain}{(+2.1)}} & \cellcolor{w16blue}\best{74.8}\,{\scriptsize\textcolor{dgain}{(+5.8)}} & \cellcolor{w16blue}59.0\,{\scriptsize\textcolor{dgain}{(+0.8)}} \\
     & \cellcolor{w16blue}\best{RAD+DC} & \cellcolor{w16blue}\best{90.0}\,{\scriptsize\textcolor{dgain}{(+10.3)}} & \cellcolor{w16blue}\best{85.0}\,{\scriptsize\textcolor{dgain}{(+14.3)}} & \cellcolor{w16blue}\best{90.0}\,{\scriptsize\textcolor{dgain}{(+12.6)}} & \cellcolor{w16blue}\best{63.3}\,{\scriptsize\textcolor{dgain}{(+11.7)}} & \cellcolor{w16blue}70.2\,{\scriptsize\textcolor{dgain}{(+1.6)}} & \cellcolor{w16blue}74.7\,{\scriptsize\textcolor{dgain}{(+5.6)}} & \cellcolor{w16blue}\best{59.9}\,{\scriptsize\textcolor{dgain}{(+1.7)}} \\
    \midrule
    \mdl{Qwen3-Next-80B}{512+shared experts, top-10}{Think/Instruct avg.} & Avg@64 & 83.9 & 75.2 & 80.6 & 57.0 & 66.5 & 69.4 & 66.5 \\
     & Cons@64 & \best{91.7}\,{\scriptsize\textcolor{dgain}{(+7.8)}} & 83.3\,{\scriptsize\textcolor{dgain}{(+8.2)}} & 81.7\,{\scriptsize\textcolor{dgain}{(+1.1)}} & 65.0\,{\scriptsize\textcolor{dgain}{(+8.0)}} & 77.0\,{\scriptsize\textcolor{dgain}{(+10.5)}} & 78.3\,{\scriptsize\textcolor{dgain}{(+8.9)}} & -- \\
     & DeepConf & \best{91.7}\,{\scriptsize\textcolor{dgain}{(+7.8)}} & \best{85.0}\,{\scriptsize\textcolor{dgain}{(+9.8)}} & 85.0\,{\scriptsize\textcolor{dgain}{(+4.4)}} & 65.0\,{\scriptsize\textcolor{dgain}{(+8.0)}} & \best{77.3}\,{\scriptsize\textcolor{dgain}{(+10.8)}} & 78.9\,{\scriptsize\textcolor{dgain}{(+9.5)}} & -- \\
     & \cellcolor{w16blue}\best{RAD} & \cellcolor{w16blue}90.0\,{\scriptsize\textcolor{dgain}{(+6.1)}} & \cellcolor{w16blue}83.3\,{\scriptsize\textcolor{dgain}{(+8.2)}} & \cellcolor{w16blue}90.0\,{\scriptsize\textcolor{dgain}{(+9.4)}} & \cellcolor{w16blue}66.7\,{\scriptsize\textcolor{dgain}{(+9.7)}} & \cellcolor{w16blue}75.5\,{\scriptsize\textcolor{dgain}{(+9.0)}} & \cellcolor{w16blue}78.1\,{\scriptsize\textcolor{dgain}{(+8.8)}} & \cellcolor{w16blue}67.1\,{\scriptsize\textcolor{dgain}{(+0.6)}} \\
     & \cellcolor{w16blue}\best{RAD+DC} & \cellcolor{w16blue}\best{91.7}\,{\scriptsize\textcolor{dgain}{(+7.8)}} & \cellcolor{w16blue}83.3\,{\scriptsize\textcolor{dgain}{(+8.2)}} & \cellcolor{w16blue}\best{91.7}\,{\scriptsize\textcolor{dgain}{(+11.1)}} & \cellcolor{w16blue}\best{71.7}\,{\scriptsize\textcolor{dgain}{(+14.7)}} & \cellcolor{w16blue}76.5\,{\scriptsize\textcolor{dgain}{(+10.0)}} & \cellcolor{w16blue}\best{79.6}\,{\scriptsize\textcolor{dgain}{(+10.2)}} & \cellcolor{w16blue}\best{69.2}\,{\scriptsize\textcolor{dgain}{(+2.7)}} \\
    \bottomrule
  \end{tabular}
\end{table}

%% file: tables/table2_swebench.tex
\providecommand{\best}[1]{\textbf{#1}}
\begin{table}[t]
  \centering
  \footnotesize
  \setlength{\tabcolsep}{5.0pt}
  \renewcommand{\arraystretch}{1.15}
  \caption{\textbf{Agentic best-of-$N$ selection on SWE-bench Verified} ($N{=}16$ rollouts $\times$ 500
    problems per model). Candidates are multi-step code patches scored pass/fail by the hidden test suite;
    there is no short answer to extract or normalize, so textual Majority and the DeepConf-vote \emph{degenerate} (exact-string votes over near-unique patches collapse to singletons).
    RAD reads routing at the \texttt{boundary@marker} anchor, the thinking$\to$action transition
    (\texttt{</think>} for Qwen3.6-35B-A3B, \texttt{<|start|>} for gpt-oss Harmony), unions the activated
    (layer, expert) pairs over the anchor rows, and selects the densest binary-Jaccard $K$-NN basin
    ($K{=}4$; binarization is a deliberate choice: it drops per-step counts, Appendix~\ref{app:swebench}). The baseline is random rollout
    selection (\emph{Avg@16}; its expectation equals the pool mean pass rate). \emph{Mixed sel@1}: accuracy
    on the decidable subset whose 16 rollouts split ($n_{\mathrm{mix}}$ problems, the only ones where
    selection can change the outcome); \emph{Avg@all}: end-to-end accuracy over all 500. RAD improves over
    random on every model. Per-model setup and analysis in Appendix~\ref{app:swebench}.}
  \label{tab:swebench}
  \begin{tabular}{@{}l c rrrr  rrr@{}}
    \toprule
    \multirow{2}{*}{\textbf{Model}} & \multirow{2}{*}{\textbf{Anchor}}
      & \multicolumn{4}{c}{\textbf{Mixed sel@1} (decidable subset)} & \multicolumn{3}{c}{\textbf{Avg@all} (500)} \\
    \cmidrule(lr){3-6}\cmidrule(l){7-9}
     & & $n_{\mathrm{mix}}$ & Avg@16 & RAD & $\Delta$ & Avg@16 & RAD & $\Delta$ \\
    \midrule
    Qwen3.6-35B-A3B & \texttt{</think>}  & 233 & 55.8 & \best{60.5} & $+4.7$ & 49.2 & \best{51.4} & $+2.2$ \\
    gpt-oss-120B    & \texttt{<|start|>} & 247 & 58.7 & \best{63.6} & $+4.8$ & 55.4 & \best{57.8} & $+2.4$ \\
    gpt-oss-20B     & \texttt{<|start|>} & 286 & 52.0 & \best{57.0} & $+5.0$ & 41.3 & \best{44.2} & $+2.9$ \\
    \bottomrule
  \end{tabular}
\end{table}

%% file: sections/limitations.tex
\section{Limitations}
\label{sec:limitations}

RAD is an answer-string-free arbitration rule, not a verifier: routing density estimates shared commitment, yet a dense route basin can still be wrong, exactly as textual Majority can follow a dense wrong consensus. It requires MoE-router access plus reliable anchors, so it does not apply to dense models or closed APIs.

RAD has a configurable access level. Strict readouts such as \texttt{boundary@marker} and \texttt{delimiter@marker} use routing rows that precede answer contents in the autoregressive conditioning order. The headline \texttt{delimiter@16} setting is not the strictest access setting: its half-open window $\mathcal{W}_i=[a_i,a_i+W)$ \emph{includes} the delimiter anchor (Eq.~\ref{eq:window}), so after the fixed delimiter later rows are conditioned on answer-opening tokens. This is an explicit signal-strength choice for the main accuracy table, not a hidden dependency: Appendix~\ref{app:marker-only} reports the strict \texttt{delimiter@marker} readout across all ten configurations and six datasets, and it recovers most of the lift while reading no answer content. The selector nonetheless reads no decoded answer strings, normalized answer classes, answer-token ids, correctness labels, execution results, or majority labels. Finally, the accuracy edge over textual Majority in the canonical-answer regime is small and not statistically significant in the natural well-posed pool ($n{=}3180$): RAD$-$Majority $+0.28$\,pp (bootstrap $95\%$ CI $[-0.50,+1.07]$, McNemar $p{=}0.52$) and RAD+DC$-$Majority $+0.57$\,pp ($[-0.22,+1.35]$, $p{=}0.18$), and it is concentrated outside GPQA; the more robust value is therefore the answer-string-free interface for code and lexically-fragmented outputs. The agentic SWE-bench result inherits the same not-a-verifier caveat and is smaller in scale---best-of-$16$ on one rollout pool per model, each under its own coding-agent harness---so it improves over the random floor but, as a consensus signal, still leaves a wide gap to the any-of-$16$ oracle ($69.8/75.8\%$ vs.\ the deployed $51.4/57.8\%$).

%% file: sections/conclusion.tex
\section{Conclusion}
\label{sec:conclusion}

The same token ids do not imply the same MoE routing state: anchor-conditioned routing leaves rollout-level structure that is strongest around boundary and delimiter readouts. RAD operationalizes this by representing each completed rollout by an anchor-window routing vector and selecting the densest center, without parsing, comparing, or voting over final answer strings. Across our 10-configuration 6-dataset MoE pool RAD is on par with textual Majority (with small positive but non-significant paired deltas) where string voting is well-posed (math and GPQA) and produces direct pass@1 selections on code-generation tasks where exact-string voting degenerates. On agentic SWE-bench Verified, it lifts best-of-16 selection over random where patches have no answer string to vote on. supporting the claim that routing agreement is a useful answer-basin signal with an analogous epistemic status: an agreement signal rather than a truth verifier. \textbf{Majority counts answer strings; RAD aggregates routing agreement.}

%% file: sections/appendix.tex
\subsection{Answer-string-free protocol details and access boundary}
\label{app:no-answer}

This appendix makes the answer-string-free access policy precise, as referenced from the protocol (\S\ref{sec:analysis:protocol}). The selector is allowed to read: token ids (used solely to locate anchors), routing tensors in the fixed anchor window, generated rollout length, and anchor occurrence counts. The selector is forbidden to read: normalized final-answer strings, textual majority labels, correctness labels, code-execution outputs, the surface-string contents of boxed-answer or code-output spans, or answer-token ids inside the window as features. RAD has configurable access levels: \texttt{boundary@marker} and \texttt{delimiter@marker} avoid answer-content-conditioned rows, while the headline \texttt{delimiter@16} readout uses a stronger answer-opening routing trace.

The routing index is prefix-causal. For a generated sequence $y_i$, the recorded row $R_i(t)$ is captured in the forward pass that predicts $y_{i,t}$ from $y_{i,<t}$; it is therefore aligned with $\texttt{tok\_token\_ids}[t]$, $\texttt{tok\_sampled\_logprob}[t]$, and $\texttt{tok\_topk\_logprobs}[t]$, but it does not condition on $y_{i,t}$ itself. A readout \texttt{type@W} is half-open, $\mathcal{W}_i=[a_i,a_i+W)$ with window width $W$ (Eq.~\ref{eq:window}). For a delimiter anchor of token length $m$, \texttt{delimiter@marker} reads the $m$ marker rows $R_i(a_i),\dots,R_i(a_i{+}m{-}1)$: the first predicts the opening delimiter token and conditions on neither the delimiter nor any answer content, and each later marker row conditions only on \emph{earlier marker tokens} (the fixed delimiter string), so \texttt{delimiter@marker} conditions on no answer-region content. And \texttt{delimiter@16} uses the same delimiter anchor with a
wider fixed window, $\mathcal{W}_i=[a_i,a_i+16)$. When $m<16$, the
additional rows are co-located with the first few answer-opening token
positions. 
The resulting distinction is an access--signal choice: 
\texttt{delimiter@marker} isolates the fixed marker rows, whereas
\texttt{delimiter@16} keeps the same anchor but widens the routing
window to answer-opening positions. Throughout the paper,
\emph{answer-string-free} denotes this interface-level property:
selection is made from routing geometry rather than from parsed,
normalized, compared, executed, or voted-over final answer strings.
Route density can correlate with answer basins because the model's
routing state near the committed answer carries basin information; this
is the mechanism under study.

\subsection{Strict marker-only readouts: routing agreement without answer-region tokens}
\label{app:marker-only}

A fair objection to the headline \texttt{delimiter@16} setting is that its window conditions on a few answer-opening tokens (an explicit signal-strength choice, not a dependency; \S\ref{sec:limitations} above). Table~\ref{tab:marker-only} isolates how much of RAD survives a \emph{strictly} answer-content-free readout: the marker-only \texttt{delimiter@marker}, which reads routing only at the answer-opening delimiter token(s) and conditions on no answer-region content. Across all ten configurations and six datasets it selects effectively. Pooled over the well-posed math$+$GPQA pool it reaches RAD $71.1$ / RAD$_{+\mathrm{DC}}$ $72.8$, well above the Avg@64 random floor ($66.3$) and close to textual Majority ($73.6$), while the wider \texttt{delimiter@16} adds only the remaining ${\sim}1$--$3$\,pp (the same ordering the $K$-NN purity diagnostic shows in Fig.~\ref{fig:evidence-a}). The routing-agreement signal is thus already present \emph{before} any answer token is read; the window is a signal-strength knob, not the source of the effect.

This is sharpest on code. A $16$-token window opened at a code fence cannot span a full program (it covers roughly the first line) so even \texttt{delimiter@16} reads no meaningful portion of the answer on LiveCodeBench, and the marker-only readout makes the point explicit: reading only the fence token(s), RAD still lifts pass@1 to $65.3$ ($+4.0$ over Avg@64) where exact-string Majority degenerates. The agentic setting is strict by construction: SWE-bench Verified (App.~\ref{app:swebench}) deploys \texttt{boundary@marker} (the thinking-to-action boundary, read once per reasoning step and unioned over the rollout's steps, with no answer-opening window) so its patch selection reads no answer content either.

In short, \texttt{delimiter@marker} and \texttt{boundary@marker} are the access-minimal configurations and the right default whenever no-answer-content routing is required; we report \texttt{delimiter@16} as the headline only because it is the strongest selector, at a pooled cost of ${\sim}1$--$3$\,pp over the marker-only readout. Weighting is not load-bearing here either: binary Jaccard is equivalent (App.~\ref{app:wjja}).

\input{tables/marker_only_table}

\subsection{MoE routing capture}
\label{app:routing-capture}

\paragraph{Capture point in the forward pass.}
Routing tensors are obtained from an internally MoE-capture-adapted vLLM v0.10.0 serving runtime with the \texttt{enable\_return\_routed\_experts} and \texttt{enable\_return\_routed\_experts\_weights} flags. A \texttt{RoutedExpertsCapturer} singleton intercepts the output of each MoE layer's \texttt{router.select\_experts(...)} call: this returns the top-$K$ expert ids and post-softmax weights \emph{after} the gate's softmax-then-top-$K$-renormalize step, so the recorded weights satisfy $\sum_{k} w_{i,\ell,t,k} \approx 1$ per token, layer, and rollout. Captured tensors are buffered on the GPU and copied to CPU staging buffers at request boundaries; we do not modify model weights, KV caches, or any compute path.

\paragraph{Storage schema.}
Per shard, captured routing for all $P$ problems and all $N{=}64$ rollouts is written under \texttt{token\_data\_*/} as four files: \texttt{moe\_meta.json} (schema version, $L$, $K$, total tokens, dtype dictionary), \texttt{moe\_row\_ptr.int64} (CSR row pointer mapping each run id to a token-index range), \texttt{moe\_topk\_expert\_ids.int16.b2} (flat $[T_{\text{shard}}\!\cdot\!L\!\cdot\!K]$, int16, blosc2 ZSTD\,+\,SHUFFLE\,+\,BYTEDELTA at compression level 5, 4\,MB chunks), and \texttt{moe\_topk\_expert\_weights.float16.b2} (same shape, float16, same compression). A parallel pair of files stores prompt-side routing. Compression yields ${\sim}1.8\times$ savings.

\paragraph{Token alignment.}
We strictly read token identities from the binary \texttt{tok\_token\_ids.int32.b2} column rather than re-tokenizing the generated text. Re-tokenization through \texttt{tokenizer.encode(generated\_text)} loses up to 7 trailing special-control tokens on OSS Harmony decoding (because the served text passes through \texttt{skip\_special\_tokens}), which causes a fixed off-by-$\Delta$ shift between text-derived token indices and the MoE routing tensor's index. All anchor-locating logic in this paper operates on \texttt{tok\_token\_ids}, so the post-marker window in RAD is exactly aligned with the routing tensor index.

\paragraph{Models and architectural diversity.}
The 10 model rollout pools span three structurally distinct sparse MoE designs (Table~\ref{tab:moe-arch}). \textbf{gpt-oss} (20B and 120B evaluated under Low/Medium/High prompt-level reasoning effort; Appendix~\ref{app:oss-reasoning-effort}) uses a Transformer MoE with top-$4$ routing and alternating full / locally-banded sparse attention plus grouped multi-query attention; the 20B variant has a small expert pool ($E{=}32$, $\sim$12.5\% activation rate per token) while the 120B variant has $E{=}128$ at $\sim$3.1\%. \textbf{Qwen3-30B-A3B} (Instruct/Thinking) is a standard Transformer MoE with $E{=}128$ and a relatively wide top-$8$ ($\sim$6.25\%) plus standard GQA attention. \textbf{Qwen3-Next-80B-A3B} (Instruct/Thinking) is a hybrid Gated DeltaNet / Gated Attention backbone with an ultra-sparse MoE: $E{=}512$ routed experts at top-$10$ ($\sim$1.95\%) plus one shared expert that contributes a constant per-token mass on top of the routed selection.

\paragraph{GPT-OSS reasoning-effort prompt protocol.}
\label{app:oss-reasoning-effort}
The gpt-oss Low/Medium/High conditions are not separate model checkpoints and do not modify model weights, MoE router parameters, expert pools, KV-cache behavior, quantization, temperature, or top-$p$. They are prompt-level conditions expressed in OpenAI's Harmony chat format. OpenAI's gpt-oss model card states that the models were trained to support three reasoning levels, \emph{low}, \emph{medium}, and \emph{high}, configured in the system prompt by inserting keywords such as \texttt{Reasoning:\ low}; increasing the reasoning level increases the model's average reasoning trace length~\citep{openai2025gptossmodelcard}. The Harmony documentation likewise places the reasoning-effort field in the system message, with \texttt{medium} as the default when no level is specified~\citep{openai2025harmony}.

Concretely, the OSS shards in this paper differ only in the following Harmony system-message line:
\begin{mdframed}[style=harmonystyle]
\begin{verbatim}
Reasoning: low
\end{verbatim}
\end{mdframed}
versus
\begin{mdframed}[style=harmonystyle]
\begin{verbatim}
Reasoning: medium
\end{verbatim}
\end{mdframed}
versus
\begin{mdframed}[style=harmonystyle]
\begin{verbatim}
Reasoning: high
\end{verbatim}
\end{mdframed}
All downstream capture code, routing hooks, sampling hyperparameters, and answer extraction logic are otherwise shared across the three conditions.

A minimal prompt-level example is:
\begin{mdframed}[style=harmonystyle]
\footnotesize
\begin{verbatim}
<|start|>system<|message|>You are ChatGPT, a large language model trained by OpenAI.
Knowledge cutoff: 2024-06
Current date: 2025-06-28

Reasoning: low

# Valid channels: analysis, commentary, final. Channel must be included for every message.<|end|>
<|start|>developer<|message|># Instructions
Solve the problem. Put the final answer in \boxed{}.<|end|>
<|start|>user<|message|>What is 17*23? Put the final answer in \boxed{}.<|end|>
<|start|>assistant
\end{verbatim}
\end{mdframed}
The Medium and High versions replace only the \texttt{Reasoning:\ low} line with \texttt{Reasoning:\ medium} or \texttt{Reasoning:\ high}. Harmony uses special control tokens such as \verb!<|start|>!, \verb!<|message|>!, and \verb!<|end|>! to delimit messages, and \verb!<|return|>! to terminate a completed assistant response. Assistant messages are further tagged by channels, including \verb!<|channel|>analysis! for the model's internal reasoning stream and \verb!<|channel|>final! for the user-facing answer~\citep{openai2025harmony}. In our OSS same-token analysis, the boundary anchor \verb!<|channel|>final<|message|>! is therefore a Harmony protocol boundary, not an answer-string feature. This is why Exp.~I can hold the measured token ids fixed at the final-channel boundary triplet while the experts that produce it still differ across Low, Medium, and High: the prompt-level reasoning-effort directive changes the hidden state entering the router before the identical protocol boundary is emitted.

\textbf{Implication for the routing representation.} The three regimes differ in how ``thick'' a per-token route vector is: gpt-oss-20B's small pool causes higher route-id collisions across rollouts, Qwen3-30B-A3B sits at the moderate sweet spot, and Qwen3-Next's ultra-sparse routing requires careful handling of the shared expert (we exclude it from the WJ histogram so it does not act as a constant noise floor that artificially inflates pairwise similarity). Despite these architectural differences, the same WJ-KNN agreement rule (Eq.~\ref{eq:knn}) drives RAD on all 10 pools.

\begin{table}[!ht]
  \centering
  \footnotesize
  \caption{\textbf{MoE architectural diversity in our pool.} The three families differ in expert pool size, top-$k$ routing density, attention block, and the presence of a shared expert; RAD's WJ-KNN agreement rule applies to all of them without modification.}
  \label{tab:moe-arch}
  \begin{tabular}{lrrrrl}
    \toprule
    Family & Total / Active & Layers & Experts & Top-$k$ & Attention \\
    \midrule
    gpt-oss-20B              & 21B / 3.6B  & 24 & 32  & 4  & alternating full / banded + GQA \\
    gpt-oss-120B             & 117B / 5.1B & 36 & 128 & 4  & alternating full / banded + GQA \\
    Qwen3-30B-A3B            & 30.5B / 3.3B & 48 & 128 & 8  & GQA (32 Q / 4 KV) \\
    Qwen3-Next-80B-A3B       & 80B / 3B     & 48 & 512 + 1 shared & 10 + shared & Gated DeltaNet + Gated Attention \\
    \bottomrule
  \end{tabular}
\end{table}

All weights are stored at float16 precision regardless of the model's inference quantization (e.g., mxfp4 on gpt-oss).

\paragraph{Capture cost.}
After the layer-id-cache and \texttt{clear\_buffer}-skip optimizations in the patched serving runtime, the routing capture path adds ${\sim}0.05$\,ms per generated token (${\sim}32$\,s for a $6{\times}10^5$ token batch), about a $7\%$ overhead relative to baseline generation on our hardware. Memory-mapped reads keep per-shard runtime memory under 8\,GB. Total disk usage for the 310{,}400-run, 4{,}850-cell pool is approximately 1.2\,TB, dominated by routing weights (the ids column compresses ${\sim}2{\times}$ better than the weights column).

\paragraph{Selection cost.}
RAD reduces each problem to a single Weighted-Jaccard $K$-NN density argmax over $N{=}64$ completed rollouts. With the sparse top-$K$ representation, each pairwise WJ is $O(L \cdot s)$ ($L \le 48$ MoE layers; $s \le \min(E, WK)$ the per-layer support after window-averaging, a small constant at the deployed $W{=}16$, larger on SWE-bench where it unions over boundary positions), so per-problem cost is dominated by the $64 \times 64$ similarity matrix and runs in tens of milliseconds on a single CPU thread; the full $4{,}850$-cell analysis pass completes in well under a minute end-to-end.

\subsection{K-NN basin purity protocol}
\label{app:knn-purity}

This appendix documents the exact computation of $A_{K_{\mathrm{nn}},r}$ (Eq.~\ref{eq:knn-purity}).

\begin{figure}[!ht]
  \centering
  \begin{subfigure}{0.49\linewidth}
    \centering
    \includegraphics[width=\linewidth]{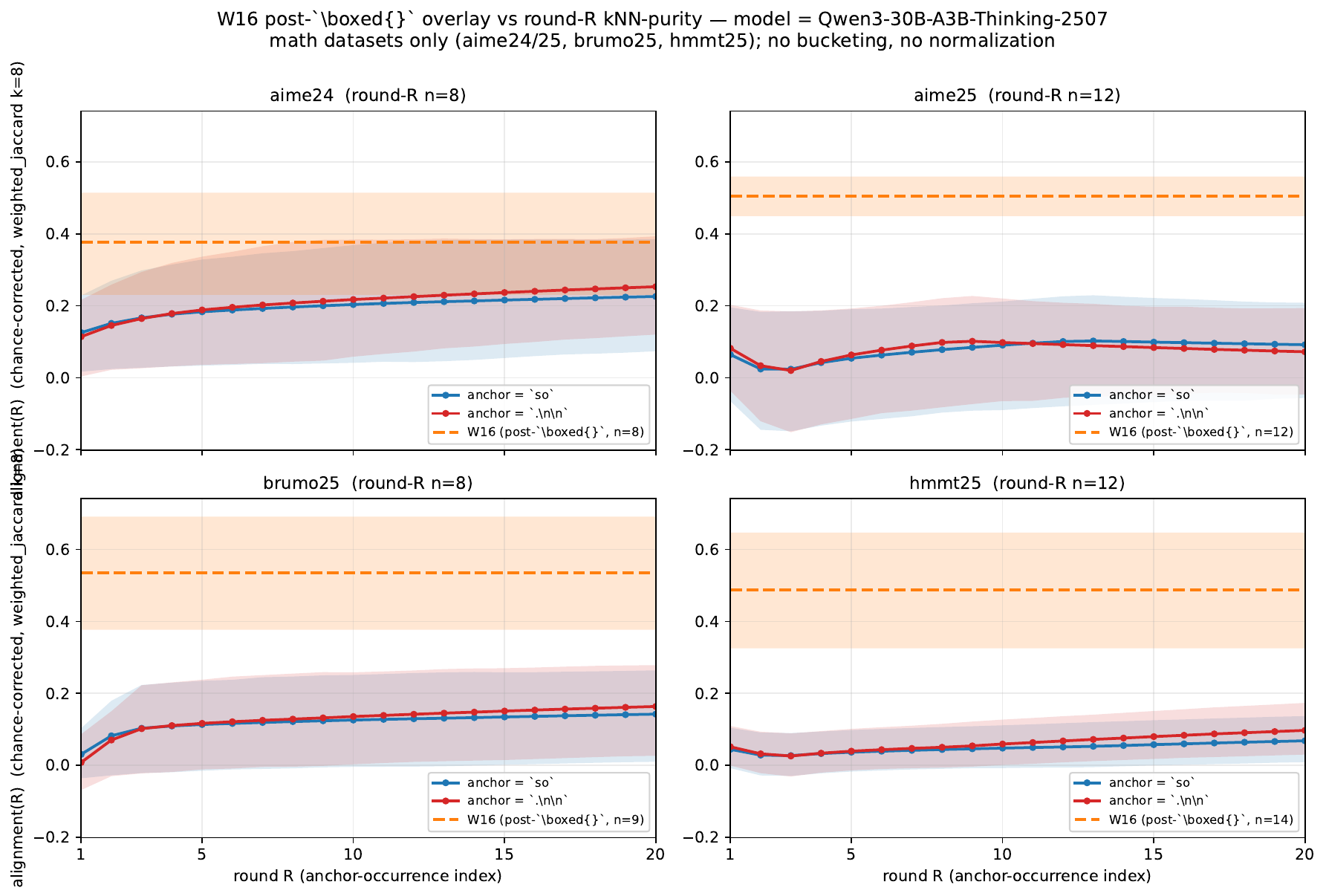}
    \caption{Qwen3-30B-A3B-Thinking-2507.}
    \label{fig:exp23-qwen30b}
  \end{subfigure}
  \hfill
  \begin{subfigure}{0.49\linewidth}
    \centering
    \includegraphics[width=\linewidth]{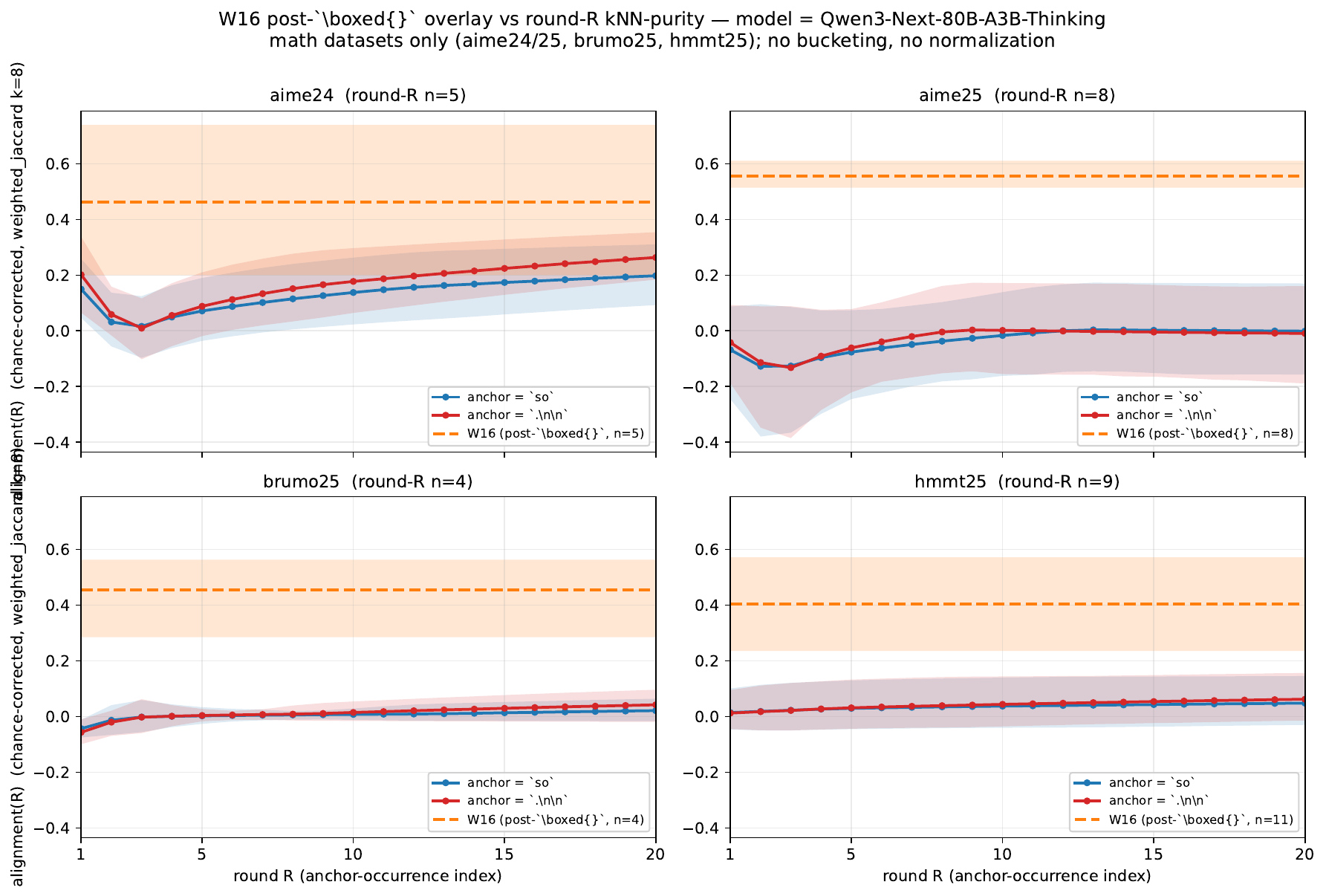}
    \caption{Qwen3-Next-80B-A3B-Thinking.}
    \label{fig:exp23-qwennext80b}
  \end{subfigure}
  \caption{\textbf{Anchor-round trajectory of $A_{K_{\mathrm{nn}},r}$ ($K_{\mathrm{nn}}{=}8$).} Lines per \emph{trajectory}-anchor surface form (shown in the in-panel legend under the heading ``anchor''): \texttt{so} in blue, \texttt{.\textbackslash n\textbackslash n} in red. The horizontal orange line (printed in-panel with the earlier label ``post-\textbackslash boxed'') is the \texttt{delimiter@16} RAD window. Shaded bands $95\%$ bootstrap CIs ($n_{\mathrm{boot}}{=}1000$). Two Qwen Thinking checkpoints; four math datasets per subfigure (AIME24, AIME25, BRUMO25, HMMT25). The early steep rise, middle stable plateau, and delimiter-window jump are visible across both models and all four datasets.}
  \label{fig:exp23}
\end{figure}

\paragraph{Cohort.}
Per problem $p$ we collect $N{=}64$ rollouts. A rollout is included iff (i) its boxed or code-output answer can be extracted and dataset-aware normalized, and (ii) the routing readouts required by the analysis are present: for the window-readout diagnostic (Fig.~\ref{fig:evidence-a}), valid delimiter-anchored windows; for the round-indexed trajectory (Fig.~\ref{fig:exp23}), at least one occurrence of the relevant trajectory-anchor token strictly before the delimiter marker. The effective neighbor count is $k_{\mathrm{eff}}=\min(K_{\mathrm{nn}}, N_{\mathrm{valid}}-1)$; we use $K_{\mathrm{nn}}{=}1$ for the window-readout basin-purity diagnostic (Fig.~\ref{fig:evidence-a}) and the purity--accuracy trajectory plot, and $K_{\mathrm{nn}}{=}8$ for the round-indexed anchor trajectory (Fig.~\ref{fig:exp23}). Problems with fewer than two valid rollouts or fewer than two distinct normalized answers are excluded; single-basin problems ($n_a{=}N$ for some $a$) yield $b{=}1$, so $A_{K_{\mathrm{nn}},r}$ is undefined and the problem is excluded. The $N{=}1{,}180$ basin-purity cells (aime25 138, brumo25 129, gpqa 913) are not fully independent (each problem is shared across the eight models) so the per-bar bootstrap (resampling cells) mildly underestimates the CI; a problem-clustered bootstrap widens the whiskers by ${\sim}5\%$ and leaves the window ordering unchanged.

\paragraph{Routing feature.}
The per-rollout routing vector at anchor round $r$ is the $L\!\cdot\!E$-dimensional sparse expert-weight histogram described in \S\ref{sec:analysis:protocol}, accumulated over the readout window and divided by the window's token count, i.e.\ the per-token mean of the routed expert mass. This per-token averaging matters when rollouts in the \emph{same} graph have different valid window lengths (from multi-token anchors, fallback, or truncation at the rollout's end) so a rollout with a shorter valid window is not penalized for carrying less total mass. (For a fixed window length it does not change the neighbor ranking, since Weighted Jaccard is invariant to a common rescaling of \emph{both} vectors, $\mathrm{WJ}(cu,cv){=}\mathrm{WJ}(u,v)$, and each readout's $K$-NN graph is built separately.) We use Weighted Jaccard throughout (for the selector, the round-indexed analyses, and the window-readout basin-purity diagnostic), having empirically verified that Weighted Jaccard and cosine produce qualitatively identical neighborhood rankings on this representation.

\paragraph{Basin label.}
The primary label space is the normalized final-answer string under dataset-aware normalization (AIME~$\to$~integer; GPQA~$\to$~first uppercase letter; LiveCodeBench~$\to$~raw string after canonicalization, which is near-unique across rollouts, usable for basin grouping but degenerate as a majority vote, which is why exact-string Majority is shown as ``--'' on code). Two diagnostic alternatives are reported in supplementary tables: binary correctness ($\{0,1\}$) and first-character of the normalized answer (coarsest, AIME-only).

\paragraph{Anchor-round trajectory (deferred from \S\ref{sec:analysis:exp2}).}
Figure~\ref{fig:exp23} traces $A_{K_{\mathrm{nn}},r}$ at successive trajectory-anchor occurrences and overlays the \texttt{delimiter@16} RAD window line, exposing the early-steep / middle-plateau / delimiter-jump three-regime structure that the main-text Figure~\ref{fig:evidence-a} only summarises in aggregate.

\begin{figure}[!ht]
  \centering
  \includegraphics[width=0.7\linewidth]{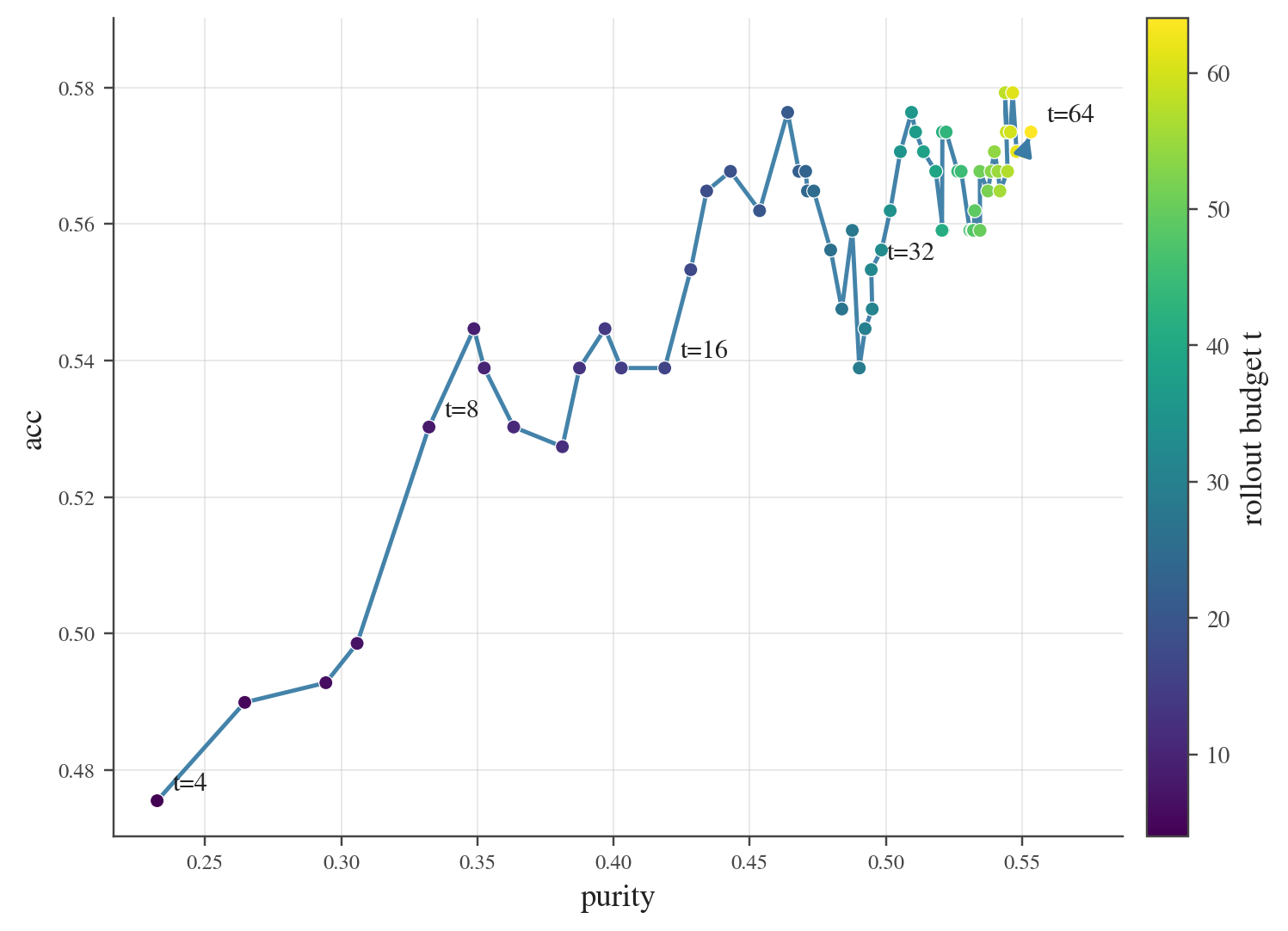}
  \caption{\textbf{Purity--accuracy trajectory parameterised by $t$.} Same balanced-panel data as Figure~\ref{fig:evidence-a} right ($347$ math problems, 8 MoE models, $K_{\mathrm{nn}}{=}1$ purity, KNN density-vote selector). Each scatter point is one $t \in \{4, \ldots, 64\}$; viridis color encodes $t$ from dark purple ($t{=}4$) to bright yellow ($t{=}64$); arrow shows direction of increasing $t$; anchor labels at $t \in \{4, 8, 16, 32, 64\}$. The curve walks northeast: each unit gain in $A_{K_{\mathrm{nn}}}$ corresponds to a roughly co-linear gain in selector accuracy.}
  \label{fig:purity-acc-phase}
\end{figure}

\paragraph{$A_{K_{\mathrm{nn}}}$--accuracy trajectory plot.}
Figure~\ref{fig:purity-acc-phase} presents the same per-$t$ data as the main-text Figure~\ref{fig:evidence-a} (right panel), reprojected from the dual-axis-vs-$t$ view into the (purity, accuracy) diagnostic plane: each scatter point is one rollout budget $t \in [4, 64]$, the viridis color encodes $t$, and the connecting curve traces the joint trajectory. The northeast walk of the trajectory is the alignment-vs-accuracy relationship made explicit: a unit increase in routing-neighbour basin purity corresponds to a measurable increase in selector accuracy on the same problems, ruling out the alternative that purity and accuracy are merely co-monotone in $t$ without coupling. Saturation lives at the upper-right end of the curve at $t \approx 32$--$64$.

\paragraph{Ground-truth vs.\ majority pseudo-label.}
The three-regime curve in Figure~\ref{fig:exp23} uses the textual majority-vote answer as a label-free pseudo-basin so that $A_{K_{\mathrm{nn}},r}$ does not require ground-truth access. We separately verify that using the held-out ground-truth answer as basin label yields qualitatively identical curves: the early-steep / middle-plateau / delimiter-jump shape and the relative gap between regimes are preserved within bootstrap CI on every model-dataset cell we checked. Majority-labelled $A_{K_{\mathrm{nn}},r}$ is therefore a faithful proxy for ground-truth-labelled $A_{K_{\mathrm{nn}},r}$ for the diagnostic claim of \S\ref{sec:analysis:exp2}.

\paragraph{Chance baseline forms.}
The implementation uses the leave-one-out form $b_{\mathrm{LOO}} = \sum_a n_a(n_a-1)/\big(N(N-1)\big)$ printed in Eq.~\ref{eq:knn-purity} for both the window-readout basin-purity diagnostic and the round-indexed analysis. The plug-in alternative $b_{\mathrm{PI}} = \sum_a (n_a/N)^2$ differs by $O(1/N)$ and yields qualitatively identical chance-corrected purity at $N{=}64$; we report the leave-one-out form throughout.

\paragraph{Cross-problem aggregation.}
Per-problem chance-corrected purity is $A_p = (\bar p_p - b_p)/(1 - b_p)$. The dataset-level value is the arithmetic mean across problems; a $95\%$ confidence interval is reported via problem-level percentile bootstrap with $n_{\mathrm{boot}}{=}2000$ resamples, resampling problems with replacement and reporting the $2.5$\textsuperscript{th}--$97.5$\textsuperscript{th} percentile of bootstrap means. For paired comparisons across two anchor rounds, the same bootstrap resamples the per-problem pair jointly and percentiles the resampled differences.

\subsection{Method details and fallbacks}
\label{app:fallback}

\paragraph{Anchor extraction.}
Trajectory, boundary, and delimiter anchors are located by binary \texttt{tok\_token\_ids} matching against a per-model token-id family (Appendix~\ref{app:routing-capture}). For each anchor, the family is built once per tokenizer by scanning the vocabulary and collecting every token id whose decoded surface form satisfies a regex defining the anchor: e.g., \texttt{so} uses \texttt{(?i)(\textasciicircum|[\textasciicircum A-Za-z])so(\$|[\textasciicircum A-Za-z])} so that \texttt{`\,so'}, \texttt{`\,So'}, and \texttt{`\textbackslash nSo'} all count as the same anchor occurrence regardless of how the BPE tokenizer happened to split surrounding context. The trajectory-anchor extractor matches the discourse-marker families \texttt{.\textbackslash n\textbackslash n}, \texttt{so}, \texttt{now}, \texttt{wait}, \texttt{.}, and \texttt{:}; the reported analyses use a \emph{single} such anchor per readout, by default \texttt{so} (with \texttt{now}, and the paragraph break \texttt{.\textbackslash n\textbackslash n} in Fig.~\ref{fig:exp23}, as the alternatives compared in App.~\ref{app:frontier}), never combined. Delimiter anchors are dataset-specific: \texttt{\textbackslash boxed\{} for math (AIME / BRUMO / HMMT) and GPQA, and the literal three-backtick code fence for LiveCodeBench. Boundary anchors use final-response transition markers such as the OSS Harmony triplet \texttt{<|channel|>\,final\,<|message|>} or end-of-thinking markers when present.

\paragraph{Delimiter-anchor location, eligibility, and baselines.}
For each rollout the controller locates the answer-opening delimiter, in order: (i) \texttt{\textbackslash boxed\{} on math/GPQA; (ii) the dataset-specific code-fence on LCB; (iii) the OSS boundary triplet on OSS rollouts. RAD scores a rollout only when its anchor (and the $16$-token window) is located; per problem this gives three cases: \textbf{(a)} all $N$ located (${\approx}75\%$ of well-posed problems), the candidate set is the full pool; \textbf{(b)} a strict subset located ($n_{\mathrm{cohort}}{<}N$, ${\approx}24\%$), the anchor-missing rollouts have no routing readout and are dropped from the candidate set, RAD selecting among the located ones; \textbf{(c)} fewer than two located (${\sim}0.8\%$), RAD abstains and the problem is scored \emph{incorrect}. Mean anchor coverage is ${\approx}93\%$ (per-model $0.73$--$1.00$). \emph{Baselines are handled the same way:} textual Majority and the DeepConf-vote act only on rollouts with a parseable answer (a rollout with none casts no vote), while the Avg@$N$ floor counts all $N$ unconditionally. RAD, Majority, and DeepConf are therefore scored on the \emph{same per-problem denominator} (cases (a)--(c) included), so RAD's anchor requirement is comparable to answer extraction, not an extra restriction. Restricting to anchor-located rollouts is a mild self-selection toward rollouts that reached the answer; it contributes the mechanical $\Delta_{\mathrm{anchor}}$ part of RAD's lift over Avg@64, with the genuine selection effect isolated as $\Delta_{\mathrm{sel}}$ (per-model in Table~\ref{tab:per-model}).

\paragraph{Weighted Jaccard computation.}
$\mathrm{WJ}(u,v) = \sum_e \min(u_e, v_e) / \sum_e \max(u_e, v_e)$ is computed on the sparse $L\!\cdot\!E$-dimensional histogram representation; only experts in the union of the two rollouts' (window-averaged) supports contribute, so the per-pair cost is $O(L \cdot s)$, $s$ the per-layer sparse support size, $s \le \min(E, WK)$, rather than $O(L \cdot E)$.

\subsection{Exact random baselines}
\label{app:exact}

\paragraph{Setup.}
For each shard, $P$ problems are sampled and each problem has $n_{\text{runs}}=64$ rollouts. Let $k_p$ be the number of correct rollouts on problem $p$ and $\hat p_p = k_p / n_{\text{runs}}$ its empirical correct rate. The random single-run selector picks one rollout uniformly at random per problem; we report its expected pass@1 $\mu_{\text{rand}} = (1/P)\sum_p \hat p_p$, its 95\% upper envelope $U_{95}^{\text{rand-1}}$, and the lift $\Delta = \text{Acc}_{\text{method}} - U_{95}^{\text{rand-1}}$.

\paragraph{Selector-variance Wald CI.}
Treating the corpus $\{\hat p_p\}_{p=1}^P$ as fixed and the random selector as the only source of randomness, the per-problem outcomes $X_p \sim \text{Bern}(\hat p_p)$ are independent and the random pass@1 has variance
\begin{equation}
\text{Var}\!\left[\tfrac{1}{P}\sum_p X_p\right] \;=\; \frac{1}{P^2}\sum_p \hat p_p(1-\hat p_p),
\label{eq:rand-var}
\end{equation}
which yields the Wald 95\% confidence interval
\begin{equation}
\text{CI}_{95} \;=\; \mu_{\text{rand}} \pm 1.96 \cdot \sqrt{\tfrac{1}{P^2}\sum_p \hat p_p(1-\hat p_p)}.
\label{eq:rand-ci}
\end{equation}
We use the upper bound $U_{95}^{\text{rand-1}}$ as the comparison line in the main text. This CI answers a fixed-corpus question (``how wide is the random selector's distribution on \emph{these} problems?'') rather than a cross-corpus question (``does the gap generalize to a new problem set?''); for the latter we additionally report problem-level paired bootstrap intervals on selected $\Delta$ values, but these are not the canonical envelope.

\subsection{Cost--certainty frontier and the routing window}
\label{app:frontier}
This appendix gives the full analysis behind the cost--certainty frontier of \S\ref{sec:exp:analysis}. We score a problem's rollouts by their local density in a $k$-NN graph ($K{=}10$) over the weighted-Jaccard similarity of the per-token-mean expert-activation histogram read in a short window at an anchor token, and return the densest rollout's correctness (\texttt{dens}). \emph{Trajectory} anchors are the recurring discourse markers \texttt{So} (default) and \texttt{Now}, each read as a single anchor (read early, near-universal; the recency-optimized variant \texttt{So$^*$} uses the latest in-budget \texttt{So}); \emph{Boundary} anchors are end-of-thinking or final-channel markers, while \emph{delimiter} anchors are fixed answer-opening delimiters such as \texttt{\textbackslash boxed\{\}} and code fences; both are read late and usually once. Per problem we compare \texttt{dens} against \texttt{avg@coh} (mean correctness over the anchor-found cohort) and the anchor-free \texttt{avg@full} ($=n_{\mathrm{correct}}/64$ over all rollouts), giving $\Delta_{\mathrm{anchor}}=\texttt{avg@coh}-\texttt{avg@full}$ (mechanical, kernel-independent), $\Delta_{\mathrm{sel}}=\texttt{dens}-\texttt{avg@coh}$ (pure selector intelligence), and $\Delta_{\mathrm{full}}=\Delta_{\mathrm{anchor}}+\Delta_{\mathrm{sel}}$. Significance is a Poisson-binomial exact test on the per-problem baseline rates (reported as $z$, one-sided $p$). We use the anchor-free \texttt{avg@full}; a deceptively common fallback ($\texttt{avg64}\times n_{\mathrm{graded}}/64$) zeroes out correct-but-anchor-absent rollouts and inflates the lift ${\approx}3.3\times$ at low coverage, which we avoid.

\paragraph{The frontier.} Table~\ref{tab:frontier} gives the per-budget $\Delta_{\mathrm{full}}$ and coverage for all five anchors. Trajectory anchors are usable from a small budget (coverage $\approx1$ throughout); boundary/delimiter anchors cover few problems until high budget but then dominate on accuracy, the two crossing at $B\approx2048$--$4096$. The families also differ in \emph{kind}: trajectory anchors carry $\Delta_{\mathrm{anchor}}\approx0$ (the markers are ubiquitous, so reaching one says nothing about correctness), so their lift is entirely selector intelligence, whereas boundary/delimiter anchors accrue $\Delta_{\mathrm{anchor}}$ up to $+3$--$5$\,pp because reaching the end-of-thinking point is itself a correctness signal, roughly half of their headline lift.

\begin{table}[!ht]
  \centering
  \footnotesize
  \caption{\textbf{Cost--certainty frontier (weighted Jaccard):} uncertainty reduction $\Delta_{\mathrm{full}}$ in pp (PB-exact $z$ in parentheses) vs.\ per-rollout token budget $B$, for the five anchors. $^\star$ marks $|z|\ge1.96$ (one-sided $p<0.025$). \texttt{cov} is the late-anchor coverage (fraction of problems with the boundary/delimiter anchor in budget); trajectory coverage is $\approx1$ from low $B$. Low-$B$ boundary/delimiter rows sit on a tiny short-trace subset (selection bias; flagged, not interpreted).}
  \label{tab:frontier}
  \begin{tabular}{rrrrrrr}
    \toprule
    $B$ & trajectory So & trajectory So$^*$ & trajectory Now & boundary & delimiter & cov(late) \\
    \midrule
    256   & $-0.38$ ($-0.8$) & $+1.58$ ($2.9$)$^\star$ & $-1.27$ ($-1.5$) & $-1.17$ ($-1.1$) & $-1.06$ ($-1.0$) & 0.19 \\
    1024  & $+0.70$ ($1.5$) & $+0.71$ ($1.4$) & $+0.06$ ($0.1$) & $+0.45$ ($0.7$) & $+0.87$ ($1.4$) & 0.45 \\
    4096  & $+1.71$ ($3.7$)$^\star$ & $+1.26$ ($2.6$)$^\star$ & $+0.39$ ($0.8$) & $+2.35$ ($4.4$)$^\star$ & $+3.69$ ($7.0$)$^\star$ & 0.68 \\
    16384 & $+2.77$ ($6.1$)$^\star$ & $+2.45$ ($5.1$)$^\star$ & $+2.37$ ($4.6$)$^\star$ & $+4.26$ ($9.0$)$^\star$ & $+5.83$ ($12.4$)$^\star$ & 0.92 \\
    32768 & $+3.45$ ($7.5$)$^\star$ & $+3.55$ ($7.4$)$^\star$ & $+3.98$ ($7.7$)$^\star$ & $+6.11$ ($13.2$)$^\star$ & $+7.65$ ($16.5$)$^\star$ & 0.98 \\
    \bottomrule
  \end{tabular}
\end{table}

\begin{figure}[!ht]
  \centering
  \includegraphics[width=0.78\linewidth]{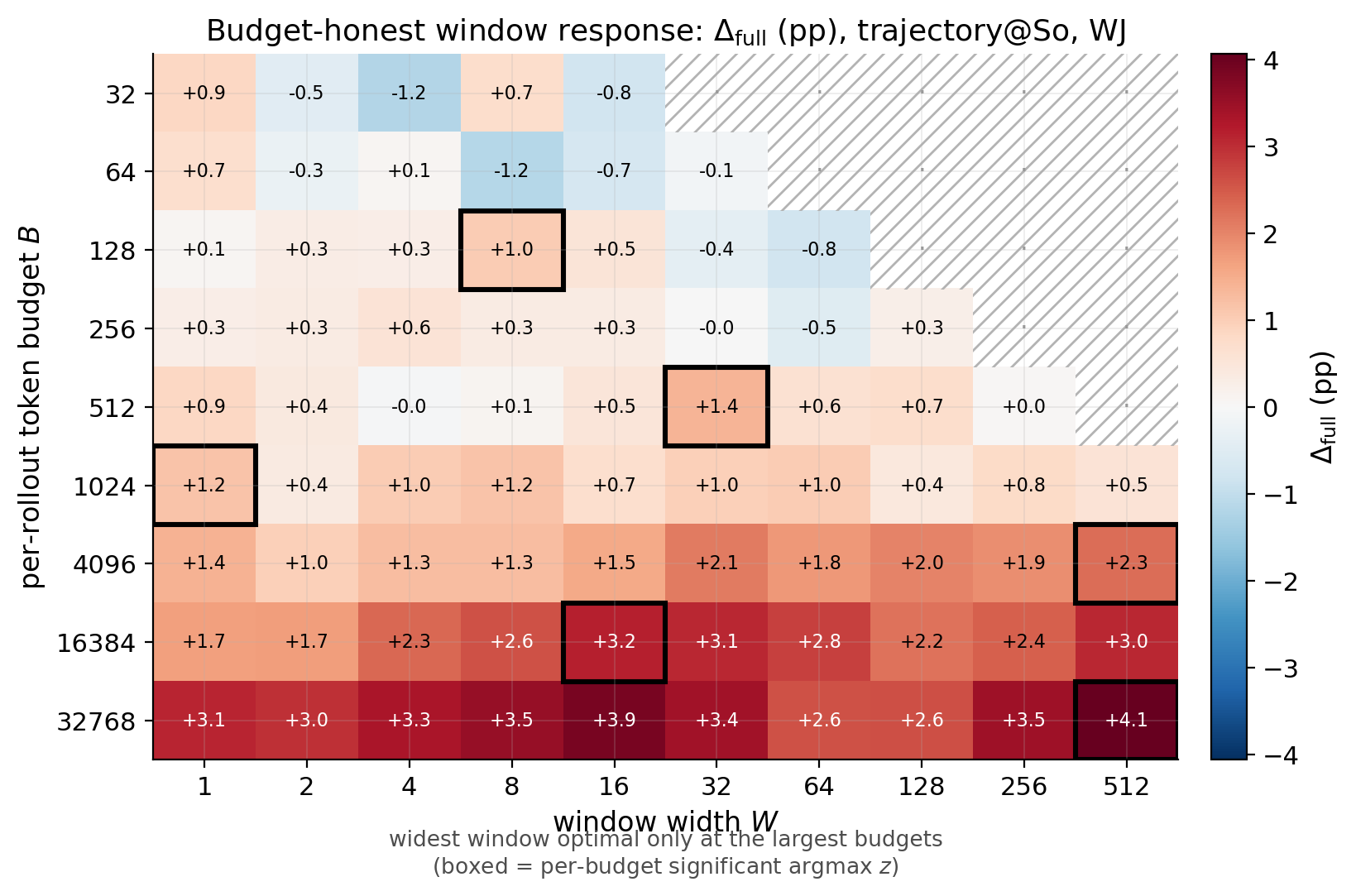}
  \caption{\textbf{Budget-honest window grid (trajectory \texttt{So}, weighted Jaccard).} $\Delta_{\mathrm{full}}$ (pp) for each (budget $B$, window width $W$); hatched cells cannot fit the budget. Boxed cells are the per-budget significant argmax ($z\ge2$). A small window is optimal at mid budgets; only the largest budget favours $W{=}512$.}
  \label{fig:window-grid}
\end{figure}

\paragraph{How much to read: the budget-honest window.} Holding the budget fixed, the routing window has a budget-honest optimum (Table~\ref{tab:window-grid}, Figure~\ref{fig:window-grid}): the whole window must fit in $[0,B)$ or the rollout is dropped. A small window ($W{\approx}8$--$16$) is best at mid budgets, and only the largest budget favours the widest fitting window ($W{=}512$, $+4.06$\,pp, $z{=}8.5$); the per-budget argmax is noisy between. A naive sweep that lets the window read past $B$ spuriously favours a mid window, a budget leak this rule removes. The deployed RAD reads a $16$-token window (\S\ref{sec:method:rad}), which sits in this flat near-optimal region at typical budgets.

\begin{table}[!ht]
  \centering
  \scriptsize
  \setlength{\tabcolsep}{3.5pt}
  \caption{\textbf{Budget-honest window grid (trajectory \texttt{So}, weighted Jaccard):} $\Delta_{\mathrm{full}}$ in pp for each (budget $B$, window width $W$); ``--'' marks windows that cannot fit the budget (rollout dropped).}
  \label{tab:window-grid}
  \begin{tabular}{r rrrrrrrrrr}
    \toprule
    $B \backslash W$ & 1 & 2 & 4 & 8 & 16 & 32 & 64 & 128 & 256 & 512 \\
    \midrule
    32    & $+0.9$ & $-0.5$ & $-1.2$ & $+0.7$ & $-0.8$ & -- & -- & -- & -- & -- \\
    64    & $+0.7$ & $-0.3$ & $+0.1$ & $-1.2$ & $-0.7$ & $-0.1$ & -- & -- & -- & -- \\
    128   & $+0.1$ & $+0.3$ & $+0.3$ & $+1.0$ & $+0.5$ & $-0.4$ & $-0.8$ & -- & -- & -- \\
    256   & $+0.3$ & $+0.3$ & $+0.6$ & $+0.3$ & $+0.3$ & $-0.0$ & $-0.5$ & $+0.3$ & -- & -- \\
    512   & $+0.9$ & $+0.4$ & $-0.0$ & $+0.1$ & $+0.5$ & $+1.4$ & $+0.6$ & $+0.7$ & $+0.0$ & -- \\
    1024  & $+1.2$ & $+0.4$ & $+1.0$ & $+1.2$ & $+0.7$ & $+1.0$ & $+1.0$ & $+0.4$ & $+0.8$ & $+0.5$ \\
    4096  & $+1.4$ & $+1.0$ & $+1.3$ & $+1.3$ & $+1.5$ & $+2.1$ & $+1.8$ & $+2.0$ & $+1.9$ & $+2.3$ \\
    16384 & $+1.7$ & $+1.7$ & $+2.3$ & $+2.6$ & $+3.2$ & $+3.1$ & $+2.8$ & $+2.2$ & $+2.4$ & $+3.0$ \\
    32768 & $+3.1$ & $+3.0$ & $+3.3$ & $+3.5$ & $+3.9$ & $+3.4$ & $+2.6$ & $+2.6$ & $+3.5$ & $+4.1$ \\
    \bottomrule
  \end{tabular}
\end{table}

\begin{figure}[!ht]
  \centering
  \includegraphics[width=\linewidth]{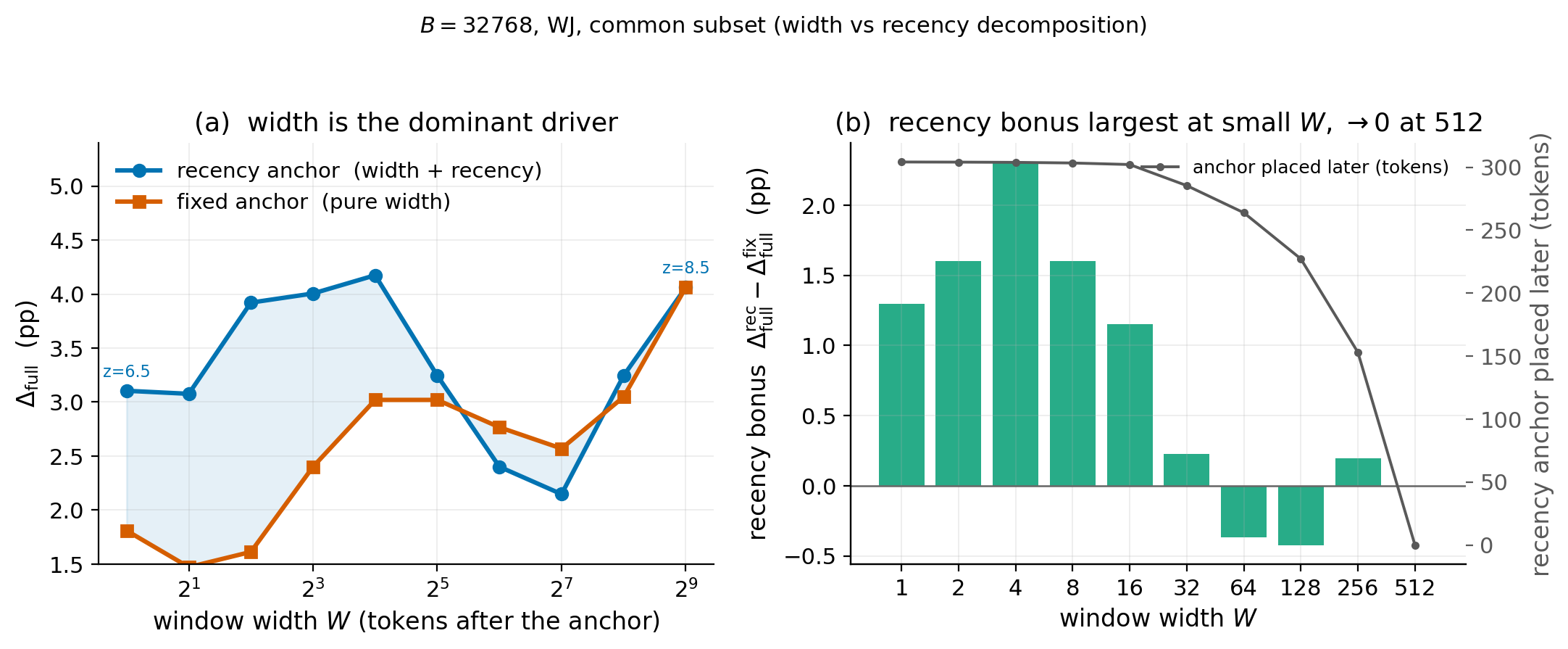}
  \caption{\textbf{Width vs.\ recency (trajectory \texttt{So}, $B{=}32768$, weighted Jaccard).} \emph{Left:} $\Delta_{\mathrm{full}}$ vs.\ window width for the \texttt{fixed} (pure-width) and \texttt{recency} (latest-anchor) strategies on a common subset. \emph{Right:} the recency bonus (\texttt{recency}$-$\texttt{fixed}) and the mean anchor depth per window. Width is the dominant driver ($+1.81\to+4.06$\,pp); the recency bonus is largest at small $W$ and vanishes by $W{=}512$.}
  \label{fig:width-recency}
\end{figure}

\paragraph{Width versus recency.} The window helps through two separable mechanisms (Table~\ref{tab:width-recency}, Figure~\ref{fig:width-recency}). On a common subset, fixing the anchor and widening the window alone (\emph{width}) lifts $\Delta_{\mathrm{full}}$ from $+1.81$ to $+4.06$\,pp, the dominant driver. Using the latest in-budget anchor adds a \emph{recency} bonus that is largest at small $W$ ($+1.29$ at $W{=}1$) and vanishes by $W{=}512$: small windows ride recency, wide windows ride width.

\begin{table}[!ht]
  \centering
  \footnotesize
  \caption{\textbf{Width vs.\ recency (trajectory \texttt{So}, $B{=}32768$, weighted Jaccard, common subset):} $\Delta_{\mathrm{full}}$ in pp ($z$). \texttt{fixed} holds the anchor and varies only the window (pure width); \texttt{recency} uses the latest in-budget anchor per window; \texttt{bonus}$=$\texttt{recency}$-$\texttt{fixed}.}
  \label{tab:width-recency}
  \begin{tabular}{rrrr}
    \toprule
    $W$ & fixed (width) & recency & bonus \\
    \midrule
    1   & $+1.81$ ($3.8$) & $+3.10$ ($6.5$) & $+1.29$ \\
    2   & $+1.47$ ($3.1$) & $+3.08$ ($6.4$) & $+1.60$ \\
    4   & $+1.61$ ($3.4$) & $+3.92$ ($8.2$) & $+2.31$ \\
    8   & $+2.40$ ($5.0$) & $+4.01$ ($8.3$) & $+1.60$ \\
    16  & $+3.02$ ($6.3$) & $+4.17$ ($8.7$) & $+1.15$ \\
    32  & $+3.02$ ($6.3$) & $+3.25$ ($6.8$) & $+0.23$ \\
    64  & $+2.77$ ($5.8$) & $+2.40$ ($5.0$) & $-0.37$ \\
    128 & $+2.57$ ($5.4$) & $+2.15$ ($4.5$) & $-0.42$ \\
    256 & $+3.05$ ($6.4$) & $+3.25$ ($6.8$) & $+0.20$ \\
    512 & $+4.06$ ($8.5$) & $+4.06$ ($8.5$) & $+0.00$ \\
    \bottomrule
  \end{tabular}
\end{table}

\paragraph{Per-model and coverage caveats.} The trajectory-anchor lift is positive on all 8 models (6/8 significant at $B{=}32768$) with $\Delta_{\mathrm{anchor}}\approx0$ throughout, whereas boundary/delimiter $\Delta_{\mathrm{anchor}}$ tracks reasoning effort / trace length (e.g.\ OSS-20B-High $+14.3$\,pp of its $+16.3$, OSS-120B-High $+6.8$ of $+8.2$), a property of \emph{whether the rollout finishes}, not of the selector. At low budget the boundary/delimiter cells sit on a tiny short-trace, low-effort subset (coverage $1$--$19\%$); those rows are selection-bias artifacts and are flagged, not interpreted. The kernel choice is within noise at the standard $W{=}16$ (Appendix~\ref{app:wjja}).

\paragraph{Domain dependence.} Selector intelligence is strongly domain-dependent (Table~\ref{tab:domain}): it orders math\,$>$\,science\,$>$\,code for the trajectory and boundary/delimiter anchors, and on code the trajectory (\texttt{So$^*$}) selector is null ($\Delta_{\mathrm{sel}}=+1.00$\,pp, $z{=}1.1$). The $+11$--$12$\,pp lift sometimes quoted for routing selectors exists only for boundary/delimiter$\times$math at high budget, and a large share of it is the mechanical $\Delta_{\mathrm{anchor}}$, not selector quality.

\begin{table}[!ht]
  \centering
  \footnotesize
  \caption{\textbf{Per-domain selector intelligence at $B{=}32768$ (weighted Jaccard):} $\Delta_{\mathrm{sel}}=\texttt{dens}-\texttt{avg@coh}$ (pure selector) and $\Delta_{\mathrm{full}}=\Delta_{\mathrm{anchor}}+\Delta_{\mathrm{sel}}$ in pp ($z$). The trajectory (\texttt{So$^*$}) selector is \emph{null} on code; the $+11$--$12$\,pp headline lives only in boundary/delimiter$\times$math, of which the mechanical $\Delta_{\mathrm{anchor}}$ is a large share.}
  \label{tab:domain}
  \begin{tabular}{lrrrrr}
    \toprule
    domain & \texttt{So$^*$} $\Delta_{\mathrm{sel}}$ & \texttt{So$^*$} $\Delta_{\mathrm{full}}$ & boundary $\Delta_{\mathrm{sel}}$ & boundary $\Delta_{\mathrm{full}}$ & boundary $\Delta_{\mathrm{anchor}}$ \\
    \midrule
    math    & $+6.88$ ($7.7$)$^\star$ & $+6.95$ ($7.7$)$^\star$ & $+6.17$ ($7.4$)$^\star$ & $+10.99$ ($12.0$)$^\star$ & $+4.82$ \\
    science & $+2.95$ ($4.1$)$^\star$ & $+3.09$ ($4.2$)$^\star$ & $+2.05$ ($2.8$)$^\star$ & $+3.74$ ($5.2$)$^\star$ & $+1.69$ \\
    code    & $+1.00$ ($1.1$) & $+1.10$ ($1.2$) & $+1.77$ ($2.3$)$^\star$ & $+5.46$ ($6.8$)$^\star$ & $+3.69$ \\
    \bottomrule
  \end{tabular}
\end{table}

\subsection{Agentic best-of-$N$: SWE-bench Verified}
\label{app:swebench}
This appendix details the agentic best-of-$N$ result reported with Table~\ref{tab:swebench}.

\paragraph{Setup.}
For each of $500$ SWE-bench Verified problems a coding agent samples $N{=}16$ independent rollouts ($8{,}000$ rollouts per model); each rollout emits a code patch whose correctness $y_i\in\{0,1\}$ is decided by the hidden test suite. The selector must pick one rollout \emph{without} executing tests, reading the reference patch, or inspecting $y_i$. It uses only the rollout's MoE routing, exactly as in the main pool. We anchor at the \emph{boundary} token marking the thinking$\to$action transition: \texttt{</think>} for Qwen3.6-35B-A3B (ChatML thinking template) and the action-turn opener \texttt{<|start|>} for the gpt-oss Harmony template (the multi-turn agentic action boundary). Both denote the same event (the switch from internal reasoning to emitting the patch) and occur exactly once per reasoning step (full coverage). For each rollout we take the union, over its boundary positions, of the activated (layer, expert) routing pairs into a binary fingerprint $S_i$, score each rollout by its mean binary Jaccard to its $k{=}4$ nearest neighbours, and select the densest center. The three backbones span distinct MoE widths (gpt-oss architectures in Appendix~\ref{app:routing-capture}; Qwen3.6-35B-A3B is a separate configuration with $40$ layers $\times$ $256$ routed experts). The baseline is random rollout selection (\emph{Avg@16}), whose expectation equals the per-problem mean pass rate with zero selector variance, matching the Avg@$N$ floor used for the main pool (Appendix~\ref{app:exact}).

\paragraph{Per-model agentic harness and configuration.}
Each backbone is run inside the multi-step coding-agent harness it is in-distribution for, rather than a single shared scaffold, so the routing recurs at a genuine action boundary in each model's native trajectory format. All three sample $N{=}16$ independent rollouts on the $500$-problem \textbf{SWE-bench Verified}~\citep{jimenez2024swebench,openai2024swebenchverified} split.
\emph{Qwen3.6-35B-A3B} runs under a \textbf{mini-swe-agent}-style loop~\citep{yang2024sweagent}: a linear message history in which the model alternates a \texttt{THOUGHT} block and a single \texttt{ACTION} \texttt{bash} command per step and terminates by issuing \texttt{submit}. Sampling is temperature $0.6$, top-$p$ $0.95$, up to $50$ steps, with a $60$\,s per-command execution timeout in a per-instance Docker environment; the candidate patch is recovered as \texttt{git diff} against the task base commit. The thinking$\to$action boundary is the ChatML \texttt{</think>} token that closes each step's reasoning span.
\emph{gpt-oss 20B and 120B} run under \textbf{harmonyagent}~\citep{mavrin2026harmony}, a from-scratch reconstruction of an in-distribution gpt-oss agent in the OpenAI \textbf{Harmony} response format~\citep{openai2025harmony}, with a Harmony tool set (\texttt{apply\_patch}, \texttt{repo\_browser.\{print\_tree,search,open\_file\}}, \texttt{container.exec}) and the analysis/final channels. Both sizes are run at a single fixed \texttt{Reasoning:\ high} reasoning-effort setting with a $131{,}072$-token context window and Harmony-default sampling (temperature $1.0$, top-$p$ $1.0$); unlike the main pool, which sweeps gpt-oss Low/Medium/High as separate configurations (Appendix~\ref{app:oss-reasoning-effort}), the agentic SWE-bench runs use only High for both models. Reasoning and tool calls are emitted in the \texttt{analysis} channel and each agentic action turn opens with the Harmony \texttt{<|start|>} control token, which is the boundary anchor. The candidate patch is recovered as \texttt{git add -A \&\& git diff --cached}.
Because the two harnesses extract patches under different conventions (base-commit \texttt{diff} vs.\ staged \texttt{diff --cached}, with different truncation and cleanup), \emph{absolute} resolve rates are not strictly comparable across the Qwen and gpt-oss families. This does not affect RAD: selection is always made \emph{within} one model's $16$-rollout cohort under a single fixed harness, where all candidates share the same extraction convention, so each model's reported lift over its own Avg@16 floor is invariant to the cross-family extraction mismatch.

\paragraph{Metrics.}
\emph{Mixed sel@1} is accuracy on the \emph{decidable} subset: problems whose 16 rollouts are split between passing and failing, the only problems on which a selector can change the outcome ($n_{\mathrm{mix}}=233/247/286$ for Qwen3.6-35B-A3B / gpt-oss-120B / gpt-oss-20B). \emph{Avg@all} is end-to-end accuracy over all 500 problems. Because roughly half of all problems are unanimous (all rollouts pass or all fail) and are therefore inert under any selector, the end-to-end gain is close to half the decidable-subset gain: the $+4.7/+4.8/+5.0$\,pp on mixed sel@1 dilute to $+2.2/+2.4/+2.9$\,pp on Avg@all (Table~\ref{tab:swebench}).

\paragraph{Cross-architecture portability.}
The identical recipe (boundary anchor, pair-level binary Jaccard, $k{=}4$ density) yields $+4$ to $+5$\,pp on the decidable subset of all three models, with no per-model tuning and no access to answer strings. The three backbones differ substantially in routing width and chat template, so the agreement signal at the action boundary is a portable property of MoE reasoning rather than an artefact of one architecture.

\paragraph{Why a binary fingerprint.}
We record only \emph{which} (layer, expert) pairs fire at the boundary, not how often, so binarizing to set membership drops the per-pair \emph{count} component (which grows with trajectory length) and keeps the portable ``which experts were recruited'' signal. The union over a rollout's per-step boundaries still leaves the fingerprint cardinality $|S_i|$ scaling with step count, but this length confound is inert at the selector level: \emph{within} each problem's rollout cohort (the only set the $k$-NN ranks), $|S_i|$ is uncorrelated with correctness and a step-count-only selector gives no lift, so RAD's $+4.7/+4.8/+5.0$\,pp is routing agreement, not trajectory length (Table~\ref{tab:swe-length}).

\paragraph{Why a single anchor.}
The signal is concentrated at the single action-start boundary. Pooling routing rows from the boundary together with other token positions mixes distinct decision phases and dilutes the agreement signal; reading each reasoning step at its lone boundary token keeps the comparison phase-aligned. This is the agentic counterpart of the marker-only readouts analysed for the main pool in \S\ref{sec:exp:analysis}.

\begin{table}[t]
  \centering\footnotesize
  \setlength{\tabcolsep}{6pt}\renewcommand{\arraystretch}{1.15}
  \caption{\textbf{The boundary-fingerprint length confound is inert at the selector level} (SWE-bench, $N{=}16$ rollouts; pooled correlations over all rollouts, within-problem correlation and lifts on the decidable subset, in pp over Avg@16). Per-rollout step counts vary widely (e.g.\ Qwen3.6 $1$--$166$ thinking$\to$action boundaries) and fingerprint cardinality $|S_i|$ scales with them. On Qwen3.6 $|S_i|$ is pooled-anticorrelated with correctness ($-0.23$), but this is an across-problem difficulty effect (harder problems run longer and fail more): it \emph{vanishes within problem} ($|r|{\le}0.06$ on all three), and the within-problem cohort is the only set the $k$-NN ranks. Accordingly the best step-count-only selector gives no positive lift, whereas RAD lifts $+4.7/+4.8/+5.0$\,pp. The agentic gain is routing agreement, not trajectory length.}
  \label{tab:swe-length}
  \begin{tabular}{@{}l c c c c c@{}}
    \toprule
    \textbf{Model} & $r(|S_i|,\mathrm{step})$ & \multicolumn{2}{c}{$r(|S_i|,\mathrm{correct})$} & step-count & RAD \\
    \cmidrule(lr){3-4}
     & (pooled) & pooled & within-prob. & selector & lift \\
    \midrule
    Qwen3.6-35B-A3B & $+0.60$ & $-0.23$ & $-0.00$ & $-3.9$ & $\mathbf{+4.7}$ \\
    gpt-oss-120B & $+0.23$ & $+0.01$ & $+0.03$ & $+0.8$ & $\mathbf{+4.8}$ \\
    gpt-oss-20B & $+0.15$ & $+0.04$ & $+0.06$ & $+0.5$ & $\mathbf{+5.0}$ \\
    \bottomrule
  \end{tabular}
\end{table}

\subsection{Cohort agreement, reasoning effort, and thinking vs.\ instruct}
\label{app:effort}
This appendix gives the full analysis behind the cohort-agreement result of \S\ref{sec:exp:analysis}, using the lift decomposition of Appendix~\ref{app:frontier} (anchor-free $\texttt{avg@full}=n_{\mathrm{correct}}/64$; $\texttt{avg@coh}$ the anchor-found cohort mean; $\Delta_{\mathrm{anchor}}=\texttt{avg@coh}-\texttt{avg@full}$, $\Delta_{\mathrm{sel}}=\texttt{dens}-\texttt{avg@coh}$ the selector intelligence vs.\ a random within-cohort pick, $\Delta_{\mathrm{full}}=\Delta_{\mathrm{anchor}}+\Delta_{\mathrm{sel}}$). The pool spans ten models (gpt-oss 20B/120B at Low/Medium/High reasoning effort and Qwen3-30B-A3B / Qwen3-Next-80B-A3B in Instruct and Thinking variants) over the math, GPQA, and well-posed (math$+$GPQA) pools, at the deployed setting (delimiter anchor, 16-token window, $K{=}10$, weighted Jaccard; the kernel is a null axis). Significance is the Poisson-binomial exact macro test ($\Delta_{\mathrm{sel}}/\Delta_{\mathrm{full}}$, the random-within-cohort-pick null), exact McNemar (paired contrasts), and a per-size paired bootstrap (effort trend); $^{*}/^{**}/^{***}$ denote $p<0.05/0.01/0.001$.

\begin{figure}[!ht]\centering
  \includegraphics[width=0.7\linewidth]{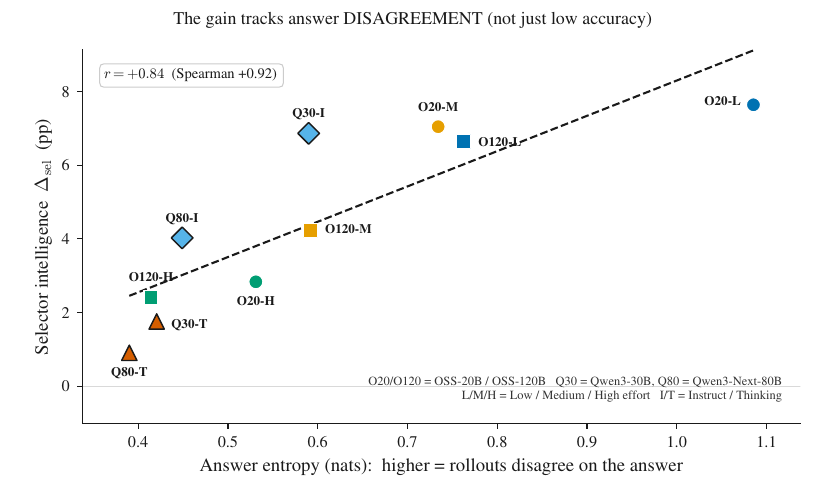}
  \caption{\textbf{The selector's gain tracks answer \emph{disagreement}, not just low accuracy.} Per-model selector intelligence $\Delta_{\mathrm{sel}}$ vs.\ full-pool answer entropy (well-posed; weighted Jaccard); cross-model $r=+0.84$. The difficulty-controlled problem-level test is Table~\ref{tab:diversity}.}
  \label{fig:diversity}
\end{figure}

\paragraph{Cohort agreement (the law).} The selector's gain concentrates on problems whose rollouts disagree. The cross-model correlation between $\Delta_{\mathrm{sel}}$ and the cohort baseline avg@coh is $r=-0.90$ (Figure~\ref{fig:cohort-agreement}), but this is \emph{largely structural}: dens and avg@coh are $99\%$ collinear ($r=0.993$), so $\Delta_{\mathrm{sel}}$ is a residual bounded by the $1-\texttt{avg@coh}$ headroom and a constant-headroom-fraction selector already yields slope $\approx-1$. The more direct test not relying on the cross-model headroom is at the problem level, within model: a problem's selector advantage (RAD-correct minus its cohort pass rate) rises with its answer entropy, simple correlation $+0.19$ ($p=1.3\times10^{-27}$, $n=3152$) and \emph{partial} correlation $+0.28$ ($p=2.9\times10^{-59}$) after partialling out difficulty (Table~\ref{tab:diversity}), which cannot be a ceiling/part-whole artifact. The cross-model answer-entropy twin ($r=+0.84$) is shown directly in Figure~\ref{fig:diversity}.

\begin{table}[!ht]\centering\footnotesize
\caption{\textbf{Problem-level: the selector helps more on higher-disagreement problems (weighted Jaccard).} Within each model, per-problem selector advantage (RAD-correct $-$ cohort pass rate) vs.\ answer entropy (well-posed, Pearson). Pooled $=$ within-model demeaned; the last row additionally partials out per-problem difficulty (cohort pass rate). The positive partial correlation shows that the entropy--advantage association is not explained away by cohort pass rate.}
\label{tab:diversity}
\begin{tabular}{@{}l r r@{}}\toprule
Model & $n$ & corr(entropy, advantage) \\\midrule
OSS-120B-High & 316 & +0.11 \\
OSS-120B-Low & 318 & +0.21 \\
OSS-120B-Medium & 318 & +0.16 \\
OSS-20B-High & 304 & +0.18 \\
OSS-20B-Low & 318 & +0.14 \\
OSS-20B-Medium & 318 & +0.31 \\
Qwen3-30B-A3B-Instruct-2507 & 318 & +0.31 \\
Qwen3-30B-A3B-Thinking-2507 & 316 & +0.15 \\
Qwen3-Next-80B-A3B-Instruct & 318 & +0.18 \\
Qwen3-Next-80B-A3B-Thinking & 308 & +0.09 \\
\midrule
\textbf{Pooled (within-model demeaned)} & 3152 & \textbf{+0.19} \;($p{=}1.3\times10^{-27}$) \\
\textbf{\;\;+ difficulty partialled out} & 3152 & \textbf{+0.28} \;($p{=}2.9\times10^{-59}$) \\
\bottomrule\end{tabular}\end{table}

\paragraph{Reasoning effort: a ceiling effect, not lost skill.} As OSS effort rises Low$\to$Medium$\to$High the cohort converges (avg@coh $42\to70\to92\%$), so the absolute $\Delta_{\mathrm{sel}}$ falls $+12.2\to+11.2\to+3.6$\,pp (Table~\ref{tab:effort-decomp}; paired-bootstrap Low$-$High $+9.3$\,pp $[+2.8,+15.6]$ $p=0.003$ for 20B, $+9.2$ $[+4.0,+14.3]$ $p<0.001$ for 120B, Table~\ref{tab:effort-trend}). Normalised by headroom the selector captures an \emph{equal-or-larger} fraction at higher effort ($21\to39\to43\%$), so this is a ceiling effect, not lost skill. The operational lift $\Delta_{\mathrm{full}}$ stays $\sim+12$--$15$\,pp at every effort, but its source flips from genuine selection (Low: $\Delta_{\mathrm{anchor}}\approx0$) to the mechanical late-anchor filter (High: $\Delta_{\mathrm{anchor}}\approx+11$, as high-effort runs less reliably emit an in-window \texttt{\textbackslash boxed}). RAD matches or edges textual majority at every effort (dens$-$maj$@64=+0.8/0/+3.1$\,pp).

\begin{table}[!ht]\centering\small
\caption{\textbf{RAD lift vs.\ OSS reasoning effort} (math, 20B+120B pooled, weighted Jaccard). $\Delta_{\mathrm{anchor}}$ = late-anchor self-selection; $\Delta_{\mathrm{sel}}$ = genuine selector intelligence (PB-exact); $\Delta_{\mathrm{full}}{=}\Delta_{\mathrm{anchor}}{+}\Delta_{\mathrm{sel}}$. As effort rises the cohort converges (avg@coh 42$\to$92\%): the absolute $\Delta_{\mathrm{sel}}$ shrinks but per-headroom it is preserved, and the lift's source flips from selection to the mechanical anchor filter.}
\label{tab:effort-decomp}
\begin{tabular}{@{}l rr rrr r rr@{}}\toprule
Effort & avg@64 & Cons@64 & $\Delta_{\mathrm{anc}}$ & $\Delta_{\mathrm{sel}}$ & $\Delta_{\mathrm{full}}$ & $\Delta_{\mathrm{sel}}$/HR & RAD & RAD+DC \\\midrule
Low & 41.9 & 53.3 & +0.1 & +12.2$^{***}$ & +12.2 & 21.0\% & 54.2 & 54.2 \\
Medium & 69.9 & 82.5 & +1.4 & +11.2$^{***}$ & +12.6 & 39.1\% & 82.5 & 84.2 \\
High & 80.6 & 92.1 & +11.0 & +3.6$^{**}$ & +14.6 & 43.1\% & 95.2 & 93.9 \\
\bottomrule\end{tabular}\end{table}

{\scriptsize
\setlength{\tabcolsep}{4pt}
\begin{longtable}{@{}l l l r l l@{}}
\caption{\textbf{Effort trend, per-size paired bootstrap (5000 resamples, shared problems).} diff $=$ lift(first) $-$ lift(second); $95\%$ CI; sign-based two-sided $p$. The trend is concentrated at Medium$\to$High; Low$\to$Medium is n.s.}\label{tab:effort-trend}\\
\toprule
Method & Size & Contrast & $n$ & $\Delta_{\mathrm{full}}$ diff [CI] ($p$) & $\Delta_{\mathrm{sel}}$ diff [CI] ($p$) \\
\midrule
\endfirsthead
\toprule
Method & Size & Contrast & $n$ & $\Delta_{\mathrm{full}}$ diff [CI] ($p$) & $\Delta_{\mathrm{sel}}$ diff [CI] ($p$) \\
\midrule
\endhead
\bottomrule
\endlastfoot
\multicolumn{6}{@{}l}{\textit{math}}\\
RAD & OSS-20B & Low-High & 111 & -3.6 [-12.1,+4.8] (0.387) & +9.3 [+2.8,+15.6] (0.003) \\
RAD+DC & OSS-20B & Low-High & 111 & -0.0 [-8.7,+8.3] (0.989) & +12.9 [+6.2,+19.3] (0.000) \\
RAD & OSS-20B & Medium-High & 111 & -0.1 [-5.5,+4.9] (0.996) & +10.4 [+5.3,+15.5] (0.000) \\
RAD+DC & OSS-20B & Medium-High & 111 & +1.7 [-4.4,+7.4] (0.558) & +12.2 [+6.4,+18.0] (0.000) \\
RAD & OSS-20B & Low-Medium & 120 & -3.6 [-11.0,+3.7] (0.347) & -1.0 [-8.0,+6.0] (0.798) \\
RAD+DC & OSS-20B & Low-Medium & 120 & -3.6 [-11.3,+4.2] (0.385) & -1.0 [-8.4,+6.3] (0.809) \\
RAD & OSS-120B & Low-High & 118 & +0.1 [-7.0,+6.9] (0.982) & +9.2 [+4.0,+14.3] (0.000) \\
RAD+DC & OSS-120B & Low-High & 118 & -0.7 [-8.1,+6.4] (0.833) & +8.3 [+2.8,+13.8] (0.003) \\
RAD & OSS-120B & Medium-High & 118 & -2.7 [-8.5,+2.8] (0.354) & +6.3 [+1.6,+10.9] (0.007) \\
RAD+DC & OSS-120B & Medium-High & 118 & -0.2 [-6.1,+5.1] (0.951) & +8.8 [+3.7,+14.0] (0.000) \\
RAD & OSS-120B & Low-Medium & 120 & +2.8 [-3.2,+8.7] (0.371) & +2.9 [-3.2,+8.8] (0.363) \\
RAD+DC & OSS-120B & Low-Medium & 120 & -0.5 [-7.5,+6.2] (0.882) & -0.5 [-7.4,+6.3] (0.897) \\
\addlinespace
\multicolumn{6}{@{}l}{\textit{well-posed}}\\
RAD & OSS-20B & Low-High & 305 & -4.2 [-8.9,+0.2] (0.064) & +5.0 [+1.1,+8.7] (0.012) \\
RAD+DC & OSS-20B & Low-High & 305 & -2.9 [-7.5,+1.4] (0.194) & +6.3 [+2.3,+10.1] (0.002) \\
RAD & OSS-20B & Medium-High & 305 & -3.4 [-7.0,+0.0] (0.052) & +4.5 [+1.3,+7.7] (0.008) \\
RAD+DC & OSS-20B & Medium-High & 305 & -3.1 [-6.8,+0.5] (0.094) & +4.8 [+1.4,+8.3] (0.008) \\
RAD & OSS-20B & Low-Medium & 318 & -0.7 [-4.9,+3.2] (0.702) & +0.6 [-3.4,+4.5] (0.792) \\
RAD+DC & OSS-20B & Low-Medium & 318 & -0.1 [-4.4,+4.0] (0.949) & +1.2 [-2.8,+5.2] (0.558) \\
RAD & OSS-120B & Low-High & 316 & -1.2 [-5.1,+2.4] (0.517) & +4.3 [+1.2,+7.4] (0.007) \\
RAD+DC & OSS-120B & Low-High & 316 & -2.5 [-6.2,+1.1] (0.183) & +3.0 [-0.1,+6.2] (0.059) \\
RAD & OSS-120B & Medium-High & 316 & -3.6 [-6.9,-0.5] (0.019) & +1.9 [-0.9,+4.6] (0.177) \\
RAD+DC & OSS-120B & Medium-High & 316 & -3.0 [-6.3,+0.1] (0.062) & +2.5 [-0.4,+5.3] (0.092) \\
RAD & OSS-120B & Low-Medium & 318 & +2.4 [-1.1,+6.0] (0.170) & +2.4 [-1.1,+6.0] (0.167) \\
RAD+DC & OSS-120B & Low-Medium & 318 & +0.5 [-3.3,+4.2] (0.818) & +0.5 [-3.3,+4.2] (0.808) \\
\end{longtable}
}

\begin{figure}[tb]\centering
  \includegraphics[width=0.7\linewidth]{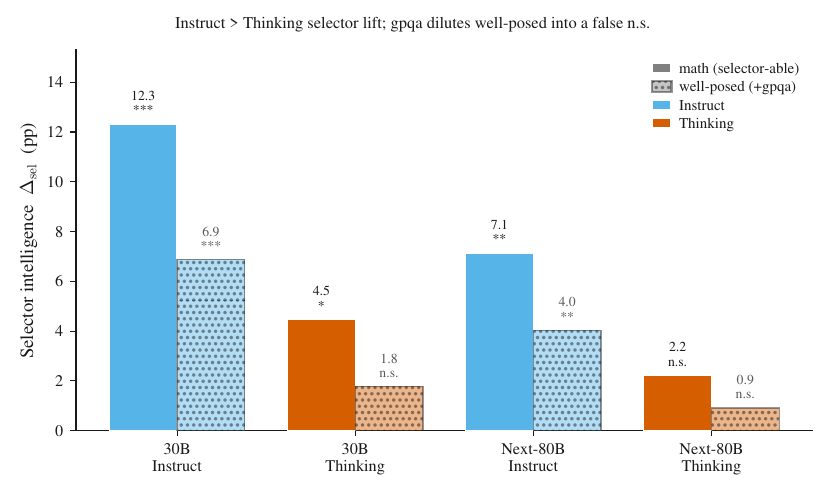}
  \caption{\textbf{Instruct $>$ thinking selector lift, and GPQA dilutes it.} $\Delta_{\mathrm{sel}}$ per Qwen model on math vs.\ well-posed (math$+$GPQA, weighted Jaccard). The selector is genuine on both variants but $2$--$3\times$ larger on instruct (diverse cohorts); GPQA is near-null for all, so the well-posed problem-weighted average flips Qwen3-30B-Thinking from significant ($+4.5^{*}$) to a false n.s.}
  \label{fig:think-instruct}
\end{figure}

\paragraph{Thinking vs.\ instruct.} The same mechanism explains the thinking/instruct gap. On the selector-able \emph{math} set both variants show genuine selector intelligence, but $2$--$3\times$ larger on instruct (Table~\ref{tab:think-instruct}, Figure~\ref{fig:think-instruct}): Qwen3-30B Instruct $+12.3^{***}$ vs.\ Thinking $+4.5^{*}$; Qwen3-Next-80B Instruct $+7.1^{**}$ vs.\ Thinking $+2.2$ (n.s.; RAD+DC $+4.9^{***}$), because thinking cohorts converge (avg@coh $87$--$92\%$) while instruct cohorts stay diverse ($61$--$69\%$). On GPQA the selector is near-null for all four ($\Delta_{\mathrm{sel}}=+0.1$ to $+3.6$, n.s.); since GPQA is ${\approx}63\%$ of the well-posed problems, the well-posed problem-weighted average flips Qwen3-30B-Thinking from a significant math effect ($+4.5^{*}$) to a misleading n.s.\ ($+1.8$), so we report \emph{math} for this contrast. The higher \emph{pick} accuracy on thinking models ($+14$--$17$\,pp) is base capability, not selector quality.

\begin{table}[!ht]\centering\small
\caption{\textbf{Thinking vs.\ instruct: the selector's intelligence is larger on instruct than thinking models, and GPQA dilutes it.} $\Delta_{\mathrm{sel}}$ (PB-exact). On math, instruct $>$ thinking; GPQA is near-null for all, so the well-posed problem-weighted average flips Qwen3-30B-Thinking to a misleading n.s.}
\label{tab:think-instruct}
\begin{tabular}{@{}l rr rrr r@{}}\toprule
 & avg@64 & avg@coh & \multicolumn{3}{c}{$\Delta_{\mathrm{sel}}$ (pp)} & RAD \\
 & {\scriptsize(math)} & {\scriptsize(math)} & & & & dens \\
\cmidrule(lr){4-6} Model & & & math & gpqa & well-posed & {\scriptsize(math)} \\\midrule
Qwen3-30B Instruct & 60.9 & 61.0 & +12.3$^{***}$ & +3.6 & +6.9$^{***}$ & 73.3 \\
Qwen3-30B Thinking & 80.0 & 87.1 & +4.5$^{*}$ & +0.1 & +1.8 & 91.5 \\
Qwen3-Next-80B Instruct & 68.6 & 68.7 & +7.1$^{**}$ & +2.1 & +4.0$^{**}$ & 75.8 \\
Qwen3-Next-80B Thinking & 83.6 & 91.6 & +2.2 & +0.1 & +0.9 & 93.9 \\
\bottomrule\end{tabular}\end{table}

\paragraph{DeepConf and fusion.} Adding DeepConf confidence weighting (RAD+DC) does not significantly change RAD in any well-posed or math cell after Holm correction (paired McNemar, Table~\ref{tab:dc}; pooled-by-effort $p=1.00/0.71/1.00$; the one nominally significant cell, Next-80B-Thinking well-posed $+1.9$\,pp $p=0.031$, does not survive Holm); its small help concentrates where routing density saturates (converged thinking cohorts), and under WJ no cell is significantly harmful. The two fusion variants (conf50, the deployed top-$50\%$ cBW filter then density, vs.\ rank-sum) are themselves indistinguishable (Table~\ref{tab:dc-fusion}), so the fusion choice is not a design axis.

{\scriptsize
\setlength{\tabcolsep}{5pt}
\begin{longtable}{@{}l r r r r r r r@{}}
\caption{\textbf{RAD+DC vs.\ RAD, paired exact McNemar (weighted Jaccard).} $\Delta=$ RAD+DC $-$ RAD (pp); \emph{rad only}/\emph{dc only} are the discordant counts. No well-posed/math cell is significant after Holm correction (one cell nominally $p{=}0.031$); DC only nudges narrow high-convergence cells.}\label{tab:dc}\\
\toprule
Group & $n$ & RAD & RAD+DC & $\Delta$ & rad only & dc only & McNemar $p$ \\
\midrule
\endfirsthead
\toprule
Group & $n$ & RAD & RAD+DC & $\Delta$ & rad only & dc only & McNemar $p$ \\
\midrule
\endhead
\bottomrule
\endlastfoot
\multicolumn{8}{@{}l}{\textit{math}}\\
OSS-20B-Low & 120 & 48.3 & 49.2 & +0.8 & 3 & 4 & 1.000 \\
OSS-20B-Medium & 120 & 83.3 & 84.2 & +0.8 & 2 & 3 & 1.000 \\
OSS-20B-High & 111 & 94.6 & 91.9 & -2.7 & 3 & 0 & 0.250 \\
OSS-120B-Low & 120 & 60.0 & 59.2 & -0.8 & 4 & 3 & 1.000 \\
OSS-120B-Medium & 120 & 81.7 & 84.2 & +2.5 & 2 & 5 & 0.453 \\
OSS-120B-High & 118 & 95.8 & 95.8 & +0.0 & 1 & 1 & 1.000 \\
OSS-pooled-Low & 240 & 54.2 & 54.2 & +0.0 & 7 & 7 & 1.000 \\
OSS-pooled-Medium & 240 & 82.5 & 84.2 & +1.7 & 4 & 8 & 0.388 \\
OSS-pooled-High & 229 & 95.2 & 93.9 & -1.3 & 4 & 1 & 0.375 \\
30B-Thinking & 118 & 91.5 & 91.5 & +0.0 & 1 & 1 & 1.000 \\
30B-Instruct & 120 & 73.3 & 74.2 & +0.8 & 0 & 1 & 1.000 \\
Next-80B-Thinking & 114 & 93.9 & 96.5 & +2.6 & 0 & 3 & 0.250 \\
Next-80B-Instruct & 120 & 75.8 & 77.5 & +1.7 & 1 & 3 & 0.625 \\
\addlinespace
\multicolumn{8}{@{}l}{\textit{gpqa}}\\
OSS-20B-Low & 198 & 60.6 & 61.6 & +1.0 & 4 & 6 & 0.754 \\
OSS-20B-Medium & 198 & 67.7 & 67.7 & +0.0 & 5 & 5 & 1.000 \\
OSS-20B-High & 194 & 69.1 & 70.1 & +1.0 & 6 & 8 & 0.791 \\
OSS-120B-Low & 198 & 69.2 & 67.7 & -1.5 & 5 & 2 & 0.453 \\
OSS-120B-Medium & 198 & 72.7 & 72.2 & -0.5 & 4 & 3 & 1.000 \\
OSS-120B-High & 198 & 81.3 & 81.3 & +0.0 & 2 & 2 & 1.000 \\
OSS-pooled-Low & 396 & 64.9 & 64.6 & -0.3 & 9 & 8 & 1.000 \\
OSS-pooled-Medium & 396 & 70.2 & 69.9 & -0.3 & 9 & 8 & 1.000 \\
OSS-pooled-High & 392 & 75.3 & 75.8 & +0.5 & 8 & 10 & 0.815 \\
30B-Thinking & 198 & 71.2 & 72.2 & +1.0 & 2 & 4 & 0.688 \\
30B-Instruct & 198 & 70.2 & 68.2 & -2.0 & 7 & 3 & 0.344 \\
Next-80B-Thinking & 194 & 77.8 & 79.4 & +1.5 & 0 & 3 & 0.250 \\
Next-80B-Instruct & 198 & 74.7 & 75.3 & +0.5 & 2 & 3 & 1.000 \\
\addlinespace
\multicolumn{8}{@{}l}{\textit{well-posed}}\\
OSS-20B-Low & 318 & 56.0 & 56.9 & +0.9 & 7 & 10 & 0.629 \\
OSS-20B-Medium & 318 & 73.6 & 73.9 & +0.3 & 7 & 8 & 1.000 \\
OSS-20B-High & 305 & 78.4 & 78.0 & -0.3 & 9 & 8 & 1.000 \\
OSS-120B-Low & 318 & 65.7 & 64.5 & -1.3 & 9 & 5 & 0.424 \\
OSS-120B-Medium & 318 & 76.1 & 76.7 & +0.6 & 6 & 8 & 0.791 \\
OSS-120B-High & 316 & 86.7 & 86.7 & +0.0 & 3 & 3 & 1.000 \\
OSS-pooled-Low & 636 & 60.8 & 60.7 & -0.2 & 16 & 15 & 1.000 \\
OSS-pooled-Medium & 636 & 74.8 & 75.3 & +0.5 & 13 & 16 & 0.711 \\
OSS-pooled-High & 621 & 82.6 & 82.4 & -0.2 & 12 & 11 & 1.000 \\
30B-Thinking & 316 & 78.8 & 79.4 & +0.6 & 3 & 5 & 0.727 \\
30B-Instruct & 318 & 71.4 & 70.4 & -0.9 & 7 & 4 & 0.549 \\
Next-80B-Thinking & 308 & 83.8 & 85.7 & +1.9 & 0 & 6 & 0.031 \\
Next-80B-Instruct & 318 & 75.2 & 76.1 & +0.9 & 3 & 6 & 0.508 \\
\addlinespace
\multicolumn{8}{@{}l}{\textit{code}}\\
OSS-20B-Low & 167 & 38.9 & 41.9 & +3.0 & 11 & 16 & 0.442 \\
OSS-20B-Medium & 167 & 72.5 & 73.7 & +1.2 & 6 & 8 & 0.791 \\
OSS-20B-High & 139 & 84.2 & 82.0 & -2.2 & 9 & 6 & 0.607 \\
OSS-120B-Low & 167 & 70.1 & 67.7 & -2.4 & 7 & 3 & 0.344 \\
OSS-120B-Medium & 167 & 79.0 & 78.4 & -0.6 & 4 & 3 & 1.000 \\
OSS-120B-High & 155 & 88.4 & 89.7 & +1.3 & 1 & 3 & 0.625 \\
OSS-pooled-Low & 334 & 54.5 & 54.8 & +0.3 & 18 & 19 & 1.000 \\
OSS-pooled-Medium & 334 & 75.7 & 76.0 & +0.3 & 10 & 11 & 1.000 \\
OSS-pooled-High & 294 & 86.4 & 86.1 & -0.3 & 10 & 9 & 1.000 \\
30B-Thinking & 162 & 74.7 & 77.2 & +2.5 & 4 & 8 & 0.388 \\
30B-Instruct & 167 & 45.5 & 44.9 & -0.6 & 5 & 4 & 1.000 \\
Next-80B-Thinking & 167 & 70.7 & 72.5 & +1.8 & 4 & 7 & 0.549 \\
Next-80B-Instruct & 167 & 63.5 & 65.9 & +2.4 & 2 & 6 & 0.289 \\
\end{longtable}
}

{\scriptsize
\setlength{\tabcolsep}{5pt}
\begin{longtable}{@{}l r r r r r r r@{}}
\caption{\textbf{DeepConf fusion robustness: top-$50\%$ cBW filter (conf50) vs.\ rank-sum (weighted Jaccard).} conf50 keeps the top-$50\%$ of rollouts by DeepConf cBW, then picks the densest by routing $K$-NN; $\Delta=$ rankfuse $-$ conf50 (pp), paired exact McNemar. The two fusions agree within noise everywhere, so the fusion choice is not a design axis.}\label{tab:dc-fusion}\\
\toprule
Group & $n$ & conf50 & rankfuse & $\Delta$ & c50 only & rf only & McNemar $p$ \\
\midrule
\endfirsthead
\toprule
Group & $n$ & conf50 & rankfuse & $\Delta$ & c50 only & rf only & McNemar $p$ \\
\midrule
\endhead
\bottomrule
\endlastfoot
\multicolumn{8}{@{}l}{\textit{math}}\\
OSS-20B-Low & 120 & 49.2 & 48.3 & -0.8 & 3 & 2 & 1.000 \\
OSS-20B-Medium & 120 & 84.2 & 83.3 & -0.8 & 6 & 5 & 1.000 \\
OSS-20B-High & 111 & 91.9 & 93.7 & +1.8 & 0 & 2 & 0.500 \\
OSS-120B-Low & 120 & 59.2 & 61.7 & +2.5 & 1 & 4 & 0.375 \\
OSS-120B-Medium & 120 & 84.2 & 84.2 & +0.0 & 2 & 2 & 1.000 \\
OSS-120B-High & 118 & 95.8 & 94.9 & -0.8 & 2 & 1 & 1.000 \\
OSS-pooled-Low & 240 & 54.2 & 55.0 & +0.8 & 4 & 6 & 0.754 \\
OSS-pooled-Medium & 240 & 84.2 & 83.8 & -0.4 & 8 & 7 & 1.000 \\
OSS-pooled-High & 229 & 93.9 & 94.3 & +0.4 & 2 & 3 & 1.000 \\
30B-Thinking & 118 & 91.5 & 91.5 & +0.0 & 0 & 0 & 1.000 \\
30B-Instruct & 120 & 74.2 & 72.5 & -1.7 & 2 & 0 & 0.500 \\
Next-80B-Thinking & 114 & 96.5 & 93.9 & -2.6 & 3 & 0 & 0.250 \\
Next-80B-Instruct & 120 & 77.5 & 77.5 & +0.0 & 1 & 1 & 1.000 \\
\addlinespace
\multicolumn{8}{@{}l}{\textit{gpqa}}\\
OSS-20B-Low & 198 & 61.6 & 59.6 & -2.0 & 9 & 5 & 0.424 \\
OSS-20B-Medium & 198 & 67.7 & 67.2 & -0.5 & 5 & 4 & 1.000 \\
OSS-20B-High & 194 & 70.1 & 69.6 & -0.5 & 5 & 4 & 1.000 \\
OSS-120B-Low & 198 & 67.7 & 69.7 & +2.0 & 1 & 5 & 0.219 \\
OSS-120B-Medium & 198 & 72.2 & 72.7 & +0.5 & 4 & 5 & 1.000 \\
OSS-120B-High & 198 & 81.3 & 81.8 & +0.5 & 1 & 2 & 1.000 \\
OSS-pooled-Low & 396 & 64.6 & 64.6 & +0.0 & 10 & 10 & 1.000 \\
OSS-pooled-Medium & 396 & 69.9 & 69.9 & +0.0 & 9 & 9 & 1.000 \\
OSS-pooled-High & 392 & 75.8 & 75.8 & +0.0 & 6 & 6 & 1.000 \\
30B-Thinking & 198 & 72.2 & 72.2 & +0.0 & 3 & 3 & 1.000 \\
30B-Instruct & 198 & 68.2 & 68.7 & +0.5 & 3 & 4 & 1.000 \\
Next-80B-Thinking & 194 & 79.4 & 78.4 & -1.0 & 2 & 0 & 0.500 \\
Next-80B-Instruct & 198 & 75.3 & 74.7 & -0.5 & 2 & 1 & 1.000 \\
\addlinespace
\multicolumn{8}{@{}l}{\textit{well-posed}}\\
OSS-20B-Low & 318 & 56.9 & 55.3 & -1.6 & 12 & 7 & 0.359 \\
OSS-20B-Medium & 318 & 73.9 & 73.3 & -0.6 & 11 & 9 & 0.824 \\
OSS-20B-High & 305 & 78.0 & 78.4 & +0.3 & 5 & 6 & 1.000 \\
OSS-120B-Low & 318 & 64.5 & 66.7 & +2.2 & 2 & 9 & 0.065 \\
OSS-120B-Medium & 318 & 76.7 & 77.0 & +0.3 & 6 & 7 & 1.000 \\
OSS-120B-High & 316 & 86.7 & 86.7 & +0.0 & 3 & 3 & 1.000 \\
OSS-pooled-Low & 636 & 60.7 & 61.0 & +0.3 & 14 & 16 & 0.856 \\
OSS-pooled-Medium & 636 & 75.3 & 75.2 & -0.2 & 17 & 16 & 1.000 \\
OSS-pooled-High & 621 & 82.4 & 82.6 & +0.2 & 8 & 9 & 1.000 \\
30B-Thinking & 316 & 79.4 & 79.4 & +0.0 & 3 & 3 & 1.000 \\
30B-Instruct & 318 & 70.4 & 70.1 & -0.3 & 5 & 4 & 1.000 \\
Next-80B-Thinking & 308 & 85.7 & 84.1 & -1.6 & 5 & 0 & 0.062 \\
Next-80B-Instruct & 318 & 76.1 & 75.8 & -0.3 & 3 & 2 & 1.000 \\
\addlinespace
\multicolumn{8}{@{}l}{\textit{code}}\\
OSS-20B-Low & 167 & 41.9 & 41.9 & +0.0 & 21 & 21 & 1.000 \\
OSS-20B-Medium & 167 & 73.7 & 74.3 & +0.6 & 7 & 8 & 1.000 \\
OSS-20B-High & 139 & 82.0 & 81.3 & -0.7 & 8 & 7 & 1.000 \\
OSS-120B-Low & 167 & 67.7 & 67.1 & -0.6 & 5 & 4 & 1.000 \\
OSS-120B-Medium & 167 & 78.4 & 80.2 & +1.8 & 2 & 5 & 0.453 \\
OSS-120B-High & 155 & 89.7 & 91.0 & +1.3 & 2 & 4 & 0.688 \\
OSS-pooled-Low & 334 & 54.8 & 54.5 & -0.3 & 26 & 25 & 1.000 \\
OSS-pooled-Medium & 334 & 76.0 & 77.2 & +1.2 & 9 & 13 & 0.523 \\
OSS-pooled-High & 294 & 86.1 & 86.4 & +0.3 & 10 & 11 & 1.000 \\
30B-Thinking & 162 & 77.2 & 77.8 & +0.6 & 4 & 5 & 1.000 \\
30B-Instruct & 167 & 44.9 & 47.3 & +2.4 & 6 & 10 & 0.454 \\
Next-80B-Thinking & 167 & 72.5 & 73.1 & +0.6 & 9 & 10 & 1.000 \\
Next-80B-Instruct & 167 & 65.9 & 66.5 & +0.6 & 9 & 10 & 1.000 \\
\end{longtable}
}

\paragraph{Per-model decomposition.} Table~\ref{tab:per-model} gives the full decomposition for all ten models across the four pools. On code (LiveCodeBench) graded rollouts are a subset of anchor-found ones, so $\Delta_{\mathrm{anchor}}$ is large and mechanical, a format/completion-compliance signal (emitting a code fence marks a complete, more-often-correct rollout, a correctness proxy rather than answer content), and string-voting baselines degenerate; only the routing density $\Delta_{\mathrm{sel}}$ is a genuine selector measure there. Even so, against the Avg@64 floor reported in Table~\ref{tab:main} RAD's code selection is positive for every model family; $\Delta_{\mathrm{sel}}$ is the stricter test (its baseline is the within-cohort random pick \texttt{avg@coh}, which already contains the mechanical $\Delta_{\mathrm{anchor}}$) so a small or occasionally negative $\Delta_{\mathrm{sel}}$ measures selection \emph{on top of} an already-elevated cohort, not a loss against the reported floor.

{\scriptsize
\setlength{\tabcolsep}{4pt}
\begin{longtable}{@{}l r r r r r r r r r@{}}
\caption{\textbf{Per-model full decomposition (RAD, weighted Jaccard).} All pools; $n$ is the \emph{cohort-eligible} count (problems with $\ge 2$ anchor-located rollouts), a subset of the full sets reported by the headline Tables~\ref{tab:main}/\ref{tab:marker-only}: these $n$ sum to $3153$ of the $3180$ well-posed and $1625$ of the $1670$ code problems (headline accuracies use the full sets; this decomposition uses the cohort-eligible subset). $\Delta_{\mathrm{sel}}$ PB-exact. On code, graded $\subseteq$ anchor-found so maj degenerates (\texttt{--}) and only $\Delta_{\mathrm{sel}}$ is meaningful.}\label{tab:per-model}\\
\toprule
Model & $n$ & avg@full & avg@coh & maj & dens & $\Delta_{\mathrm{anc}}$ & $\Delta_{\mathrm{sel}}$ & $\Delta_{\mathrm{full}}$ & $\Delta_{\mathrm{sel}}$/HR \\
\midrule
\endfirsthead
\toprule
Model & $n$ & avg@full & avg@coh & maj & dens & $\Delta_{\mathrm{anc}}$ & $\Delta_{\mathrm{sel}}$ & $\Delta_{\mathrm{full}}$ & $\Delta_{\mathrm{sel}}$/HR \\
\midrule
\endhead
\bottomrule
\endlastfoot
\multicolumn{10}{@{}l}{\textit{math}}\\
OSS-120B-High & 118 & 84.4 & 93.4 & 95.8 & 95.8 & +9.1 & +2.3 & +11.4 & 35.3\% \\
OSS-120B-Low & 120 & 48.8 & 48.8 & 58.3 & 60.0 & +0.0 & +11.2$^{***}$ & +11.2 & 21.9\% \\
OSS-120B-Medium & 120 & 73.3 & 73.3 & 84.2 & 81.7 & +0.1 & +8.3$^{**}$ & +8.4 & 31.3\% \\
OSS-20B-High & 111 & 76.5 & 89.6 & 88.3 & 94.6 & +13.1 & +5.0$^{**}$ & +18.1 & 48.2\% \\
OSS-20B-Low & 120 & 35.1 & 35.2 & 48.3 & 48.3 & +0.1 & +13.1$^{***}$ & +13.3 & 20.3\% \\
OSS-20B-Medium & 120 & 66.5 & 69.2 & 80.8 & 83.3 & +2.7 & +14.2$^{***}$ & +16.9 & 45.9\% \\
Qwen3-30B-A3B-Instruct-2507 & 120 & 60.9 & 61.0 & 70.8 & 73.3 & +0.1 & +12.3$^{***}$ & +12.4 & 31.6\% \\
Qwen3-30B-A3B-Thinking-2507 & 118 & 80.0 & 87.1 & 88.1 & 91.5 & +7.0 & +4.5$^{*}$ & +11.5 & 34.5\% \\
Qwen3-Next-80B-A3B-Instruct & 120 & 68.6 & 68.7 & 74.2 & 75.8 & +0.0 & +7.1$^{**}$ & +7.2 & 22.8\% \\
Qwen3-Next-80B-A3B-Thinking & 114 & 83.6 & 91.6 & 90.4 & 93.9 & +8.0 & +2.2 & +10.2 & 26.6\% \\
\addlinespace
\multicolumn{10}{@{}l}{\textit{gpqa}}\\
OSS-120B-High & 198 & 75.4 & 78.8 & 83.3 & 81.3 & +3.4 & +2.5 & +5.9 & 11.7\% \\
OSS-120B-Low & 198 & 65.3 & 65.3 & 65.7 & 69.2 & +0.0 & +3.9 & +3.9 & 11.2\% \\
OSS-120B-Medium & 198 & 71.0 & 71.0 & 73.2 & 72.7 & +0.0 & +1.7 & +1.7 & 6.0\% \\
OSS-20B-High & 194 & 60.5 & 67.5 & 72.2 & 69.1 & +7.0 & +1.6 & +8.6 & 4.9\% \\
OSS-20B-Low & 198 & 56.3 & 56.3 & 61.6 & 60.6 & +0.0 & +4.3 & +4.3 & 9.9\% \\
OSS-20B-Medium & 198 & 64.4 & 64.9 & 68.2 & 67.7 & +0.6 & +2.7 & +3.3 & 7.8\% \\
Qwen3-30B-A3B-Instruct-2507 & 198 & 66.2 & 66.6 & 69.2 & 70.2 & +0.5 & +3.6 & +4.0 & 10.7\% \\
Qwen3-30B-A3B-Thinking-2507 & 198 & 71.1 & 71.1 & 71.2 & 71.2 & +0.0 & +0.1 & +0.1 & 0.5\% \\
Qwen3-Next-80B-A3B-Instruct & 198 & 72.2 & 72.6 & 75.3 & 74.7 & +0.4 & +2.1 & +2.5 & 7.8\% \\
Qwen3-Next-80B-A3B-Thinking & 194 & 62.0 & 77.7 & 79.9 & 77.8 & +15.7 & +0.1 & +15.8 & 0.6\% \\
\addlinespace
\multicolumn{10}{@{}l}{\textit{well-posed}}\\
OSS-120B-High & 316 & 78.7 & 84.3 & 88.0 & 86.7 & +5.6 & +2.4 & +8.0 & 15.4\% \\
OSS-120B-Low & 318 & 59.1 & 59.1 & 62.9 & 65.7 & +0.0 & +6.6$^{***}$ & +6.6 & 16.2\% \\
OSS-120B-Medium & 318 & 71.9 & 71.9 & 77.4 & 76.1 & +0.0 & +4.2$^{**}$ & +4.3 & 15.0\% \\
OSS-20B-High & 305 & 66.3 & 75.5 & 78.0 & 78.4 & +9.2 & +2.8$^{*}$ & +12.0 & 11.6\% \\
OSS-20B-Low & 318 & 48.3 & 48.3 & 56.6 & 56.0 & +0.1 & +7.6$^{***}$ & +7.7 & 14.8\% \\
OSS-20B-Medium & 318 & 65.2 & 66.5 & 73.0 & 73.6 & +1.4 & +7.0$^{***}$ & +8.4 & 21.1\% \\
Qwen3-30B-A3B-Instruct-2507 & 318 & 64.2 & 64.5 & 69.8 & 71.4 & +0.3 & +6.9$^{***}$ & +7.2 & 19.3\% \\
Qwen3-30B-A3B-Thinking-2507 & 316 & 74.4 & 77.0 & 77.5 & 78.8 & +2.6 & +1.8 & +4.4 & 7.7\% \\
Qwen3-Next-80B-A3B-Instruct & 318 & 70.9 & 71.1 & 74.8 & 75.2 & +0.3 & +4.0$^{**}$ & +4.3 & 14.0\% \\
Qwen3-Next-80B-A3B-Thinking & 308 & 70.0 & 82.9 & 83.8 & 83.8 & +12.8 & +0.9 & +13.7 & 5.3\% \\
\addlinespace
\multicolumn{10}{@{}l}{\textit{code}}\\
OSS-120B-High & 155 & 76.9 & 85.5 & -- & 88.4 & +8.7 & +2.9 & +11.5 & 19.7\% \\
OSS-120B-Low & 167 & 64.3 & 64.3 & -- & 70.1 & +0.0 & +5.7$^{**}$ & +5.7 & 16.1\% \\
OSS-120B-Medium & 167 & 76.1 & 76.2 & -- & 79.0 & +0.1 & +2.9 & +3.0 & 12.1\% \\
OSS-20B-High & 139 & 61.7 & 77.3 & -- & 84.2 & +15.6 & +6.9$^{*}$ & +22.4 & 30.3\% \\
OSS-20B-Low & 167 & 33.5 & 33.8 & -- & 38.9 & +0.3 & +5.1 & +5.4 & 7.7\% \\
OSS-20B-Medium & 167 & 67.3 & 68.3 & -- & 72.5 & +1.0 & +4.1 & +5.1 & 13.0\% \\
Qwen3-30B-A3B-Instruct-2507 & 167 & 47.1 & 47.1 & -- & 45.5 & +0.0 & -1.6 & -1.5 & -3.0\% \\
Qwen3-30B-A3B-Thinking-2507 & 162 & 71.4 & 73.8 & -- & 74.7 & +2.3 & +0.9 & +3.3 & 3.6\% \\
Qwen3-Next-80B-A3B-Instruct & 167 & 62.8 & 62.9 & -- & 63.5 & +0.0 & +0.6 & +0.6 & 1.6\% \\
Qwen3-Next-80B-A3B-Thinking & 167 & 70.1 & 71.2 & -- & 70.7 & +1.1 & -0.5 & +0.5 & -1.8\% \\
\end{longtable}
}

\subsection{Binary vs.\ weighted Jaccard kernel}
\label{app:wjja}
RAD scores agreement with a $k$-NN density in routing space; the similarity kernel can be either binary Jaccard (JA, set overlap of activated experts) or weighted Jaccard (WJ, $\sum_e\min(u_e,v_e)/\sum_e\max(u_e,v_e)$ of routing weights). We test whether the routing \emph{weights} matter with a within-cell paired ablation that holds the anchor, cohort, baselines, and all $15{,}188$ problem cells (8 thinking models $\times$ 6 datasets, cohort $\geq 2$) fixed and varies \emph{only} the kernel between two otherwise-identical runs. The two kernels pick a different rollout on just $816$ cells ($5.4\%$); only these carry kernel signal.

The difference is non-significant in every anchor mode and all Holm-adjusted p-values are $1.000$ (Table~\ref{tab:wjja-main}); the pooled gap is WJ$-$JA $=+0.22$\,pp with a shard cluster-bootstrap 95\% CI $[-0.18,+0.59]$ that crosses zero. No category cell is significant (Table~\ref{tab:wjja-cat}), and no per-model / per-dataset / category cut survives BH-FDR ($0/26$); the two Qwen models lean in opposite directions, consistent with noise rather than a systematic edge. The ablation is adequately powered (Table~\ref{tab:wjja-power}): pooled power to detect a true $1$\,pp effect is ${\approx}\,1.00$ and the pooled significance boundary ($0.38$\,pp) exceeds the pooled observed gap ($0.22$\,pp), so a paper-relevant ($\geq 1$\,pp) \emph{pooled} kernel effect would almost certainly have been detected, an informative null, not a false negative. The kernel is therefore not load-bearing: the agreement signal lives in the routing geometry, not in the weighting, so binary and weighted Jaccard are interchangeable in practice. Numbers reproduce one-click via \texttt{wj\_ja\_compare.py}.

\input{tables/wj_ja_appendix}

\subsection{Layer-depth profile of the routing signal}
\label{app:layer-depth}
This appendix gives the full layer-depth ablation summarised in \S\ref{sec:exp:analysis}. We profile the selector at depth-decile resolution: the ten deciles of the MoE stack indexed from the shallowest, each ablated three ways, a single decile alone ($s_0\!\ldots\!s_9$), the cumulative shallow band ($c_0\!\ldots\!c_8$, growing from the top), and leave-one-decile-out ($x_0\!\ldots\!x_9$), under all four anchor configurations (boundary vs.\ delimiter anchor $\times$ anchor-only vs.\ 16-token window) and both the weighted (WJ) and binary (JA) Jaccard kernels. Picks reproduce an independent band-level recomputation bit-for-bit ($45{,}608$ problem-level cells, zero mismatches), so the decile profile and the coarser band picks are the same underlying selection. Significance uses the exact paired McNemar test (full vs.\ each band, $\Delta=$ full $-$ band) with Holm correction within each configuration, plus an omnibus Cochran's $Q$ over each band family; all numbers regenerate via \texttt{claim\_audit.py}.

\paragraph{Necessity and sufficiency.} Across all four configurations and both kernels, no single decile is necessary: every leave-one-decile-out change is $\le 0.47$\,pp and $0/10$ survive Holm in each cell. The shallowest tenth alone (cumulative band $c_0$) is within $0.7$\,pp of the full network at the point estimate ($95\%$ deficit upper bound $\le 1.7$\,pp, $\le 1.2$\,pp for the windowed configurations) and is never significantly worse (per-config McNemar $p\ge 0.23$; pooled $p=0.19$). For the marker-only (anchor-token) configurations no single decile differs significantly from the full network (every per-config McNemar Holm-adjusts to n.s.); for the boundary anchor the single-decile omnibus is also not significant (Cochran's $Q$ $p=0.67$, Table~\ref{tab:layer-deciles}). So the sufficiency of a small slice is \emph{not} a shallow-only effect.

\begin{table}[!ht]
  \centering
  \footnotesize
  \caption{\textbf{Single-decile layer profile for the marker-only (boundary anchor) configuration} (weighted Jaccard, $n=3786$ problems, full-network accuracy $71.16\%$). Each row uses \emph{only} one tenth of the MoE layers, indexed from the shallowest ($s_0$) to the deepest ($s_9$); $\Delta$ is full $-$ decile (positive $=$ full better), with the exact paired McNemar $p$ and its Holm-adjusted value. No single decile (shallow, middle, or deep) differs significantly from the full network (omnibus Cochran's $Q=6.72$, $\mathrm{df}=9$, $p=0.67$; $0/10$ survive Holm), so a small slice suffices regardless of depth.}
  \label{tab:layer-deciles}
  \begin{tabular}{lrrrr}
    \toprule
    Decile & Acc.\ (\%) & $\Delta$ (pp) & McNemar $p$ & Holm \\
    \midrule
    $s_0$ (shallowest) & 70.73 & $+0.42$ & 0.456 & 1.000 \\
    $s_1$              & 71.24 & $-0.08$ & 0.921 & 1.000 \\
    $s_2$              & 70.31 & $+0.85$ & 0.134 & 1.000 \\
    $s_3$              & 70.05 & $+1.11$ & 0.045 & 0.448 \\
    $s_4$              & 70.73 & $+0.42$ & 0.451 & 1.000 \\
    $s_5$              & 70.81 & $+0.34$ & 0.560 & 1.000 \\
    $s_6$              & 70.71 & $+0.45$ & 0.422 & 1.000 \\
    $s_7$              & 70.36 & $+0.79$ & 0.168 & 1.000 \\
    $s_8$              & 70.36 & $+0.79$ & 0.164 & 1.000 \\
    $s_9$ (deepest)    & 70.29 & $+0.87$ & 0.130 & 1.000 \\
    \midrule
    full               & 71.16 & ---     & ---   & ---   \\
    \bottomrule
  \end{tabular}
\end{table}

\begin{figure}[!ht]
  \centering
  \includegraphics[width=0.8\linewidth]{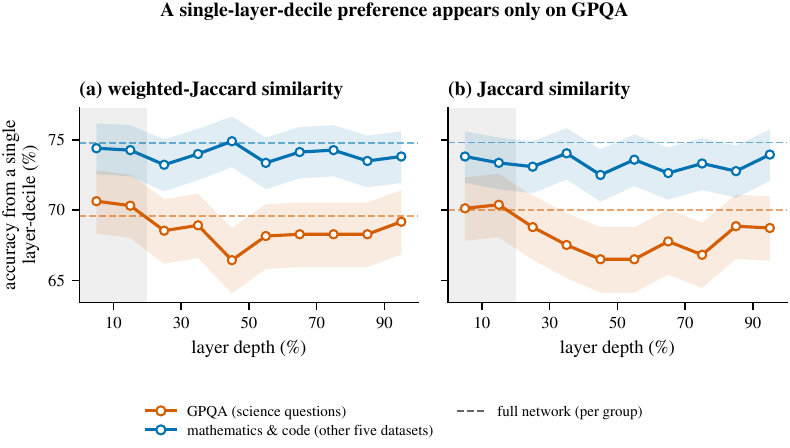}
  \caption{\textbf{The single-layer-decile preference is specific to GPQA.} Selection accuracy using \emph{only one} layer-decile of routing, as a function of layer depth, for the boundary anchor with a 16-token routing window, split into GPQA (graduate-level science questions) vs.\ the five mathematics and code datasets, under the \textbf{(a)} weighted-Jaccard (primary) and \textbf{(b)} Jaccard similarity kernel; dashed lines mark each group's full-network accuracy. On GPQA the shallow deciles match the full network while the middle deciles fall about 3 percentage points lower (single-decile omnibus Cochran's-$Q$ test $p<10^{-4}$ under both kernels); on mathematics and code the profile is flat ($p=0.25$ weighted-Jaccard, $0.30$ Jaccard; not significant). The pooled preference is therefore driven by GPQA and is not a general depth-localization law; it does not raise full-network accuracy.}
  \label{fig:layer-localization-gpqa}
\end{figure}

\begin{figure}[!ht]
  \centering
  \includegraphics[width=0.68\linewidth]{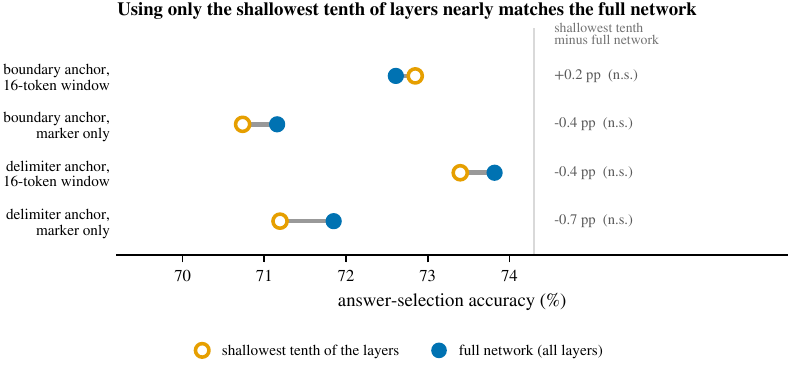}
  \caption{\textbf{Using only the shallowest tenth of the layers nearly matches the full network, for every selector setting.} For each of the four settings (anchor point $\times$ routing window), open markers give the answer-selection accuracy when the selector reads only the shallowest tenth of the mixture-of-experts layers, and filled markers give the full-network accuracy (weighted-Jaccard similarity kernel); the bar links the two. The shallowest tenth is within $0.7$ percentage points of the full network in every setting (for the boundary$+$16-token setting nominally $0.2$\,pp above), and the difference is not statistically significant in any of them (paired McNemar exact test, $p\ge0.23$). The Jaccard kernel reproduces this (gaps $\le0.8$ percentage points; not significant), and the audit script regenerates all values.}
  \label{fig:shallow-vs-full-dumbbell}
\end{figure}

\paragraph{Where a single-decile preference does appear, it is GPQA-specific.} At decile resolution the two \emph{windowed} configurations do show a shallow-favoring localization (single-decile omnibus Cochran's $Q$ significant: boundary$+$16-token $p=6.5\times10^{-3}$ WJ / $2\times10^{-4}$ JA). This significance is driven by GPQA: the shallow$-$middle gap is $+2.9$\,pp (WJ) / $+3.3$\,pp (JA) on GPQA ($Q\,p<10^{-4}$) but only $+0.2$\,pp (WJ, $Q\,p=0.25$) / $+0.7$\,pp (JA, $Q\,p=0.30$) outside GPQA, not significant in either kernel (Figure~\ref{fig:layer-localization-gpqa}). It does not raise full-network accuracy and is best read as a GPQA-specific interpretability observation, not a general depth-localization law. Figure~\ref{fig:shallow-vs-full-dumbbell} gives the per-configuration shallowest-tenth-vs-full comparison underlying the main-text sufficiency result.

\subsection{Composition control: what the routing selector selects}
\label{app:rollout}
This appendix details the composition control that dissects the RAD selector (\S\ref{sec:exp:analysis}, Figure~\ref{fig:rad-composition}). We pool the eight thinking-class model configurations (gpt-oss-20B/120B at Low/Medium/High effort, Qwen3-30B-Thinking, Qwen3-Next-80B-Thinking) over the six datasets at $N{=}64$ rollouts per problem and study two things: (i) how accuracy scales with the cohort size $m\in\{2,4,8,16,32,64\}$, and (ii) a composition control that fixes $m$ and injects exactly $c$ correct rollouts, sweeping the correct fraction $f{=}c/m$. RAD uses the weighted-Jaccard $k$-NN density ($k{=}10$, window $16$); RAD+DC additionally filters by DeepConf confidence (top-$50\%$ by $cBW$, then density within the kept set); maj is the answer-aware plurality reference and oracle is pass@$m$. Every selector is answer-string-free except maj. We report both the \emph{boundary} anchor (end-of-thinking / final-response marker; $n{=}2832$ at pool$\ge$64) and the deployed \emph{delimiter} anchor (\texttt{\textbackslash boxed}\,/\,code-fence marker; $n{=}2636$); the paper default is the delimiter anchor (Table~\ref{tab:main}). Uncertainty is a two-stage bootstrap ($B{=}200$ sub-cohort draws per cell, then a $2000$-resample cluster bootstrap over problems). \textbf{Eval-conditioning.} The pool is the anchor-found $\wedge$ labeled $\wedge$ feature-present rollouts (mean anchor coverage ${\approx}0.90$); the \emph{absolute} accuracies below are therefore eval-conditioned and sit several points above an all-problems full-$64$ deployment. They evaluate the selector on its applicable subset, not a deployment number. The lift-over-chance composition results (controlled $f$) and all paired CIs are unaffected by this conditioning.

\begin{figure}[!ht]
  \centering
  \includegraphics[width=0.72\linewidth]{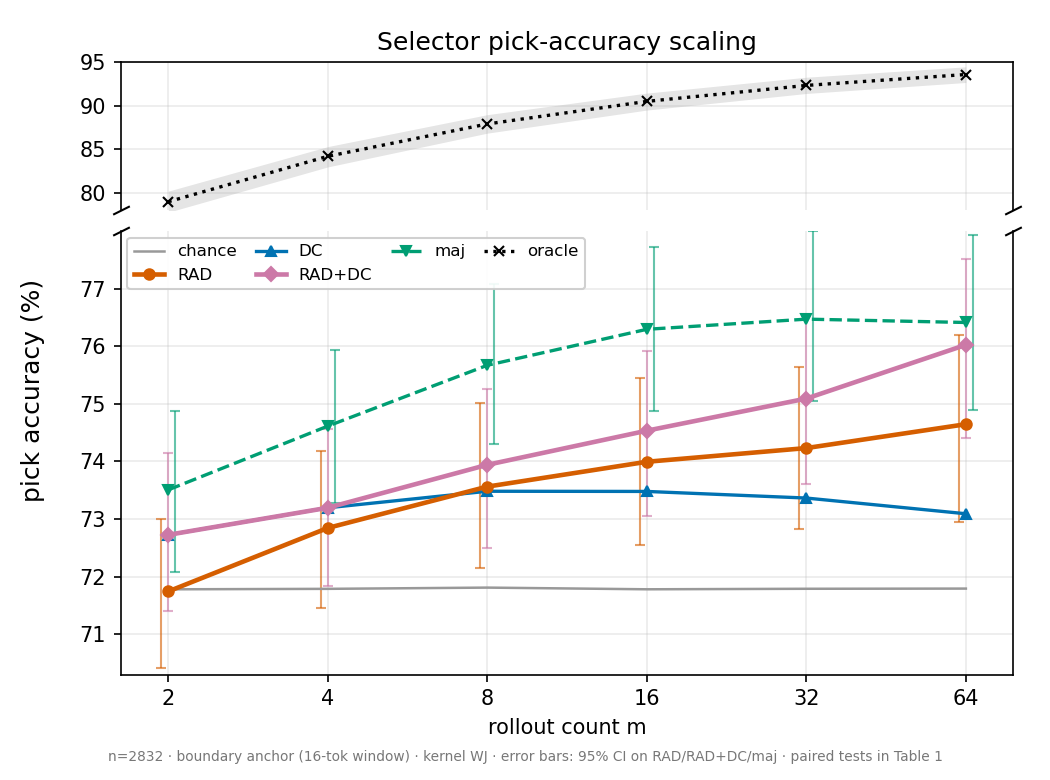}
  \caption{\textbf{Pick accuracy scales with the rollout count $m$} (boundary anchor, $n{=}2832$, weighted Jaccard; broken $y$-axis: the ${\approx}70$--$78\%$ selector comparison below, the $78$--$95\%$ oracle ceiling above). RAD rises with $m$ while DeepConf saturates by $m{\approx}8$; fused, RAD+DC tracks the answer-aware Majority. Error bars are $95\%$ cluster-bootstrap CIs on RAD/RAD+DC/maj (paired tests in Table~\ref{tab:rollout-scaling}).}
  \label{fig:rollout-scaling}
\end{figure}

\begin{figure}[!ht]
  \centering
  \includegraphics[width=\linewidth]{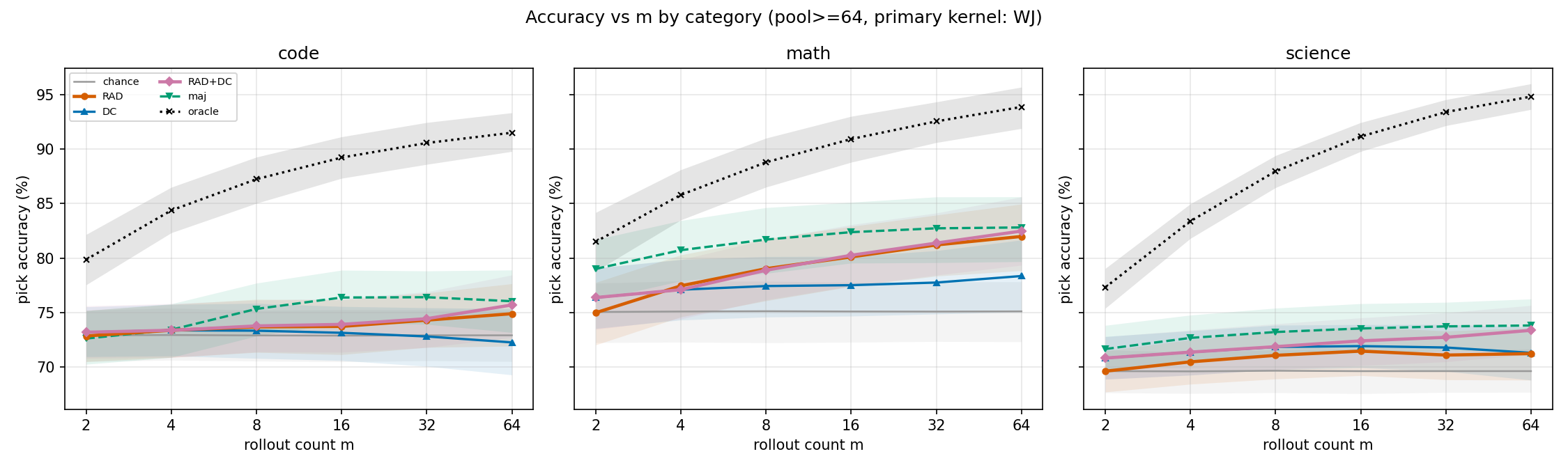}
  \caption{\textbf{Accuracy vs.\ $m$ by domain} (pool$\ge$64, weighted Jaccard, boundary anchor). The selector comparison (RAD/DC/RAD+DC vs.\ Majority and the oracle ceiling) on code, math, and science; the RAD-to-Majority gap and the oracle headroom are largest on mathematics.}
  \label{fig:rollout-scaling-cat}
\end{figure}

\paragraph{Scaling with the rollout count.} RAD's pick accuracy rises monotonically with $m$ (Table~\ref{tab:rollout-scaling}, Figure~\ref{fig:rollout-scaling}): the paired slope from $m{=}2$ to $64$ is $+3.26$\,pp at the delimiter anchor ($+2.90$ boundary), with a $95\%$ CI that excludes $0$. DeepConf does \emph{not} scale (slope CI includes $0$ at both anchors; it peaks by $m{\approx}8$ and mildly declines thereafter), the signature of a per-rollout quality signal, not a consensus one. Fused, RAD+DC scales and at $m{=}64$ statistically ties the answer-aware Majority: the paired gap RAD+DC$-$maj is $+0.11$\,pp $[-0.83,+1.06]$ at the delimiter anchor and $-0.39$\,pp $[-1.45,+0.64]$ at the boundary anchor, both including $0$ (the binary-Jaccard kernel reproduces the tie, $-0.64$\,pp $[-1.66,+0.35]$). The per-domain split is Figure~\ref{fig:rollout-scaling-cat}.

\begin{table}[!ht]
  \centering\footnotesize
  \setlength{\tabcolsep}{5pt}
  \caption{\textbf{Pick accuracy (\%) vs.\ cohort size $m$} (pool$\ge$64; weighted Jaccard; answer-string-free selectors except maj). Paired slopes $m{=}2\!\to\!64$ (pp): boundary RAD $+2.90\,[{+}2.04,{+}3.74]$, DC $+0.37\,[{-}0.56,{+}1.35]$, RAD+DC $+3.30\,[{+}2.53,{+}4.10]$; delimiter RAD $+3.26\,[{+}2.38,{+}4.09]$, DC $-0.36\,[{-}1.33,{+}0.60]$, RAD+DC $+2.98\,[{+}2.27,{+}3.75]$. Paired gap RAD+DC$-$maj at $m{=}64$: boundary $-0.39\,[{-}1.45,{+}0.64]$, delimiter $+0.11\,[{-}0.83,{+}1.06]$ (both include $0$).}
  \label{tab:rollout-scaling}
  \begin{tabular}{rrrrrrr}
    \toprule
    $m$ & chance & RAD & DC & RAD+DC & maj & oracle \\
    \midrule
    \multicolumn{7}{l}{\emph{Boundary anchor} ($n=2832$)}\\
    2  & 71.78 & 71.75 & 72.73 & 72.73 & 73.51 & 78.99 \\
    4  & 71.79 & 72.85 & 73.20 & 73.20 & 74.61 & 84.21 \\
    8  & 71.81 & 73.56 & 73.48 & 73.94 & 75.67 & 87.92 \\
    16 & 71.78 & 74.00 & 73.48 & 74.53 & 76.29 & 90.52 \\
    32 & 71.79 & 74.23 & 73.37 & 75.09 & 76.47 & 92.36 \\
    64 & 71.79 & 74.65 & 73.09 & 76.02 & 76.41 & 93.61 \\
    \midrule
    \multicolumn{7}{l}{\emph{Delimiter anchor} ($n=2636$, deployed)}\\
    2  & 74.38 & 74.36 & 74.98 & 74.98 & 75.57 & 81.10 \\
    4  & 74.38 & 75.83 & 75.25 & 75.25 & 76.42 & 85.90 \\
    8  & 74.40 & 76.48 & 75.41 & 76.45 & 77.32 & 89.32 \\
    16 & 74.37 & 76.79 & 75.30 & 76.99 & 77.82 & 91.68 \\
    32 & 74.38 & 77.05 & 75.06 & 77.31 & 77.93 & 93.38 \\
    64 & 74.39 & 77.62 & 74.62 & 77.96 & 77.85 & 94.50 \\
    \bottomrule
  \end{tabular}
\end{table}

\begin{figure}[!ht]
  \centering
  \includegraphics[width=\linewidth]{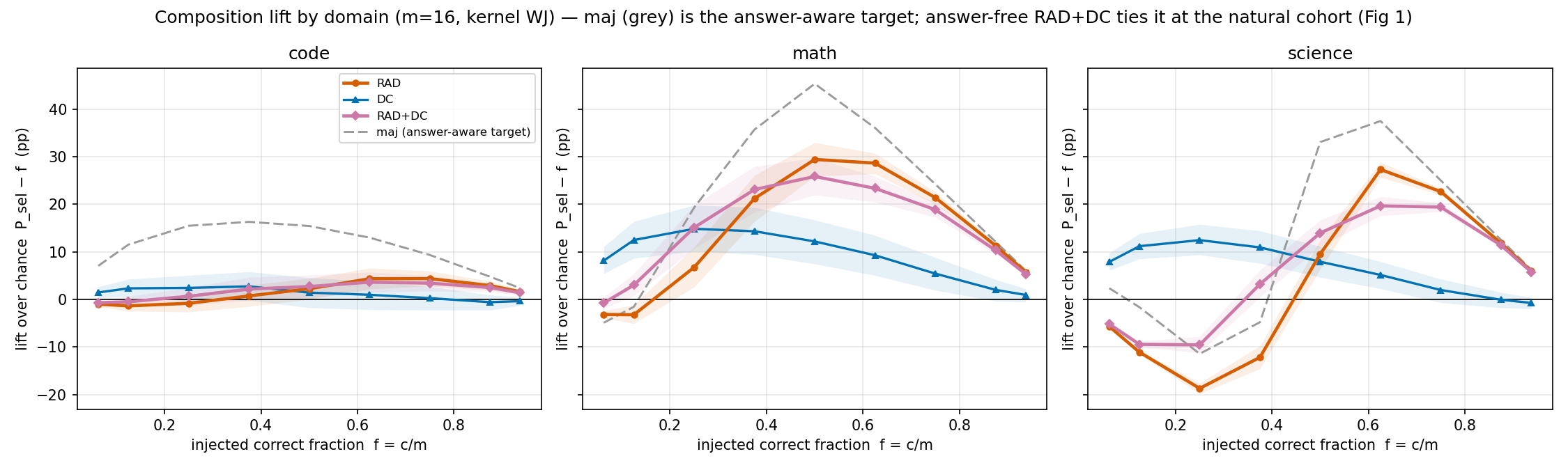}
  \caption{\textbf{Composition lift by domain} ($m{=}16$, weighted Jaccard, boundary anchor): lift over chance vs.\ injected correct fraction $f$, with the answer-aware Majority (grey) as the target. The RAD consensus S-curve is strongest on mathematics and flattest on code, matching the domain ordering of Table~\ref{tab:rollout-domain}.}
  \label{fig:rollout-composition-cat}
\end{figure}

\begin{figure}[!ht]
  \centering
  \includegraphics[width=0.62\linewidth]{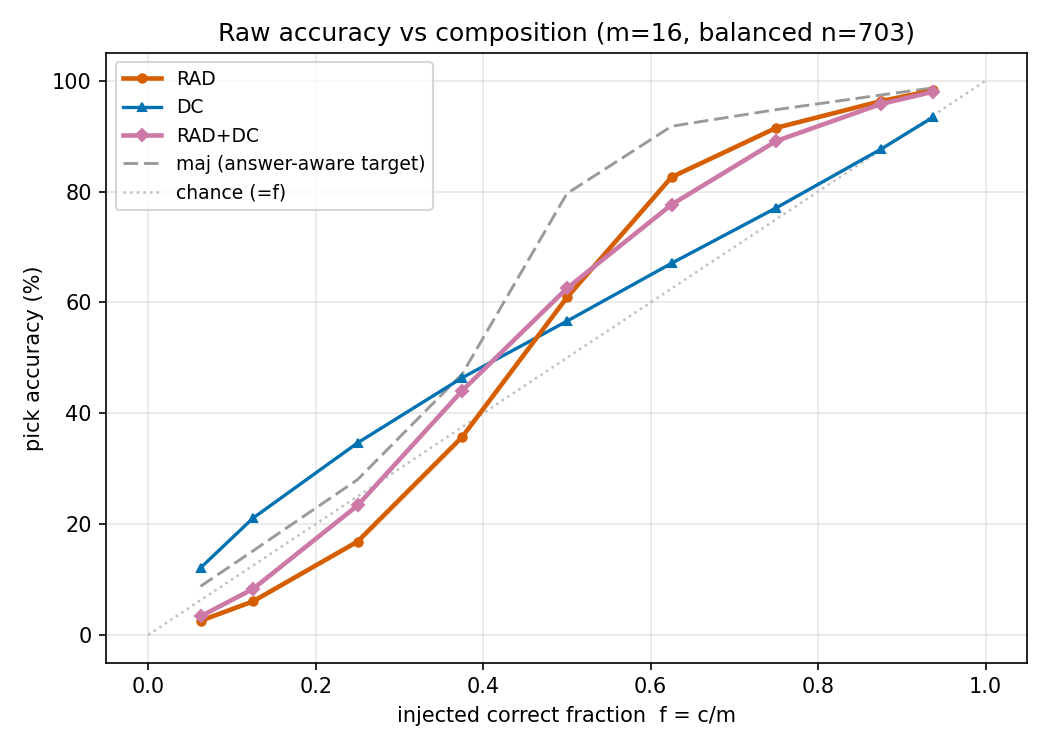}
  \caption{\textbf{Raw pick accuracy behind the composition lift} ($m{=}16$ balanced, $n{=}703$, boundary anchor): absolute pick accuracy vs.\ $f$, with the $y{=}f$ chance diagonal. RAD overtakes DeepConf once correct answers approach a majority ($f{\gtrsim}0.4$); below that DeepConf is higher.}
  \label{fig:rollout-composition-raw}
\end{figure}

\begin{figure}[!ht]
  \centering
  \includegraphics[width=0.62\linewidth]{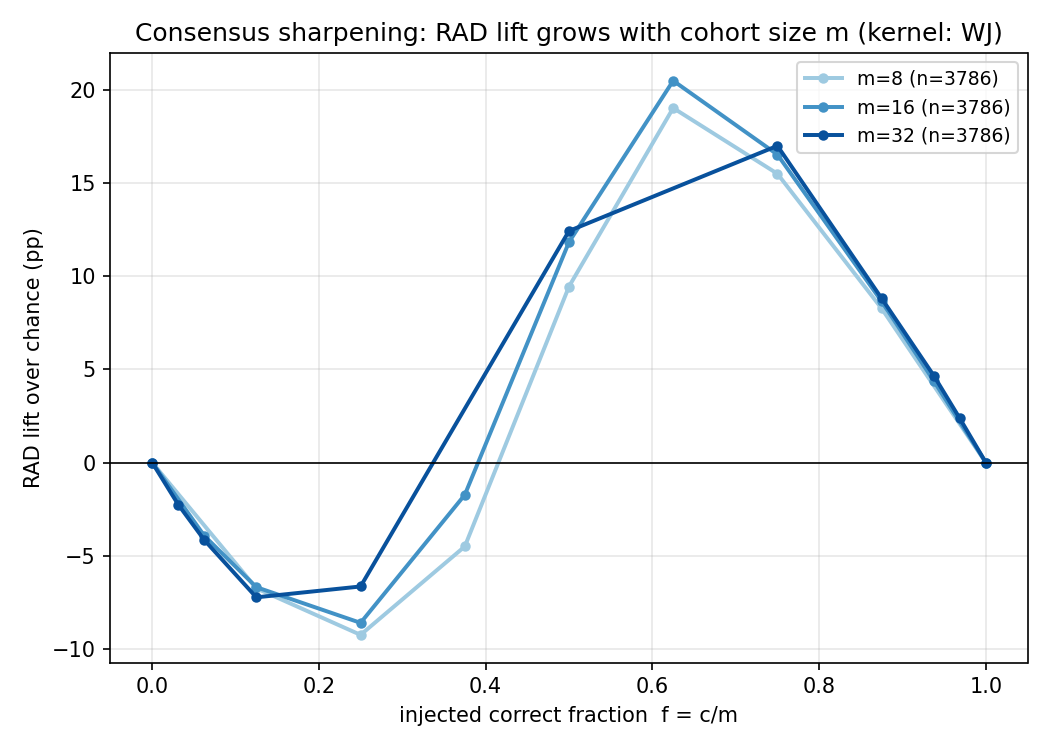}
  \caption{\textbf{The consensus S-curve sharpens with cohort size.} RAD lift over chance vs.\ $f$ for $m{=}8/16/32$ (feasible set, $n{=}3786$, boundary anchor): the S-curve amplitude grows with $m$, the mechanism behind RAD's scaling in Table~\ref{tab:rollout-scaling}.}
  \label{fig:rollout-sharpen}
\end{figure}

\paragraph{Composition control.} Fixing $m{=}16$ on a difficulty-balanced fixed set ($\ge 15$ correct $\wedge \ge 15$ incorrect, $n{=}703$ boundary / $664$ delimiter) and injecting $c$ correct rollouts, RAD's lift over chance is a clean S-curve in $f$ (Table~\ref{tab:rollout-composition}, Figures~\ref{fig:rad-composition} and \ref{fig:rollout-composition-cat}): strongly negative when correct answers are a minority ($-8.2$\,pp at $f{=}0.25$), crossing zero near $f{\approx}0.4$, and large-positive once a majority ($+20$ boundary / $+24$ delimiter\,pp at $f{=}0.62$). DeepConf is the complementary signal, an inverted-U positive in the rare-correct regime where RAD is negative (peak $+9.7$\,pp at $f{=}0.25$), with no sign flip. So RAD+DC inherits both (e.g.\ boundary $c{=}8$: RAD $+10.9 \to$ RAD+DC $+12.5$; $c{=}1$: $-3.7 \to -2.9$). Beyond this regime-complementarity, the two are near-orthogonal as per-rollout signals (mean within-problem Pearson $r=0.03$ between the DeepConf cBW scalar and the $K$-NN routing density, under both kernels; Table~\ref{tab:dc-density-corr}), so RAD+DC fuses non-redundant axes rather than re-weighting one. The answer-aware maj $c{=}1$ entry is a lowest-id tie-break artifact (it flips $+2.5$ boundary $\to -1.4$ delimiter) and is omitted. The raw pick accuracies behind these lifts (with the $y{=}f$ chance diagonal) are in Figure~\ref{fig:rollout-composition-raw}, and the S-curve amplitude grows with $m$ (Figure~\ref{fig:rollout-sharpen}).

\begin{table}[!ht]
  \centering\footnotesize
  \setlength{\tabcolsep}{4pt}
  \caption{\textbf{Composition control: lift over chance (pp) at $m{=}16$} on the difficulty-balanced fixed set (inject $c$ correct rollouts, $f{=}c/16$; $n{=}703$ boundary / $664$ delimiter). RAD's S-curve (negative when correct is a minority, positive once a majority) is the consensus signature; DeepConf is positive in the rare-correct (low-$f$) regime where RAD is negative.}
  \label{tab:rollout-composition}
  \begin{tabular}{rr rrr rrr}
    \toprule
    & & \multicolumn{3}{c}{Boundary anchor} & \multicolumn{3}{c}{Delimiter anchor}\\
    \cmidrule(lr){3-5}\cmidrule(lr){6-8}
    $c$ & $f$ & RAD & DC & RAD+DC & RAD & DC & RAD+DC \\
    \midrule
    1  & 0.06 & $-3.70$ & $+5.85$ & $-2.90$ & $-3.81$ & $+3.55$ & $-3.02$ \\
    2  & 0.12 & $-6.47$ & $+8.55$ & $-4.22$ & $-6.69$ & $+5.39$ & $-4.53$ \\
    4  & 0.25 & $-8.16$ & $+9.65$ & $-1.65$ & $-8.16$ & $+6.07$ & $-0.95$ \\
    6  & 0.38 & $-1.77$ & $+8.91$ & $+6.58$ & $-0.01$ & $+5.47$ & $+6.36$ \\
    8  & 0.50 & $+10.89$ & $+6.61$ & $+12.53$ & $+17.23$ & $+3.57$ & $+13.45$ \\
    10 & 0.62 & $+20.11$ & $+4.56$ & $+15.15$ & $+23.77$ & $+2.07$ & $+16.86$ \\
    12 & 0.75 & $+16.49$ & $+2.06$ & $+14.12$ & $+18.08$ & $+0.31$ & $+15.72$ \\
    14 & 0.88 & $+8.82$ & $+0.15$ & $+8.30$ & $+9.43$ & $-0.84$ & $+9.07$ \\
    15 & 0.94 & $+4.51$ & $-0.30$ & $+4.27$ & $+4.77$ & $-0.84$ & $+4.57$ \\
    \bottomrule
  \end{tabular}
\end{table}

\paragraph{The single-needle regime.} When exactly one rollout is correct ($c{=}1$, $f{=}1/m$) the dichotomy is sharpest (Table~\ref{tab:rollout-needle}): DeepConf finds the needle above chance at every $m$ ($+7.2/+5.2/+2.9$\,pp boundary at $m{=}8/16/32$) while RAD is below chance ($-6.8/-3.9/-2.3$), since a lone correct rollout is by definition outside the consensus basin. The operational corollary is that low-pass-rate / single-needle regimes should prefer DeepConf to RAD. The DeepConf needle lift is family-skewed (Qwen-heavy); we report it pooled and do not break it out per family.

\begin{table}[!ht]
  \centering\footnotesize
  \caption{\textbf{Single-needle regime ($c{=}1$): lift over chance (pp).} Exactly one correct rollout ($f{=}1/m$). DeepConf is above chance at every $m$; RAD is below chance, since a lone correct rollout is outside the consensus basin.}
  \label{tab:rollout-needle}
  \begin{tabular}{rrrrr}
    \toprule
    $m$ & RAD & DC & RAD+DC & $n$ \\
    \midrule
    \multicolumn{5}{l}{\emph{Boundary anchor}}\\
    8  & $-6.75$ & $+7.23$ & $-4.05$ & 1440 \\
    16 & $-3.93$ & $+5.23$ & $-2.94$ & 1140 \\
    32 & $-2.26$ & $+2.92$ & $-1.72$ & 738 \\
    \midrule
    \multicolumn{5}{l}{\emph{Delimiter anchor}}\\
    8  & $-7.21$ & $+5.01$ & $-4.80$ & 1418 \\
    16 & $-3.96$ & $+3.27$ & $-3.10$ & 1091 \\
    32 & $-2.07$ & $+2.03$ & $-1.69$ & 699 \\
    \bottomrule
  \end{tabular}
\end{table}

\paragraph{Domain dependence.} RAD's lift over chance at $m{=}64$ orders mathematics $\gg$ science, code (Table~\ref{tab:rollout-domain}): $+6.9$\,pp on math vs.\ $+1.6$ (science) / $+2.0$ (code) at the boundary anchor, with all three CIs excluding $0$ and a Holm correction over the three domains leaving the calls unchanged. DeepConf mirrors RAD on math but is null on code at both anchors (and on science at the delimiter anchor). This is the same math\,$>$\,science\,$>$\,code ordering reported for selector intelligence $\Delta_{\mathrm{sel}}$ in Table~\ref{tab:domain}, here measured as lift over chance rather than over the cohort baseline; the value of routing-space consensus is bounded by how routing-coherent correct reasoning is: high on math and marginal elsewhere.

\begin{table}[!ht]
  \centering\footnotesize
  \setlength{\tabcolsep}{4pt}
  \caption{\textbf{Domain dependence: RAD/DC/RAD+DC lift over chance (pp) at $m{=}64$}, with $95\%$ cluster-bootstrap CIs. Mathematics $\gg$ science, code; a Holm correction over the three domains leaves the RAD calls unchanged. ($n$ per domain: math $605/592$, science $1355/1223$, code $872/821$ for boundary\,/\,delimiter.)}
  \label{tab:rollout-domain}
  \begin{tabular}{lrrr}
    \toprule
    Domain & RAD & DC & RAD+DC \\
    \midrule
    \multicolumn{4}{l}{\emph{Boundary anchor}}\\
    math    & $+6.88\,[{+}5.20,{+}8.54]$ & $+3.25\,[{+}1.13,{+}5.42]$ & $+7.38\,[{+}5.62,{+}9.16]$ \\
    science & $+1.61\,[{+}0.41,{+}2.85]$ & $+1.68\,[{+}0.19,{+}3.19]$ & $+3.75\,[{+}2.62,{+}4.94]$ \\
    code    & $+1.99\,[{+}0.35,{+}3.63]$ & $-0.65\,[{-}2.42,{+}1.10]$ & $+2.79\,[{+}1.05,{+}4.43]$ \\
    \midrule
    \multicolumn{4}{l}{\emph{Delimiter anchor}}\\
    math    & $+6.19\,[{+}4.75,{+}7.77]$ & $+3.32\,[{+}1.24,{+}5.46]$ & $+6.70\,[{+}5.02,{+}8.31]$ \\
    science & $+2.20\,[{+}1.06,{+}3.23]$ & $-0.67\,[{-}2.14,{+}0.77]$ & $+2.36\,[{+}1.26,{+}3.55]$ \\
    code    & $+2.64\,[{+}0.90,{+}4.38]$ & $-0.65\,[{-}2.41,{+}1.13]$ & $+3.12\,[{+}1.58,{+}4.75]$ \\
    \bottomrule
  \end{tabular}
\end{table}

\paragraph{DeepConf on the natural distribution.} The composition control above injects a uniform sweep of $f$; on the \emph{natural} pool, by contrast, most cohorts are already converged (correct answers a clear majority), the regime where DeepConf's complementary lift is smallest. This is why adding DeepConf does not significantly change RAD on the natural distribution (paired exact McNemar n.s.\ in every well-posed cell after Holm correction, Table~\ref{tab:dc}) even though the composition control and the per-rollout correlation (Table~\ref{tab:dc-density-corr}) show the two signals are genuinely near-orthogonal: that near-orthogonality is realised mainly in the rare-correct tail, which is under-represented at the natural cohort. Under weighted Jaccard no cell is significantly harmful.

\begin{table}[t]
  \centering
  \footnotesize
  \caption{\textbf{DeepConf confidence and RAD routing density are near-orthogonal axes.} Mean within-problem Pearson correlation between the per-rollout DeepConf cBW scalar and the $K$-NN routing density ($K{=}10$) over $3{,}765$ pools ($\mathrm{sd}=0.26$); near zero under both kernels and in every split, so RAD+DC fuses two non-redundant axes rather than re-weighting one.}
  \label{tab:dc-density-corr}
  \begin{tabular}{@{}lcccc@{}}
    \toprule
    Kernel & All & Well-posed & Code & \% pools $|r|<0.2$ \\
    \midrule
    Weighted Jaccard (deployed) & $0.03$ & $0.02$ & $0.05$ & $61$ \\
    Binary Jaccard              & $0.03$ & $0.02$ & $0.06$ & $60$ \\
    \bottomrule
  \end{tabular}
\end{table}

\begin{figure}[!ht]
  \centering
  \includegraphics[width=\linewidth]{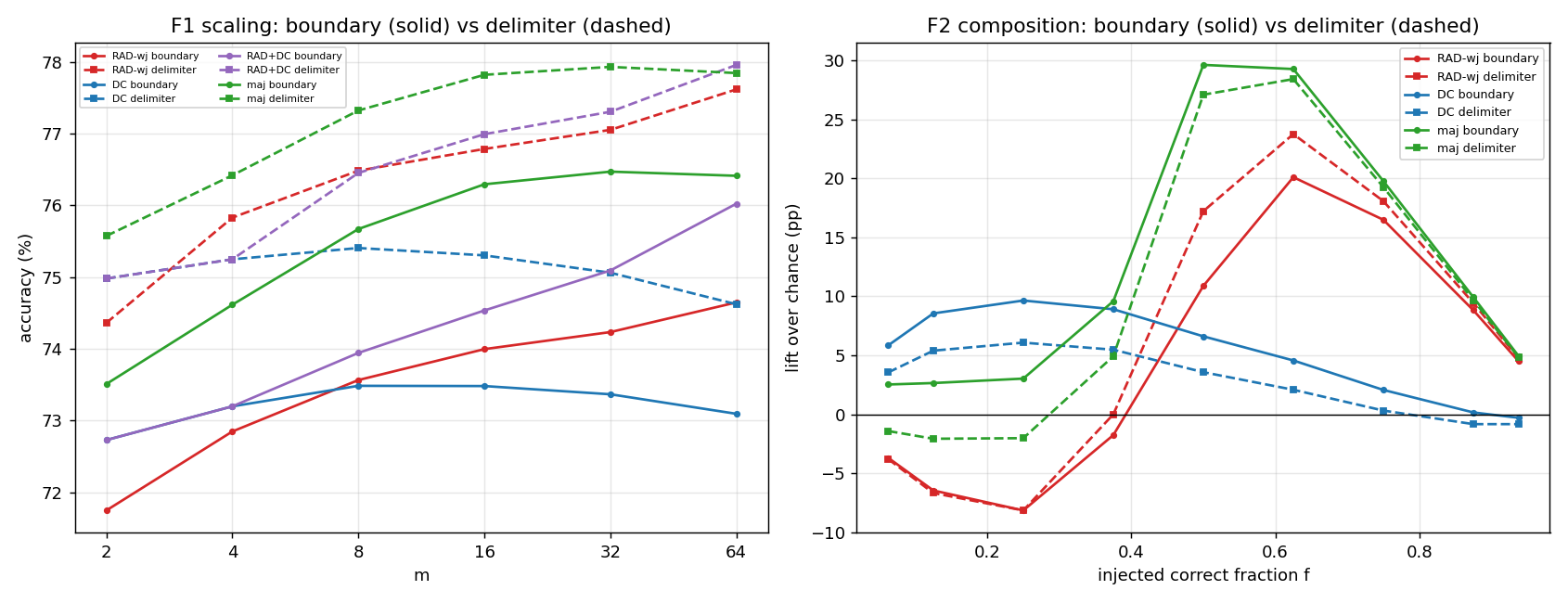}
  \caption{\textbf{Cross-anchor robustness: boundary (solid) vs.\ delimiter (dashed).} \emph{Left:} accuracy vs.\ $m$. \emph{Right:} composition lift vs.\ $f$. The scaling, saturation, S-curve, and Majority-tie all replicate at both anchors (weighted Jaccard).}
  \label{fig:rollout-anchor}
\end{figure}

\paragraph{Cross-anchor robustness.} Every finding above replicates at both anchors (Figure~\ref{fig:rollout-anchor}): RAD scales (slope $+2.90$ boundary / $+3.26$ delimiter), DeepConf does not, the composition S-curve preserves its shape and sign flip, the single-needle dichotomy holds, and RAD+DC ties Majority at $m{=}64$ ($-0.39$ / $+0.11$\,pp). The boundary anchor (read at the end-of-thinking marker) and the delimiter anchor (read at the boxed-answer / code-fence delimiter) thus expose the same mechanism; the deployed selector uses the delimiter anchor (\S\ref{sec:method}).

\subsection{Failure case studies}
\label{app:case-studies}

Table~\ref{tab:case-studies} summarises two qualitative regimes that route geometry exposes; we then anchor each regime to the data via a structural signature.

\begin{table}[!ht]
  \centering
  \footnotesize
  \caption{\textbf{Qualitative regimes for routing-based selection.} Two classes of cases that recur across the benchmark and motivate RAD's selection rule.}
  \label{tab:case-studies}
  \begin{tabular}{p{0.21\linewidth}p{0.22\linewidth}p{0.22\linewidth}p{0.22\linewidth}}
    \toprule
    Regime & What route geometry shows & What RAD does & Lesson \\
    \midrule
    Code-generation case & Exact-string voting degenerates; one route basin is denser & RAD selects the basin centre and returns a direct pass@1 trajectory & Routing density operates where lexical voting cannot \\
    False consensus & The largest route basin is wrong & RAD follows the dense wrong basin & Basin agreement is not truth \\
    \bottomrule
  \end{tabular}
\end{table}

We identify the two regimes using deterministic structural predicates over the rollout pool:

\begin{itemize}
\item \emph{Tie-breaking success.} Problems on which textual Majority's tie set has size $\ge 3$ but RAD's route density has a clear single argmax that matches the correct basin. These are over-represented on AIME25, BRUMO25, and (for OSS Thinking models) LiveCodeBench.
\item \emph{False consensus.} Problems on which the route basin selected by RAD has $\ge 32$ rollouts and is wrong. These are concentrated on the hardest difficulty bucket of GPQA (where many rollouts converge to the same plausible-but-wrong answer choice) and a small subset of HMMT25.
\end{itemize}

%% file: tables/marker_only_table.tex
\providecommand{\best}[1]{\textbf{#1}}
\providecommand{\mdltwo}[2]{\multirow{2}{*}{\shortstack[l]{#1\\{\scriptsize #2}}}}
\definecolor{dgain}{RGB}{15,120,75}
\definecolor{dloss}{RGB}{176,0,32}
\definecolor{w16blue}{RGB}{225,244,248}
\begin{table}[t]
  \centering
  \footnotesize
  \setlength{\tabcolsep}{4.0pt}
  \renewcommand{\arraystretch}{1.15}
  \caption{\textbf{Strict marker-only readout (\texttt{delimiter@marker}) across all ten configurations and six datasets.} Density-rank selector ($K{=}10$, Weighted-Jaccard) at the answer-opening \emph{delimiter} anchor (\texttt{\textbackslash boxed} on math/GPQA, a code fence on LCBv5), with the routing window reduced to the \emph{marker token(s) themselves} (\texttt{delimiter@marker}), so it conditions on \emph{no} answer-region tokens (App.~\ref{app:no-answer}), unlike the headline \texttt{delimiter@16} of Table~\ref{tab:main}. Each cell is density-rank pick accuracy (\%); the green/red {\scriptsize\textcolor{dgain}{($\pm\Delta$)}} is the gain over that cell's Avg@64 random-rollout floor. All RAD rows are the marker-only readout. Pooled over the well-posed math$+$GPQA pool ($n{=}3180$, problem-weighted), \texttt{delimiter@marker} reaches RAD $71.1$ / RAD$_{+\mathrm{DC}}$ $72.8$ versus Majority $73.6$ and the Avg@64 floor $66.3$, i.e.\ it recovers most of the lift while reading no answer content, and the headline \texttt{delimiter@16} adds only the remaining ${\sim}1$--$3$\,pp (reaching $73.9$ / $74.2$). On code ($n{=}1670$), where a $16$-token window cannot span a full solution and exact-string voting degenerates, \texttt{delimiter@marker} still lifts RAD to $65.3$ ($+4.0$ over Avg@64). Weighting is not load-bearing (binary Jaccard is equivalent, App.~\ref{app:wjja}).}
  \label{tab:marker-only}
  \begin{tabular}{@{}ll rrrr r r @{\hspace{9pt}} r@{}}
    \toprule
    \multirow{2}{*}{\textbf{Model}} & \multirow{2}{*}{\textbf{Method}} & \multicolumn{4}{c}{\textbf{Math Reasoning}} & \multirow{2}{*}{\textbf{GPQA}} & \textbf{Avg.} & \multirow{2}{*}{\textbf{Code}} \\
    \cmidrule(lr){3-6}
     & & \textbf{AIME24} & \textbf{AIME25} & \textbf{BRUMO25} & \textbf{HMMT25} & & {\scriptsize math+GPQA} & \\
    \midrule
    \mdltwo{gpt-oss-20B}{Low} & \cellcolor{w16blue}\best{RAD} & \cellcolor{w16blue}50.0\,{\scriptsize\textcolor{dgain}{(+6.4)}} & \cellcolor{w16blue}36.7\,{\scriptsize\textcolor{dgain}{(+2.3)}} & \cellcolor{w16blue}46.7\,{\scriptsize\textcolor{dgain}{(+3.4)}} & \cellcolor{w16blue}23.3\,{\scriptsize\textcolor{dgain}{(+4.3)}} & \cellcolor{w16blue}57.1\,{\scriptsize\textcolor{dgain}{(+0.8)}} & \cellcolor{w16blue}50.3\,{\scriptsize\textcolor{dgain}{(+2.0)}} & \cellcolor{w16blue}35.9\,{\scriptsize\textcolor{dgain}{(+2.4)}} \\
     & \cellcolor{w16blue}\best{RAD$_{+\mathrm{DC}}$} & \cellcolor{w16blue}63.3\,{\scriptsize\textcolor{dgain}{(+19.7)}} & \cellcolor{w16blue}43.3\,{\scriptsize\textcolor{dgain}{(+9.0)}} & \cellcolor{w16blue}50.0\,{\scriptsize\textcolor{dgain}{(+6.8)}} & \cellcolor{w16blue}30.0\,{\scriptsize\textcolor{dgain}{(+10.9)}} & \cellcolor{w16blue}59.1\,{\scriptsize\textcolor{dgain}{(+2.8)}} & \cellcolor{w16blue}54.4\,{\scriptsize\textcolor{dgain}{(+6.1)}} & \cellcolor{w16blue}34.7\,{\scriptsize\textcolor{dgain}{(+1.2)}} \\
    \addlinespace[2pt]
    \mdltwo{gpt-oss-20B}{Med} & \cellcolor{w16blue}\best{RAD} & \cellcolor{w16blue}80.0\,{\scriptsize\textcolor{dgain}{(+5.2)}} & \cellcolor{w16blue}73.3\,{\scriptsize\textcolor{dgain}{(+4.6)}} & \cellcolor{w16blue}86.7\,{\scriptsize\textcolor{dgain}{(+14.1)}} & \cellcolor{w16blue}56.7\,{\scriptsize\textcolor{dgain}{(+6.9)}} & \cellcolor{w16blue}65.2\,{\scriptsize\textcolor{dgain}{(+0.8)}} & \cellcolor{w16blue}68.6\,{\scriptsize\textcolor{dgain}{(+3.4)}} & \cellcolor{w16blue}70.1\,{\scriptsize\textcolor{dgain}{(+2.7)}} \\
     & \cellcolor{w16blue}\best{RAD$_{+\mathrm{DC}}$} & \cellcolor{w16blue}86.7\,{\scriptsize\textcolor{dgain}{(+11.8)}} & \cellcolor{w16blue}86.7\,{\scriptsize\textcolor{dgain}{(+18.0)}} & \cellcolor{w16blue}86.7\,{\scriptsize\textcolor{dgain}{(+14.1)}} & \cellcolor{w16blue}56.7\,{\scriptsize\textcolor{dgain}{(+6.9)}} & \cellcolor{w16blue}68.2\,{\scriptsize\textcolor{dgain}{(+3.8)}} & \cellcolor{w16blue}72.3\,{\scriptsize\textcolor{dgain}{(+7.2)}} & \cellcolor{w16blue}70.7\,{\scriptsize\textcolor{dgain}{(+3.3)}} \\
    \addlinespace[2pt]
    \mdltwo{gpt-oss-20B}{High} & \cellcolor{w16blue}\best{RAD} & \cellcolor{w16blue}86.7\,{\scriptsize\textcolor{dgain}{(+10.1)}} & \cellcolor{w16blue}86.7\,{\scriptsize\textcolor{dgain}{(+11.4)}} & \cellcolor{w16blue}90.0\,{\scriptsize\textcolor{dgain}{(+13.3)}} & \cellcolor{w16blue}83.3\,{\scriptsize\textcolor{dgain}{(+28.5)}} & \cellcolor{w16blue}67.2\,{\scriptsize\textcolor{dgain}{(+7.9)}} & \cellcolor{w16blue}74.5\,{\scriptsize\textcolor{dgain}{(+10.9)}} & \cellcolor{w16blue}71.3\,{\scriptsize\textcolor{dgain}{(+19.8)}} \\
     & \cellcolor{w16blue}\best{RAD$_{+\mathrm{DC}}$} & \cellcolor{w16blue}86.7\,{\scriptsize\textcolor{dgain}{(+10.1)}} & \cellcolor{w16blue}86.7\,{\scriptsize\textcolor{dgain}{(+11.4)}} & \cellcolor{w16blue}90.0\,{\scriptsize\textcolor{dgain}{(+13.3)}} & \cellcolor{w16blue}76.7\,{\scriptsize\textcolor{dgain}{(+21.8)}} & \cellcolor{w16blue}66.2\,{\scriptsize\textcolor{dgain}{(+6.9)}} & \cellcolor{w16blue}73.3\,{\scriptsize\textcolor{dgain}{(+9.6)}} & \cellcolor{w16blue}67.7\,{\scriptsize\textcolor{dgain}{(+16.2)}} \\
    \midrule
    \mdltwo{gpt-oss-120B}{Low} & \cellcolor{w16blue}\best{RAD} & \cellcolor{w16blue}66.7\,{\scriptsize\textcolor{dgain}{(+11.0)}} & \cellcolor{w16blue}56.7\,{\scriptsize\textcolor{dgain}{(+4.4)}} & \cellcolor{w16blue}60.0\,{\scriptsize\textcolor{dgain}{(+3.5)}} & \cellcolor{w16blue}43.3\,{\scriptsize\textcolor{dgain}{(+12.5)}} & \cellcolor{w16blue}66.7\,{\scriptsize\textcolor{dgain}{(+1.4)}} & \cellcolor{w16blue}62.9\,{\scriptsize\textcolor{dgain}{(+3.8)}} & \cellcolor{w16blue}66.5\,{\scriptsize\textcolor{dgain}{(+2.1)}} \\
     & \cellcolor{w16blue}\best{RAD$_{+\mathrm{DC}}$} & \cellcolor{w16blue}53.3\,{\scriptsize\textcolor{dloss}{(-2.3)}} & \cellcolor{w16blue}66.7\,{\scriptsize\textcolor{dgain}{(+14.4)}} & \cellcolor{w16blue}53.3\,{\scriptsize\textcolor{dloss}{(-3.2)}} & \cellcolor{w16blue}36.7\,{\scriptsize\textcolor{dgain}{(+5.8)}} & \cellcolor{w16blue}68.2\,{\scriptsize\textcolor{dgain}{(+2.9)}} & \cellcolor{w16blue}62.3\,{\scriptsize\textcolor{dgain}{(+3.2)}} & \cellcolor{w16blue}67.1\,{\scriptsize\textcolor{dgain}{(+2.7)}} \\
    \addlinespace[2pt]
    \mdltwo{gpt-oss-120B}{Med} & \cellcolor{w16blue}\best{RAD} & \cellcolor{w16blue}73.3\,{\scriptsize\textcolor{dloss}{(-4.4)}} & \cellcolor{w16blue}80.0\,{\scriptsize\textcolor{dgain}{(+1.9)}} & \cellcolor{w16blue}73.3\,{\scriptsize\textcolor{dloss}{(-2.6)}} & \cellcolor{w16blue}70.0\,{\scriptsize\textcolor{dgain}{(+8.6)}} & \cellcolor{w16blue}72.2\,{\scriptsize\textcolor{dgain}{(+1.2)}} & \cellcolor{w16blue}73.0\,{\scriptsize\textcolor{dgain}{(+1.1)}} & \cellcolor{w16blue}78.4\,{\scriptsize\textcolor{dgain}{(+2.4)}} \\
     & \cellcolor{w16blue}\best{RAD$_{+\mathrm{DC}}$} & \cellcolor{w16blue}80.0\,{\scriptsize\textcolor{dgain}{(+2.3)}} & \cellcolor{w16blue}70.0\,{\scriptsize\textcolor{dloss}{(-8.1)}} & \cellcolor{w16blue}83.3\,{\scriptsize\textcolor{dgain}{(+7.4)}} & \cellcolor{w16blue}76.7\,{\scriptsize\textcolor{dgain}{(+15.3)}} & \cellcolor{w16blue}72.7\,{\scriptsize\textcolor{dgain}{(+1.7)}} & \cellcolor{w16blue}74.5\,{\scriptsize\textcolor{dgain}{(+2.7)}} & \cellcolor{w16blue}76.6\,{\scriptsize\textcolor{dgain}{(+0.6)}} \\
    \addlinespace[2pt]
    \mdltwo{gpt-oss-120B}{High} & \cellcolor{w16blue}\best{RAD} & \cellcolor{w16blue}93.3\,{\scriptsize\textcolor{dgain}{(+6.5)}} & \cellcolor{w16blue}90.0\,{\scriptsize\textcolor{dgain}{(+5.1)}} & \cellcolor{w16blue}93.3\,{\scriptsize\textcolor{dgain}{(+8.2)}} & \cellcolor{w16blue}93.3\,{\scriptsize\textcolor{dgain}{(+17.8)}} & \cellcolor{w16blue}80.8\,{\scriptsize\textcolor{dgain}{(+5.4)}} & \cellcolor{w16blue}85.2\,{\scriptsize\textcolor{dgain}{(+6.9)}} & \cellcolor{w16blue}79.0\,{\scriptsize\textcolor{dgain}{(+7.7)}} \\
     & \cellcolor{w16blue}\best{RAD$_{+\mathrm{DC}}$} & \cellcolor{w16blue}93.3\,{\scriptsize\textcolor{dgain}{(+6.5)}} & \cellcolor{w16blue}90.0\,{\scriptsize\textcolor{dgain}{(+5.1)}} & \cellcolor{w16blue}93.3\,{\scriptsize\textcolor{dgain}{(+8.2)}} & \cellcolor{w16blue}90.0\,{\scriptsize\textcolor{dgain}{(+14.5)}} & \cellcolor{w16blue}82.8\,{\scriptsize\textcolor{dgain}{(+7.4)}} & \cellcolor{w16blue}86.2\,{\scriptsize\textcolor{dgain}{(+7.9)}} & \cellcolor{w16blue}85.6\,{\scriptsize\textcolor{dgain}{(+14.3)}} \\
    \midrule
    \mdltwo{Qwen3-30B-A3B}{Think} & \cellcolor{w16blue}\best{RAD} & \cellcolor{w16blue}93.3\,{\scriptsize\textcolor{dgain}{(+5.9)}} & \cellcolor{w16blue}93.3\,{\scriptsize\textcolor{dgain}{(+12.6)}} & \cellcolor{w16blue}96.7\,{\scriptsize\textcolor{dgain}{(+11.6)}} & \cellcolor{w16blue}76.7\,{\scriptsize\textcolor{dgain}{(+14.7)}} & \cellcolor{w16blue}70.7\,{\scriptsize\textcolor{dloss}{(-0.4)}} & \cellcolor{w16blue}78.0\,{\scriptsize\textcolor{dgain}{(+4.0)}} & \cellcolor{w16blue}74.3\,{\scriptsize\textcolor{dgain}{(+4.9)}} \\
     & \cellcolor{w16blue}\best{RAD$_{+\mathrm{DC}}$} & \cellcolor{w16blue}93.3\,{\scriptsize\textcolor{dgain}{(+5.9)}} & \cellcolor{w16blue}93.3\,{\scriptsize\textcolor{dgain}{(+12.6)}} & \cellcolor{w16blue}96.7\,{\scriptsize\textcolor{dgain}{(+11.6)}} & \cellcolor{w16blue}80.0\,{\scriptsize\textcolor{dgain}{(+18.0)}} & \cellcolor{w16blue}71.2\,{\scriptsize\textcolor{dgain}{(+0.1)}} & \cellcolor{w16blue}78.6\,{\scriptsize\textcolor{dgain}{(+4.6)}} & \cellcolor{w16blue}71.9\,{\scriptsize\textcolor{dgain}{(+2.6)}} \\
    \addlinespace[2pt]
    \mdltwo{Qwen3-30B-A3B}{Inst} & \cellcolor{w16blue}\best{RAD} & \cellcolor{w16blue}80.0\,{\scriptsize\textcolor{dgain}{(+8.0)}} & \cellcolor{w16blue}66.7\,{\scriptsize\textcolor{dgain}{(+6.1)}} & \cellcolor{w16blue}76.7\,{\scriptsize\textcolor{dgain}{(+7.0)}} & \cellcolor{w16blue}40.0\,{\scriptsize\textcolor{dloss}{(-1.2)}} & \cellcolor{w16blue}67.7\,{\scriptsize\textcolor{dgain}{(+1.5)}} & \cellcolor{w16blue}67.0\,{\scriptsize\textcolor{dgain}{(+2.8)}} & \cellcolor{w16blue}42.5\,{\scriptsize\textcolor{dloss}{(-4.5)}} \\
     & \cellcolor{w16blue}\best{RAD$_{+\mathrm{DC}}$} & \cellcolor{w16blue}86.7\,{\scriptsize\textcolor{dgain}{(+14.6)}} & \cellcolor{w16blue}70.0\,{\scriptsize\textcolor{dgain}{(+9.4)}} & \cellcolor{w16blue}80.0\,{\scriptsize\textcolor{dgain}{(+10.3)}} & \cellcolor{w16blue}43.3\,{\scriptsize\textcolor{dgain}{(+2.1)}} & \cellcolor{w16blue}68.2\,{\scriptsize\textcolor{dgain}{(+2.0)}} & \cellcolor{w16blue}68.9\,{\scriptsize\textcolor{dgain}{(+4.7)}} & \cellcolor{w16blue}44.3\,{\scriptsize\textcolor{dloss}{(-2.7)}} \\
    \midrule
    \mdltwo{Qwen3-Next-80B}{Think} & \cellcolor{w16blue}\best{RAD} & \cellcolor{w16blue}93.3\,{\scriptsize\textcolor{dgain}{(+5.1)}} & \cellcolor{w16blue}90.0\,{\scriptsize\textcolor{dgain}{(+7.7)}} & \cellcolor{w16blue}93.3\,{\scriptsize\textcolor{dgain}{(+9.0)}} & \cellcolor{w16blue}76.7\,{\scriptsize\textcolor{dgain}{(+12.9)}} & \cellcolor{w16blue}76.8\,{\scriptsize\textcolor{dgain}{(+16.0)}} & \cellcolor{w16blue}81.1\,{\scriptsize\textcolor{dgain}{(+13.2)}} & \cellcolor{w16blue}70.7\,{\scriptsize\textcolor{dgain}{(+0.5)}} \\
     & \cellcolor{w16blue}\best{RAD$_{+\mathrm{DC}}$} & \cellcolor{w16blue}93.3\,{\scriptsize\textcolor{dgain}{(+5.1)}} & \cellcolor{w16blue}90.0\,{\scriptsize\textcolor{dgain}{(+7.7)}} & \cellcolor{w16blue}96.7\,{\scriptsize\textcolor{dgain}{(+12.3)}} & \cellcolor{w16blue}86.7\,{\scriptsize\textcolor{dgain}{(+22.9)}} & \cellcolor{w16blue}77.3\,{\scriptsize\textcolor{dgain}{(+16.5)}} & \cellcolor{w16blue}82.7\,{\scriptsize\textcolor{dgain}{(+14.8)}} & \cellcolor{w16blue}70.7\,{\scriptsize\textcolor{dgain}{(+0.5)}} \\
    \addlinespace[2pt]
    \mdltwo{Qwen3-Next-80B}{Inst} & \cellcolor{w16blue}\best{RAD} & \cellcolor{w16blue}80.0\,{\scriptsize\textcolor{dgain}{(+0.5)}} & \cellcolor{w16blue}73.3\,{\scriptsize\textcolor{dgain}{(+5.3)}} & \cellcolor{w16blue}76.7\,{\scriptsize\textcolor{dloss}{(-0.2)}} & \cellcolor{w16blue}53.3\,{\scriptsize\textcolor{dgain}{(+3.1)}} & \cellcolor{w16blue}70.7\,{\scriptsize\textcolor{dloss}{(-1.5)}} & \cellcolor{w16blue}70.8\,{\scriptsize\textcolor{dloss}{(-0.1)}} & \cellcolor{w16blue}64.7\,{\scriptsize\textcolor{dgain}{(+1.8)}} \\
     & \cellcolor{w16blue}\best{RAD$_{+\mathrm{DC}}$} & \cellcolor{w16blue}83.3\,{\scriptsize\textcolor{dgain}{(+3.8)}} & \cellcolor{w16blue}76.7\,{\scriptsize\textcolor{dgain}{(+8.6)}} & \cellcolor{w16blue}83.3\,{\scriptsize\textcolor{dgain}{(+6.5)}} & \cellcolor{w16blue}60.0\,{\scriptsize\textcolor{dgain}{(+9.8)}} & \cellcolor{w16blue}74.7\,{\scriptsize\textcolor{dgain}{(+2.5)}} & \cellcolor{w16blue}75.2\,{\scriptsize\textcolor{dgain}{(+4.3)}} & \cellcolor{w16blue}64.7\,{\scriptsize\textcolor{dgain}{(+1.8)}} \\
    \bottomrule
  \end{tabular}
\end{table}

%% file: tables/wj_ja_appendix.tex

\begin{table}[t]
  \centering
  \caption{\textbf{Binary (JA) vs.\ weighted (WJ) Jaccard kernel, per anchor
    mode.} Within-cell paired ablation over 15{,}188 problem cells (8 thinking
    models $\times$ 6 datasets, cohort $\geq 2$); only the kNN similarity kernel
    differs between the two runs. $b$/$c$ are the discordant picks
    (JA-only-correct / WJ-only-correct); the exact two-sided McNemar test acts on
    them. All four anchor modes are non-significant and Holm-adjust to $1.000$;
    pooled WJ$-$JA $=+0.22$\,pp (95\% CI $[-0.18,+0.59]$, cluster bootstrap over
    shards). \emph{No detectable difference under the primary paired test.}}
  \label{tab:wjja-main}
  \small
  \begin{tabular}{lrrrrrrrr}
    \toprule
    Anchor mode & $n$ & JA (\%) & WJ (\%) & WJ$-$JA & $b$ & $c$ & McNemar $p$ & Holm $p$ \\
    \midrule
    \texttt{boundary@marker}          & 3786 & 70.71 & 71.16 & $+0.45$ & 103 & 120 & 0.284 & 1.000 \\
    \texttt{delimiter@marker}     & 3808 & 71.43 & 71.85 & $+0.42$ & 109 & 125 & 0.327 & 1.000 \\
    \texttt{delimiter@16} & 3808 & 73.58 & 73.82 & $+0.24$ &  72 &  81 & 0.518 & 1.000 \\
    \texttt{boundary@16}      & 3786 & 72.82 & 72.61 & $-0.21$ & 107 &  99 & 0.626 & 1.000 \\
    \midrule
    \textbf{Pooled}      & \textbf{15188} & \textbf{72.14} & \textbf{72.36} & \textbf{$+0.22$} & \textbf{391} & \textbf{425} & \textbf{0.248} & --- \\
    \bottomrule
  \end{tabular}
\end{table}

\begin{table}[t]
  \centering
  \caption{\textbf{JA vs.\ WJ by category, per anchor mode.} math~$=$~\{aime24,
    aime25, brumo25, hmmt25\}; code~$=$~\{livecodebench\_v5\};
    science~$=$~\{gpqa\}. No cell is significant; the largest gap
    (\texttt{delimiter@marker}$\times$math, $+1.17$\,pp) is non-significant ($p=0.108$) and does not
    survive FDR (App.~\ref{app:wjja}, BH $q=0.70$).}
  \label{tab:wjja-cat}
  \small
  \begin{tabular}{llrrrrrr}
    \toprule
    Anchor mode & Category & $n$ & JA (\%) & WJ (\%) & WJ$-$JA & $b$ & $c$ \\
    \midrule
    \texttt{boundary@marker}          & math    &  934 & 75.91 & 76.66 & $+0.75$ & 15 & 22 \\
    \texttt{boundary@marker}          & code    & 1278 & 69.41 & 70.34 & $+0.94$ & 35 & 47 \\
    \texttt{boundary@marker}          & science & 1574 & 68.68 & 68.55 & $-0.13$ & 53 & 51 \\
    \addlinespace
    \texttt{delimiter@marker}     & math    &  941 & 75.56 & 76.73 & $+1.17$ & 14 & 25 \\
    \texttt{delimiter@marker}     & code    & 1291 & 70.49 & 70.64 & $+0.15$ & 49 & 51 \\
    \texttt{delimiter@marker}     & science & 1576 & 69.73 & 69.92 & $+0.19$ & 46 & 49 \\
    \addlinespace
    \texttt{delimiter@16} & math    &  941 & 80.77 & 80.87 & $+0.11$ &  6 &  7 \\
    \texttt{delimiter@16} & code    & 1291 & 71.88 & 71.88 & $+0.00$ & 51 & 51 \\
    \texttt{delimiter@16} & science & 1576 & 70.69 & 71.19 & $+0.51$ & 15 & 23 \\
    \addlinespace
    \texttt{boundary@16}      & math    &  934 & 80.73 & 79.76 & $-0.96$ & 16 &  7 \\
    \texttt{boundary@16}      & code    & 1278 & 70.50 & 71.13 & $+0.63$ & 51 & 59 \\
    \texttt{boundary@16}      & science & 1574 & 70.01 & 69.57 & $-0.44$ & 40 & 33 \\
    \bottomrule
  \end{tabular}
\end{table}

\begin{table}[t]
  \centering
  \caption{\textbf{The kernel ablation is adequately powered.} Exact
    \emph{conditional} McNemar power (conditioning on the discordant count
    $n_{\mathrm{disc}}=b+c$) at $\alpha=0.05$ (two-sided), $1-\beta=0.80$. MDE@80\%
    and the post-hoc significance boundary are WJ$-$JA accuracy gaps (pp). Pooled
    power to detect a true $1$\,pp effect is $\approx\!1.00$ and the pooled
    significance boundary ($0.38$\,pp) exceeds the pooled observed gap ($0.22$\,pp), so a
    paper-relevant ($\geq 1$\,pp) \emph{pooled} kernel effect would almost
    certainly have been detected. The null in Table~\ref{tab:wjja-main} is
    informative, not a false negative. Per anchor mode, power@1pp is
    $0.68$--$0.84$, so a $\geq 1$\,pp effect confined to one mode could
    occasionally be missed.}
  \label{tab:wjja-power}
  \small
  \begin{tabular}{lrrrrrr}
    \toprule
    Scope & $n_{\mathrm{disc}}$ & MDE@80\% & sig.\ bound & power@1pp & power@0.5pp & obs.\ $|$WJ$-$JA$|$ \\
    \midrule
    \texttt{boundary@marker}          & 223 & 1.12 & 0.82 & 0.70 & 0.23 & 0.45 \\
    \texttt{delimiter@marker}     & 234 & 1.15 & 0.84 & 0.68 & 0.22 & 0.42 \\
    \texttt{delimiter@16} & 153 & 0.95 & 0.71 & 0.84 & 0.29 & 0.24 \\
    \texttt{boundary@16}      & 206 & 1.08 & 0.79 & 0.74 & 0.24 & 0.21 \\
    \midrule
    \textbf{Pooled}      & \textbf{816} & \textbf{0.53} & \textbf{0.38} & \textbf{$\approx$1.00} & \textbf{0.75} & \textbf{0.22} \\
    \bottomrule
  \end{tabular}
\end{table}